\documentclass[runningheads]{llncs}

\usepackage{eccv}

\usepackage{eccvabbrv}

\usepackage{graphicx}
\usepackage{booktabs}

\usepackage{amsmath}
\usepackage{amssymb}
\usepackage{pifont}
\usepackage{multirow}
\usepackage{comment}
\usepackage{xcolor}
\usepackage{gensymb}
\usepackage{microtype}

\usepackage{algorithm}
\usepackage{algpseudocode}

\algnewcommand\algorithmicinput{\textbf{Input:}}
\algnewcommand\algorithmicoutput{\textbf{Output:}}
\algnewcommand\Input{\item[\algorithmicinput]}
\algnewcommand\Output{\item[\algorithmicoutput]}

\usepackage{tikz}
\usetikzlibrary{spy,positioning}

\definecolor{light_yellow}{HTML}{FFF2CC}
\definecolor{light_green}{rgb}{0.850980392, 0.917647059, 0.82745098}
\definecolor{light_blue}{rgb}{0.788235294, 0.854901961, 0.97254902}
\definecolor{light_purple}{rgb}{0.850980392, 0.823529412, 0.91372549}
\definecolor{light_red}{rgb}{0.901960784, 0.721568627, 0.68627451}

\usepackage[accsupp]{axessibility}  

\usepackage{hyperref}

\usepackage{orcidlink}

\usepackage{dsfont}

\newif\ifshowedits

\newcommand{\addeditor}[3]{%
  \definecolor{#1color}{rgb}{#3}
  \expandafter\newcommand\csname #1\endcsname[1]{
  \ifshowedits
    {\color{#1color} ##1}
  \else
    {##1}
  \fi
  }%
  \expandafter\newcommand\csname #1rmk\endcsname[1]{
  \ifshowedits
    {\color{#1color} {\bf [#2: ##1]}}
  \else
    {}
  \fi
  }%
}

\newcommand{\createtextvar}[1]{
  \expandafter\newcommand\csname #1\endcsname{%
  {\text{#1}}
}%
}
\DeclareMathOperator*{\argmin}{arg\,min}

\newcommand{\doublezoom}[3]{

\begin{tikzpicture}[
    spy using outlines
]

\node (img) {\includegraphics[width=0.285\linewidth]{#1}};

\spy [rectangle, magnification=3.5, size=1.5cm, draw=red, connect spies, line width=0.7pt] on #2 in node [right] at (1.8, 0.80);

\spy [rectangle, magnification=3.5, size=1.5cm, draw=blue, connect spies, line width=0.7pt] on #3 in node [right] at (1.8, -0.80);

\end{tikzpicture}
}

\begin{document}

\title{Scene Generation at Absolute Scale: Utilizing Semantic and Geometric Guidance From Text for Accurate and Interpretable 3D Indoor Scene Generation} 


\titlerunning{Scene Generation at Absolute Scale}


\author{Stefan Ainetter\inst{1} \and
Thomas Deixelberger \inst{1,3} \and
Edoardo A. Dominici\inst{2,3} \and
Philipp Drescher\inst{1,3} \and
Konstantinos Vardis\inst{2} \and
Markus Steinberger\inst{1,3}
}

\authorrunning{Ainetter et al.}

\institute{Huawei Technologies, Austria \and
Huawei Technologies, Switzerland \and
Graz University of Technology, Austria
}

\maketitle

\begin{abstract}
\begin{figure}[]
    \centering
        \includegraphics[width=.98\linewidth]{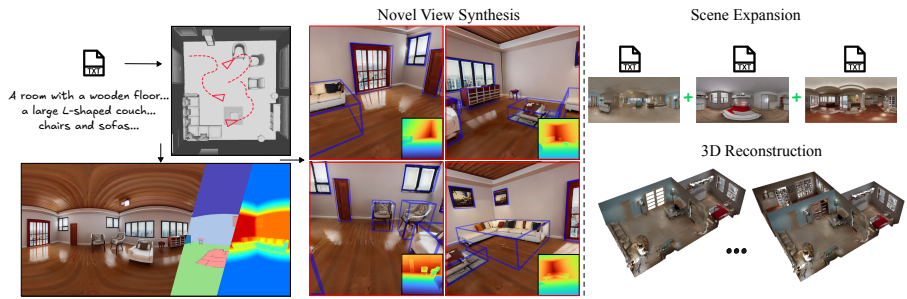} 
    \caption{\textbf{Left:} Given a textual description of an indoor scene as input, our method accurately generates a corresponding 3D representation at absolute metric scale by first predicting a global 3D scene layout.
    \textbf{Middle:} Our method additionally provides accurate 9D object poses and semantic object instance segmentation for the generated scene, bridging the gap between text-to-3D scene generation, and 3D scene interpretability and understanding.
    \textbf{Right:} One useful application of our method is progressive and seamless indoor scene expansion. As our generated scenes are in a global metric coordinate frame, they can be naturally expanded without any additional alignment or scaling.}
    \label{fig:teaser}
\end{figure}

We present GuidedSceneGen, a text-to-3D generation framework that produces metrically accurate, globally consistent, and semantically interpretable indoor scenes. Unlike prior text-driven methods that often suffer from geometric drift or scale ambiguity, our approach maintains an absolute world coordinate frame throughout the entire generation process. Starting from a textual scene description, we predict a global 3D layout encoding both semantic and geometric structure, which serves as a guiding proxy for downstream stages. A semantics- and depth-conditioned panoramic diffusion model then synthesizes 360\textdegree\ imagery aligned with the global layout, substantially improving spatial coherence. To explore unobserved regions, we employ a video diffusion model guided by optimized camera trajectories that balances coverage and collision avoidance, achieving up to 10× faster sampling compared to exhaustive path exploration. The generated views are fused using 3D Gaussian Splatting, yielding a consistent and fully navigable 3D scene in absolute scale. GuidedSceneGen enables accurate transfer of object poses and semantic labels from layout to reconstruction, and supports progressive scene expansion without re-alignment. Quantitative results and a user study demonstrate greater 3D consistency and layout plausibility compared to recent text-to-3D baselines. Project page: \url{https://d3ixi.github.io/GuidedSceneGen/}
\end{abstract}

\section{Introduction}
\label{sec:intro}

Accurate and interpretable 3D scene generation is an essential capability for computer vision, graphics, and robotics. Applications such as virtual environment design, embodied AI, and autonomous navigation rely on realistic and metrically consistent 3D reconstructions. However, synthesizing such scenes manually is time-consuming and requires extensive domain expertise. The recent success of large-scale text-to-image and video diffusion models~\cite{rombach2022high,flux2024,blattmann2023stable} has inspired a surge of text-to-3D scene generation methods~\cite{schneider_hoellein_2025_worldexplorer,yang2024layerpano3d,zhou2024dreamscene360}, where a simple textual prompt can produce complex 3D content. Yet, despite impressive visual fidelity, current models struggle to maintain global geometric coherence, metric scale, and semantic interpretability.

Existing methods typically lift a few generated 2D views into 3D space, relying on iterative inpainting or autoregressive view expansion \cite{hoellein2023text2room, scenescape2023, schneider_hoellein_2025_worldexplorer}. While effective for local consistency, such pipelines tend to accumulate semantic drift and geometric distortion, particularly when exploring larger or multi-room scenes. Even panoramic-based generation \cite{huang2025dreamcube,yang2024layerpano3d,zhou2024dreamscene360} often introduces spatial distortion and lacks absolute alignment between objects. As a result, the environments are visually plausible but physically inconsistent, limiting their usability for tasks that demand accurate scene geometry, object localization, or downstream interaction.

A promising direction is to introduce explicit geometric and semantic guidance early in the generation process. By conditioning generative models on structured 3D layouts or semantic proxies, prior work has demonstrated improved spatial grounding \cite{dominici2025dreamanywhere,schult2024controlroom3d,yang2024scenecraft}. However, these methods typically rely on discrete object retrieval or relative-scale geometry, preventing globally coherent and metrically meaningful reconstructions. Similarly, video diffusion models~\cite{VDM,wan2025wan, zhou2025stable} offer temporal consistency but lack explicit control over scene scale or camera motion, resulting in misaligned multi-view imagery.

In this work, we propose GuidedSceneGen, a text-to-3D framework that addresses these challenges by generating scenes at absolute scale while preserving semantic and geometric interpretability. Given a textual description of an indoor scene, our method first predicts a global 3D scene layout that establishes a world coordinate frame and encodes both object semantics and spatial geometry. This layout serves as a guiding proxy for all subsequent stages. We then introduce a semantics- and depth-conditioned panoramic diffusion model that generates 360° imagery consistent with the layout, producing coherent object arrangements. 
To explore unobserved areas, we leverage a video diffusion model guided by optimized camera trajectories designed for maximal coverage and collision avoidance, achieving up to 10× faster sampling than exhaustive exploration. The generated multi-view frames are fused using 3D Gaussian Splatting (3DGS) \cite{kerbl20233d}, producing a navigable 3D scene aligned in metric world space.
This approach bridges the gap between text-driven generative modeling and explicit 3D scene understanding. By maintaining absolute scale and semantic grounding throughout the pipeline, GuidedSceneGen enables interpretable, extendable, and physically meaningful scene generation. Moreover, because all components share a consistent world frame, scenes can be progressively expanded without re-scaling or re-alignment---an important property for building large, connected environments.

\noindent Our contributions can be summarized as follows:
\begin{itemize}
\item \textbf{Absolute-scale text-to-3D generation}: We present the first framework that preserves metric scale and global consistency throughout the full text-to-3D generation process.
\item \textbf{Guided panoramic diffusion}: A multi-view diffusion model conditioned on semantic and depth cues produces 360° panoramas aligned with the global layout.
\item \textbf{Optimized novel-view synthesis}: A coverage- and collision-aware trajectory sampling strategy improves geometric consistency while providing a 10× speed-up over exhaustive approaches.
\item \textbf{Interpretable 3D reconstruction}: The fused 3D scenes are metrically aligned, semantically labeled, and support seamless expansion to larger environments.
\end{itemize}
\section{Related Work}
\label{sec:related_work}

\paragraph{Iterative Image Inpainting and Lifting.}
A common approach to text-to-3D scene generation is to first synthesize a reference image with a large-scale 2D diffusion model~\cite{rombach2022high} and the lift it into 3D via iterative depth inpainting and geometry fusion. Methods like Text2Room~\cite{hoellein2023text2room} and SceneScape~\cite{scenescape2023} follow this strategy, warping images to new viewpoints and using text-guided inpainting to fill disocclusions. Others distilling color and depth through diffusion-based priors~\cite{poole2023dreamfusion, shriram2024realmdreamertextdriven3dscene, yu2023wonderjourney,yu2025wonderworld, Zhang2023Text2NeRFT3} or leverage video diffusion models to ensure locally coherent transitions~\cite{chen2025flexworld, liang2024wonderland}. More recent feed-forward architectures \cite{szymanowicz2025bolt3d} attempt to directly denoise geometry buffers for faster and more consistent reconstruction, but have not scaled to fully globally coherent navigable scenes, in part due to the limited training data.
Existing methods often rely on well-defined camera trajectories and keyframes to guide the generation~\cite{vistadream2025, schneider_hoellein_2025_worldexplorer}, limiting the coverage space that can be optimized.

\paragraph{Panoramic-based Scene Generation.}
To create immersive environments, users must be able to freely explore and perceive coherent content in all directions, which is difficult to achieve with incrementally inpainted or locally composited views. To this end, 360\degree panoramas emerged as robust and compact scene representations that capture global spatial context.
DreamAnywhere~\cite{dominici2025dreamanywhere} decomposes the panoramic image into individual objects, reconstructs them individually, and recomposes the full scene. An agentic-oriented approach based on this concepts has been proposed in Hunyuan-world \cite{hunyuanworld2025tencent}.
WorldPrompter~\cite{zhang2025generating} generates a panoramic walk-through video from text and reconstructs it into a 3DGS scene, but without an absolute metric frame or layout-grounded semantics. DreamScene360~\cite{zhou2024dreamscene360} optimizes novel off-origin views for semantic and geometric similarity with the reference views. This results in stable geometry under offset views but no new content being synthesized. LayerPano3D~\cite{yang2024layerpano3d} introduced a layered scene representation and approach that progressively peels off geometry from the panoramic images, estimates panoramic depth and harmonizes the layers for improved parallax and occlusions. 
WorldExplorer~\cite{schneider_hoellein_2025_worldexplorer} bridges panoramic scene synthesis with video diffusion, expanding the environment outward from the origin. However, it relies on extensive exploratory trajectories, which accumulate geometric and appearance drift over long sequences, limiting global scene consistency.
Together with LayerPano3D~\cite{yang2024layerpano3d} and~\cite{zhou2024dreamscene360}, WorldExplorer is the closest to our work in leveraging panoramic images and diffusion models for 3D scene generation.

\paragraph{Camera Controlled Video Generation.}
Recent video diffusion models~\cite{VDM, wan2025wan, blattmann2023stable} exhibit remarkable temporal coherence and photorealism, making them promising candidates for extending 2D generative priors into the 3D domain. To enable explicit camera trajectory conditioning, several strategies have been explored: Some works finetune pre-trained text-to-video diffusion models by augmenting their latent representations with Plücker embeddings~\cite{zhou2025stable}; others adopt view reprojection or inpainting-based conditioning~\cite{yu2024viewcrafter, ren2025gen3c3dinformedworldconsistentvideo, gu2025diffusion}. 
Voyager~\cite{voyager} generates world-consistent RGB-D video along a user-defined camera path from a single input image, accumulating a point-cloud world cache for long-range consistency; unlike our setting, it is image- and trajectory-conditioned rather than text-driven.
While such extensions make these models attractive for scene generation, their temporal consistency does not automatically translate into 3D coherence, and precise camera control or geometry-aware guidance remains challenging.

\paragraph{Compositional Scene Generation.}
Relying purely on 2D diffusion models to arrange objects end-to-end often succeeds only for small, localized compositions~\cite{epstein2024disentangled}, and typically requires complex discrete–continuous optimization~\cite{zhang2023scenewiz3d}. Introducing 3D proxies has proven beneficial: they have been combined with score distillation for layout optimization~\cite{cohen2023set, dreamsceneLi2024} and for reconstructing individual objects before scene assembly~\cite{chen2025comboverse}. SceneCraft~\cite{yang2024scenecraft} uses pre-defined bounding-boxes as proxy to semantically guide view refinement with a diffusion model; similarly ControlRoom3D~\cite{schult2024controlroom3d} iteratively transforms a pre-defined geometric proxy into a complete scene through depth-guided diffusion. Ctrl-Room~\cite{fang2025ctrl} operates directly from a pre-defined description of the 3D scene and objects, then estimates a layout of bounding boxes and a panoramic image.
SpatialGen~\cite{SpatialGen} synthesizes a scene from a separately supplied 3D layout via a layout-conditioned multi-view diffusion model, with text only conditioning the appearance branch. All these methods rely on axis-aligned bounding-box proxies, which prevent transfer of object geometry, pose, and semantics to the final scene.
In contrast to these methods, we generate the 3D proxy information entirely from text, enabling flexible and fully generative scene synthesis with high 3D consistency, while preserving accurate object pose and semantics end-to-end.
\section{Method}
\label{sec:method}

\begin{figure*}[] 
    \centering
    \includegraphics[trim={3.3cm 3.7cm 1.2cm 0.9cm},clip,width=.93\linewidth]{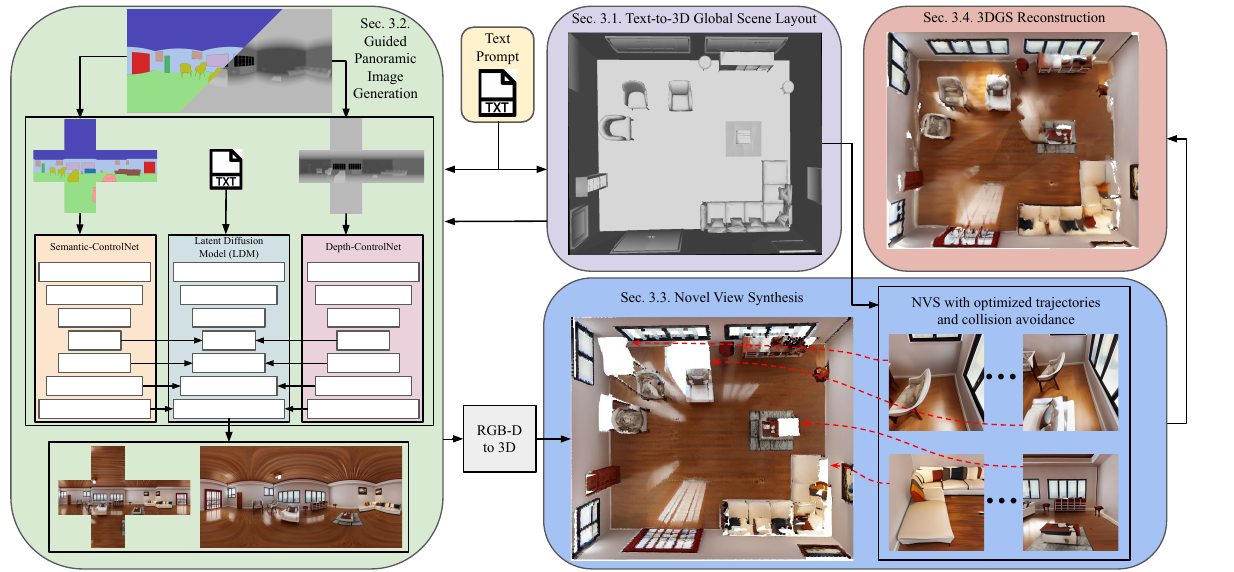}
    \caption{\textbf{Method Overview.} Given a text prompt \fcolorbox{black}{light_yellow}{\rule{0pt}{2pt}\rule{2pt}{0pt}} describing an indoor scene, we first estimate a 3D global scene layout \fcolorbox{black}{light_purple}{\rule{0pt}{2pt}\rule{2pt}{0pt}}. From this, we derive the guidance signals for panorama generation in the form of depth and semantic renderings. Our guided panoramic image generator \fcolorbox{black}{light_green}{\rule{0pt}{2pt}\rule{2pt}{0pt}} utilizes these renderings to generate a consistent and realistic RGB panoramic image that accurately aligns with the 3D global scene layout. As shown in \fcolorbox{black}{light_blue}{\rule{0pt}{2pt}\rule{2pt}{0pt}}, we then derive optimized camera trajectories for novel view synthesis (NVS) to cover unobserved areas, with the focus on avoiding collisions of the camera trajectory with the objects in the scene. Finally, we directly utilize all generated RGB frames and the corresponding metric scale camera poses to reconstruct the 3D scene in the absolute world coordinate frame \fcolorbox{black}{light_red}{\rule{0pt}{2pt}\rule{2pt}{0pt}}.
    }
    \label{fig:overview}
\vspace{-10pt}
\end{figure*}

Given a textual description of an indoor scene, our method follows a coarse-to-fine approach to generate a diverse and globally consistent 3D representation.

We first build a global proxy layout that provides semantic and structural guidance using a text-to-layout model (Sec.~\ref{subsec:proxygeneration}), which establishes the absolute world frame and guides subsequent stages. 
Next, we synthesize a 360\degree cubemap panorama using a novel multiview text-to-image diffusion model, augmented by two conditional adapters, one for semantics and one for depth (Sec.~\ref{subsec:panogen}), enforcing consistency and adherence to the original layout.
To reveal regions not covered by the panorama, we generate coverage-aware camera trajectories from the proxy depth, render novel views with a camera-guided video diffusion model for novel view synthesis (NVS) and align them to the global coordinate frame (Sec.~\ref{subsec:traj_est_sampling}).
As last step, we reconstruct the scene with 3D Gaussian Splatting (3DGS) using the aligned RGB information, without any re-scaling or external pose estimation, as all RGB frames are already aligned to our 3D global scene layout (Sec.~\ref{subsec:3dgs_recon}).

\subsection{Text-to-3D Global Scene Layout Estimation}
\label{subsec:proxygeneration}
We leverage Holodeck~\cite{Yang_2024_CVPR_holodeck} to generate the 3D proxy environment, which generates plausible and diverse 3D room descriptions from a given input prompt. The resulting scene layout encodes both semantics and geometry, from which we extract proxy geometry together with depth, semantic and object instance renderings. Additionally, object-specific attributes including 9D (position, orientation and scale) and semantic information are also available. 
An example layout visualized as a mesh is illustrated in Figure~\ref{fig:overview} \fcolorbox{black}{light_purple}{\rule{0pt}{2pt}\rule{2pt}{0pt}}. 

From this 3D proxy, we render metric depth and semantic maps arranged as cubemaps, where six perspective images as faces of a cube represent a 360°$\times$180° panoramic image. For the semantic maps, we use the NYUv2-40 labels~\cite{silberman2012indoor}, which provide a standardized taxonomy for indoor scenes. These serve as the geometric and semantic guidance throughout our pipeline.
\subsection{Guided Panoramic Image Generation}
\label{subsec:panogen}

Our panoramic image generator is implemented as a cubemap-based multi-view latent diffusion model, similar to~\cite{kalischek2025cubediff}. The representation of a panoramic image as multiple perspective views ensures a natural distribution of pixels across the entire field of view, minimizing distortion compared to standard spherical representation of 360\textdegree\ panoramic images. 
We employ Stable Diffusion 2~\cite{rombach2022high} as our backbone and extend the architecture from single to multi-view generation. This is done by integrating inflated self- and cross-attention layers~\cite{kalischek2025cubediff,shi2023MVDream} in the U-Net blocks of the diffusion model. Inflated attention is implemented by reshaping the token sequence length from $b \times (hw) \times l$  to $b \times (thw) \times l$, where $b$ is the batch size, $hw$ is the flattened spatial size, $t=6$ is the number of views, and $l$ is the sequence length. This operation enables us to re-use initial pre-trained weights of the image diffusion model, and enables the information sharing between different views of the cubemap. 

Our generator receives two additional input signals: First, same as in~\cite{kalischek2025cubediff}, we utilize a binary mask to enable the usage of a reference input image for generation. Second, we incorporate a xyz-positional encoding~\cite{huang2025dreamcube} for each pixel in each face of the cubemap, which allows the multi-view generator to learn the spatial relation between the different faces of the cubemap and generate consistent geometry and object relations across views.

After training our panoramic image generator to produce realistic and diverse cubemaps of indoor scenes, we extend the architecture with a Multi-ControlNet~\cite{zhang2023controlnet} to enable the generation of panoramas which are aligned with the 3D proxy.
More precisely, we add two separate ControlNet branches for depth and semantic information, where each one is initialized with a copy of the U-Net model from our trained panoramic image generator. This means that our inflated attention layers as well as the xyz-positional encoding are also integrated in each of the ControlNet branches. During finetuning the ControlNets, we freeze the down and mid blocks, where the up blocks are replaced by zero convolutions and features are fused into the main U-Net branch, as proposed in~\cite{zhang2023controlnet}.

\subsection{Optimized Camera Trajectory Estimation and Sampling}
\label{subsec:traj_est_sampling}
After guided panoramic image generation, we can re-project the RGB and depth information to form an initial point cloud of the scene. As shown in Figure~\ref{fig:overview} \fcolorbox{black}{light_blue}{\rule{0pt}{2pt}\rule{2pt}{0pt}}, this point cloud lacks completeness due to multiple unobserved areas. 
To explore and generate content for these regions we utilize a camera-controlled video diffusion model for NVS~\cite{zhou2025stable}, but this requires careful consideration: 
First, exhaustive generation of a large number of views leads to high computational cost, making the overall 3D scene generation process slow~\cite{schneider_hoellein_2025_worldexplorer}. Second, and arguably more crucial, the higher the number of generated frames for unobserved areas, the more complex it is to ensure that the visible content is 3D consistent, as each new frame can potentially add inconsistencies. This can significantly reduce the overall quality of the final reconstruction. To solve these issues, we propose an efficient video trajectory estimation and sampling approach. In the first step, we estimate camera trajectories by taking into account collision avoidance and scene coverage. Second, we align these camera trajectories so that the frames of the generated videos align well with our global scene layout. This ensures that the novel frames cover unobserved regions of the scene as desired.

\paragraph{Collision-Aware Trajectory Estimation.}
\label{subsubsec:opt_traj_est}
To construct camera paths that avoid collisions while covering unobserved content, we follow a grid-based exploration strategy. We first divide the scene into equally spaced areas. More precisely, for standard-sized indoor scenes, we partition the space into four quadrants. For each quadrant, we define a sparse camera trajectory in a circular pattern, where each camera pose is oriented toward the center of the quadrant. The extent of the pattern is chosen by utilizing the metric 3D proxy information, so that the camera stays within the quadrant, and to avoid collisions with walls and ceiling. Additionally, we check if objects are in close proximity of the sparse camera trajectory, by rendering depth maps from the 3D proxy and discarding camera poses where the rendered depth is $< 0.3m$. Empirically, it showed that this threshold avoids sampling degenerate frames, which happens when the camera is too close to the layout or objects (see Figure~\ref{fig:ablation_trajectory} right, $scale=2.0$). 
As last step, we interpolate the remaining poses to obtain smooth trajectories that emphasize coverage of the unobserved regions. Note that this simple trajectory definition turned out to be sufficient in our experiments to avoid collisions with layout elements and objects, and to ensure high coverage of the room. However, for highly complex or uncommon scene layouts, the full information of our 3D proxy can be utilized to implement more general sampling strategies~\cite{viewselectionOLAGUE2002,viewselectionsun2021}. We refer to the supplementary material for a comparison of our proposed trajectory estimation algorithm to standard mesh-based view planning.

\paragraph{Depth-guided Trajectory Alignment.}
\label{subsubsec:traj_alignment}
Directly using the defined trajectories as input to a video diffusion model is non-trivial, since such models operate in their own relative coordinate systems. We therefore align our 3D proxy layout to the model's internal representation.
Specifically, we utilize Stable Virtual Camera~\cite{zhou2025stable} for NVS, which maps the scene into a cube of $[-2,2]^3$ units. Without re-normalization, the NVS camera trajectory can deviate from the target trajectory, leading to potentially unusable frames.

Zhou et al.~\cite{zhou2025stable} address this issue by introducing a specific camera-scale parameter. The optimal value for this parameter is estimated by sweeping values in $[0.1,2.0]$, and comparing the generated output against ground-truth. Since we are missing ground-truth information in our generative setting, we adapt and optimize their approach in two ways. First, we utilize our proxy layout as a ground-truth representative. For each scale candidate, we generate intermediate frames along the camera trajectory and predict the corresponding depth using a metric depth estimator~\cite{yang2024depth}, and optimize the scale by minimizing the L1 error against the proxy depth renderings. Formally, this can be defined as

\begin{equation}
    \theta_{cs}^* = \argmin_{\theta_{cs}}(\mathcal{L}_{dpt}),
\end{equation}
where $\theta_{cs}^*$ is the optimal camera scale parameter and $\mathcal{L}_{dpt}$ the depth alignment loss. $\mathcal{L}_{dpt}$ is defined as

\begin{equation}
\mathcal{L}_{dpt} = \frac{1}{N} \sum_{i=1}^{N} ||D^i_{gt} - D^i_{\theta_{cs}} ||_1, 
\end{equation}
where $D^i_{\theta_{cs}}$ is the metric depth estimated for the generated novel views for a specific camera scale parameter $\theta_{cs}$, and $D^i_{gt}$ is the metric depth rendered from the 3D proxy for the same camera pose $i$. Our second adaptation compared to~\cite{zhou2025stable} is that we perform bisection search instead of linear search for the camera scale parameter, which increases the computational efficiency. We refer to the supplemental for a detailed description of our search algorithm. 

This procedure yields NVS frames that are consistently aligned to the proxy's global coordinate system, and the camera poses of novel views accurately fit to the desired target poses. 

\subsection{3D Scene Reconstruction using 3DGS}
\label{subsec:3dgs_recon}

The final stage of our method involves generating a 3D reconstruction of the scene from all synthesized frames, as illustrated in Figure~\ref{fig:overview} \fcolorbox{black}{light_red}{\rule{0pt}{2pt}\rule{2pt}{0pt}}. We adopt 3D Gaussian Splatting (3DGS)~\cite{kerbl20233d} as the underlying scene representation due to its ability to provide high-quality, real-time rendering. 

Due to the precise alignment of the generated novel views with the global scene layout, our approach allows for direct use of depth renderings to estimate the initial point cloud. Moreover, all camera poses and intrinsics used to define the camera trajectories can be employed as ground-truth parameters for reconstruction. This provides a substantial advantage over prior methods~\cite{dominici2025dreamanywhere,yang2024layerpano3d,schneider_hoellein_2025_worldexplorer,zhou2024dreamscene360}, which depend on estimated depth or camera poses, and are therefore prone to error accumulation potentially degrading the accuracy of the final 3D reconstruction. Our training loss $\mathcal{L}_{\text{}}$ is defined as
\begin{equation}
\label{eq:our_full_loss_main}
\mathcal{L}_{\text{}} =\mathcal{L}_{\text{3DGS}} +  \mathcal{L}_{\text{geom}} +\mathcal{L}_{\text{NN}} + \mathcal{L}_{\text{depth}},
\end{equation}
where $\mathcal{L}_{\text{3DGS}}$ is the original 3DGS loss proposed in~\cite{kerbl20233d}, and $\mathcal{L}_{\text{geom}}$ describes the normal, geometric, and photometric multi-view loss introduced in~\cite{chen2024pgsr}. $\mathcal{L}_{\text{NN}}$ minimizes the k-nearest-neighbor distances between the Gaussian means and the initial point cloud, promoting structural coherence in the reconstructed geometry. $\mathcal{L}_{\text{depth}}$ enables depth-guided supervision using our metric ground-truth depth maps, which ensures improved depth fidelity. A detailed description of the individual loss terms, training procedure, optimization details, and parameter settings is provided in the supplementary material. Note that for 3DGS reconstruction, we use all NVS generated frames from our four trajectories with 42 frames each, and additionally sample 170 frames from the panoramic image, leading to a total of 338 frames.
\section{Experiments and Evaluation}
\label{sec:experiments}
To evaluate the full capabilities of our model, we split the evaluation into two parts: First, we compare our guided panoramic image generator against strong text-to-panorama baselines to quantify the overall image quality and prompt alignment. Second, we evaluate the overall quality and consistency of generated 3D scenes using novel-view renderings from camera poses that deviate strongly from the reconstruction views, complemented by qualitative comparisons and a user study.
For both parts, we use diverse indoor scenes from the Structured3D~\cite{Structured3D} test set. We select a set of scenes with varying layout complexity and scene type, and infer textual descriptions using the publicly available Janus Pro model~\cite{chen2025janus}. These textual descriptions are then used as input for both evaluation experiments.

\subsection{Training, and Inference Setup for Panorama Generation}
As mentioned in Subsection~\ref{subsec:panogen}, the training of our panorama generator consists of two consecutive steps. First, we train our cubemap-based multi-view diffusion model for 120k iterations with batch size 2 using AdamW ($\beta_1=0.9, \beta_2=0.999, \gamma=1e^{-2}$), and a learning rate of 1e-5 linearly ramped up over 10k steps. Gradient clipping with a maximum norm of 1 is applied. The model is trained with v-prediction using a DDPM noise scheduler with 1000 timesteps. As training data, we utilize various public panoramic image datasets~\cite{Structured3D,Matterport3D,gardner2017laval_indoor}.

In the second step, we extend the multi-view diffusion model with two ControlNet branches for depth and semantic guidance, each initialized with the previously trained U-Net, including all architectural modifications. We then finetune the full model for 90k iterations with batch size 1 on the Structured3D~\cite{Structured3D} dataset, using the same optimizer settings as for the multi-view diffusion model.

At test time, the model takes as input a text prompt along with depth and semantic renderings from the 3D proxy scene. We use DDIM sampling with 50 steps and apply classifier-free guidance of 8.0.

\subsection{Evaluation of Panorama Generation}
In this section, we provide the results of our quantitative evaluation for panoramic image generation using text descriptions from the Structured3D dataset.
\paragraph{Evaluation Metrics.}
To measure image quality and aesthetics, we use CLIP-IQA+~\cite{clipiqa}, Q-Align~\cite{qalign}, and the Inception Score (IS)~\cite{salimans2016improved} as reference-free metrics. In addition, we compute the CLIP score~\cite{clipscore} between the input textual description and the generated image to quantify prompt alignment.
\paragraph{Baselines.}
Diffusion360~\cite{feng2023diffusion360} introduces a circular blending strategy within diffusion models to generate seamless 360° panoramic images by ensuring geometric continuity across the panorama’s edges during denoising and decoding. MVDiffusion~\cite{tang2023MVDiffusion} presents a correspondence-aware diffusion approach that simultaneously generates multiple views of a scene with cross-view interactions to enforce consistent multi-view image synthesis without iterative warping. The panoramic image generator of LayerPano3D~\cite{yang2024layerpano3d} is a finetuned version of a Flux~\cite{flux2024} text-to-image model with a panorama LoRA~\cite{hu2022lora} for panorama generation and inpainting. 
\paragraph{Results.} As shown in Table~\ref{tab:panoeval}, our panoramic image generator performs competitively compared to recent baselines in terms of both image quality and aesthetics metrics. Additionally, our superior CLIP score indicates high alignment between the textual description and the generated panoramic image.
 We provide a qualitative comparison to the competitor methods in the supplemental, as well as visualizations to highlight that our Multi-ControlNet helps to avoid geometric and semantic inaccuracies, effectively increasing the plausibility and visual quality of the results.

\begin{table}[]
\centering
\scalebox{0.9}{
\setlength{\tabcolsep}{4pt}
\begin{tabular}{lcccc}
\toprule
Method & CLIP Score $\uparrow$ & Q-Align $\uparrow$ & CLIP-IQA+ $\uparrow$ & IS $\uparrow$ \\
\midrule
Diffusion360~\cite{feng2023diffusion360} & 30.73 & \textbf{4.80} & \textbf{0.77} & 3.10 \\
MVDiffusion~\cite{tang2023MVDiffusion} & 31.28 & 4.10 & 0.63 & \textbf{3.46} \\
LayerPano3D~\cite{yang2024layerpano3d} & 30.95 & 4.38 & 0.64 & 3.24 \\
\midrule
Ours (full, text + Multi-ControlNet) & \textbf{32.48} & \underline{4.54} & \underline{0.71} & \underline{3.26} \\
\midrule
\midrule
\textbf{Ablation} &&&&\\
Ours (text only) & 31.17 & 4.45 & 0.69 & 3.33 \\
Ours (text + Depth-ControlNet) & 31.85 & 4.49 & 0.70 & 3.33 \\
Ours (text + Semantic-ControlNet) & 32.23 & 4.49 & 0.70 & 3.41 \\
\bottomrule
\end{tabular}
}
\vspace{3pt}
\caption{\textbf{Quantitative comparison of panoramic image generators.} Our guided panorama generator achieves the highest CLIP score, highlighting the high alignment between input text prompt and generated panorama. The competitive results for the other metrics show the overall high quality of our generated panoramic images.}
\label{tab:panoeval}
\end{table}

\subsection{Evaluation of NVS and 3D Consistency of Generated Scenes}
To ensure a fair comparison for 3D consistency, we generate a common set of high-quality panorama images from the textual descriptions serving as input for all methods. This ensures that all scenes have the same level of complexity, and we can reliably compare the scene reconstruction accuracy across different methods. 
For the evaluation, we aim to generate comparable camera paths for each method. To achieve this, we move the camera in a circular trajectory within the layout boundaries, with the camera oriented toward the center of the scene, generating videos consisting of $480$ frames per scene.
\paragraph{Evaluation Metrics.} We provide the MEt3R score~\cite{asim24met3r}, a metric for multi-view consistency for a sequence of consecutive frames, and additionally use the Inception Score to assess the visual quality of rendered novel views.
Furthermore, we conduct a user study and ask the participants to assess the Perceptual Quality (PQ), 3D Consistency (3DC) and Overall Scene Quality (SQ) of the scene reconstruction results on a scale from 1-5, where higher means better. 
\paragraph{Baselines.} We compare against available state-of-the-art methods that turn a 360\degree panorama into a 3D scene representation. DreamScene360~\cite{zhou2024dreamscene360} initializes a set of 3D Gaussian splats from the panoramic image, then optimizes them by enforcing geometric and semantic correspondence between training and novel views. Novel views are obtained by perturbing the training views. LayerPano3D~\cite{yang2024layerpano3d} synthesizes novel content by erasing objects and inpainting the panoramic images, then estimating and fusing depth maps for each layered 3DGS representation. WorldExplorer~\cite{schneider_hoellein_2025_worldexplorer} autoregressively samples camera trajectories for novel view synthesis conditioned on the starting perspective views and closest frames in scene memory, which is built incrementally.

\begin{figure}
\centering
\scalebox{.95}{
\begin{tabular}{c}
\includegraphics[trim={0.0cm 13.6cm 0.0cm 19.2cm},clip, width=0.95\linewidth]{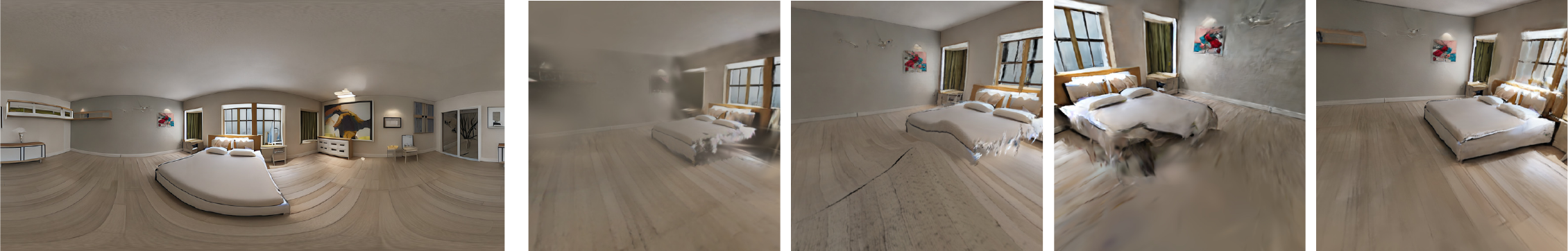} \\
\includegraphics[trim={0.0cm 26.35cm 0.0cm 6.4cm},clip, width=0.95\linewidth]{figures/visual_compare/fig_3_compressed.pdf} \\
\includegraphics[trim={0.0cm 0.0cm 0.0cm 32.1cm},clip, width=0.95\linewidth]{figures/visual_compare/fig_3_compressed.pdf} \\
\end{tabular}
}
\vspace{-.3cm}
\caption{\textbf{Visual comparison of 3D reconstruction results for different methods.} The 2D renderings show comparable viewpoints for different methods. Our method generates more convincing and complete geometries compared to other methods. We refer to the supplemental for additional visualizations.
}
\label{fig:qual_results}
\end{figure}

\begin{figure}[]
\centering
\includegraphics[width=.95\linewidth]{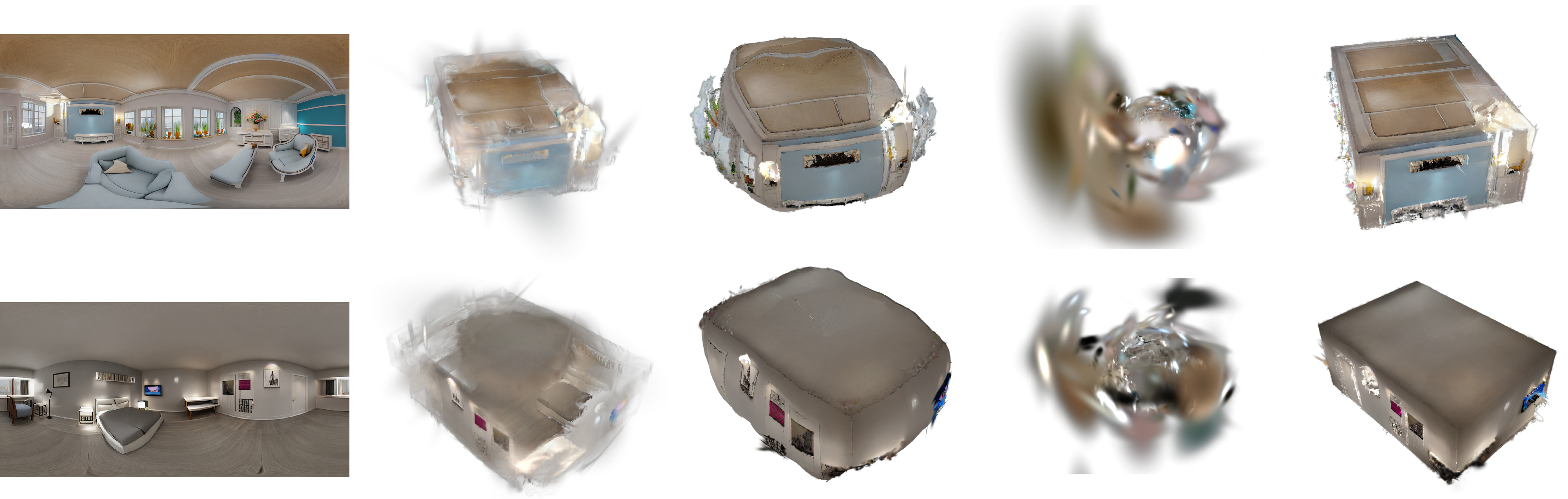}\\
\scriptsize{Initial Panorama \hspace{0.5cm} DreamScene360 \hspace{.5cm} LayerPano3D \hspace{.4cm} WorldExplorer \hspace{.5cm} Ours}
\vspace{-.1cm}
\caption{\textbf{Scene bounding comparison of 3D reconstruction results for different methods.} Our 3DGS point cloud follows the scene geometry accurately, while other methods have significant distortions and floaters. Note that 3D scenes of competitor methods are in relative scale and have been manually re-scaled for better visualization. In contrast, our method provides results in absolute metric scale.}
\label{fig:scene_bounding}
\vspace{-10pt}
\end{figure}

\paragraph{Results.} As shown in Table~\ref{tab:eval_3Drecon}, our method was overall preferred by both the image metrics and the user participants. We note that DreamScene360 performs generally well, as it tends to blur out the unobserved areas in the scene, which can be perceptually pleasant to the observer. LayerPano3D produces very sharp details by relying less on a full 3DGS optimization, but suffers from distortion due to the equirectangular projection and the flat geometry used to represent the objects. Finally, WorldExplorer optimizes a large number of incoherent views, causing instability and error accumulation, thus performed worst. 
On a single consumer graphics card, our runtime is comparable to DreamScene360 and LayerPano3D, while being $\sim$10$\times$ faster than WorldExplorer thanks to our optimized trajectory sampling, which significantly reduces the dominant NVS cost. 
Figures~\ref{fig:qual_results},\ref{fig:scene_bounding} show that our method generates high-fidelity novel views and 3D reconstructions that preserve the panorama’s appearance and the proxy structure under large viewpoint changes. Additional renderings and user study details are shown in the supplementary material.

\begin{table}[]
\centering

\setlength{\tabcolsep}{4pt}
\scalebox{0.9}{
\begin{tabular}{lccccccc}
\toprule
&\multicolumn{2}{c}{2D Metrics}&&\multicolumn{3}{c}{User Study} & Mean\\
\cmidrule{2-3} \cmidrule{5-7}
Method & IS $\uparrow$ & MEt3R $\downarrow$ && PQ $\uparrow$ & 3DC $\uparrow$ &  SQ $\uparrow$ &Runtime\\
\midrule
DreamScene360~\cite{zhou2024dreamscene360}                  & 1.91 & \underline{0.026} && \underline{2.34} & \underline{2.44} & \underline{2.38} & 0.5h\\
LayerPano3D~\cite{yang2024layerpano3d}                      & 2.00 & 0.031 && 2.08 & 1.79 & 1.94 & 0.2h\\
WorldExplorer~\cite{schneider_hoellein_2025_worldexplorer}  & \textbf{2.27} & 0.033 && 1.42 & 1.51 & 1.35& 7h \\

\midrule
GuidedSceneGen (ours)                                        & \underline{2.22} & \textbf{0.024} && \textbf{3.03} & \textbf{3.05} & \textbf{3.01}& 0.75h \\
\bottomrule
\end{tabular}
}
\vspace{3pt}
\caption{\textbf{Quantitative comparison for NVS and 3D consistency of generated scenes.} We report results for 2D metrics on novel views, a conducted user study, and average runtime per scene. The generated scenes from our method are preferred in the user study, which also reflects in competitive results for 2D metrics.}
\label{tab:eval_3Drecon}
\vspace{-25pt}
\end{table}

\subsection{Ablation Study}
    
\paragraph{Influence of Guidance Signals for Panorama Generation.}
Our guided panoramic image generator ensures the alignment between global 3D proxy and the generated 2D RGB frames, which enables optimized trajectory calculation, metric alignment of the reconstructed scene, and good adherence to the input prompt. 
The ablation in Table~\ref{tab:panoeval} shows the effect of using different guidance signals. \textit{Ours (text only)} already achieves competitive performance, since we re-purpose a strong pretrained text-to-image diffusion model with only minimal architectural changes. It shows that the design choice of representing the panoramic image as cubemap enables our model to retain the ability of the base image diffusion model to predict high quality perspective images. Adding the Multi-ControlNet (\textit{Ours (full, text + Multi-ControlNet)}) further boosts performance by injecting proxy-derived depth and semantic signals: depth reduces geometric inconsistencies, semantic maps improve semantic correctness and text-image alignment (highlighted by higher CLIP score). We refer to the supplemental for corresponding qualitative results highlighting the visual improvements when using the additional guidance signals. 

\paragraph{Depth-guided Trajectory Alignment.}
To evaluate the effect of our trajectory alignment step, we compare our trajectories using depth-guided alignment with trajectories using the default setting from~\cite{zhou2025stable}, where the camera scale parameters are $[1,2]$.
To measure the alignment, we utilize MapAnything~\cite{keetha2025mapanything} to estimate accurate camera poses for the generated RGB frames of the trajectories in metric coordinates, and compare them with the target trajectory derived from the 3D proxy.

\begin{figure}[]
\centering
\scriptsize
\scalebox{.95}{
\begin{tabular}{cccc}
  \includegraphics[width=0.24\linewidth]{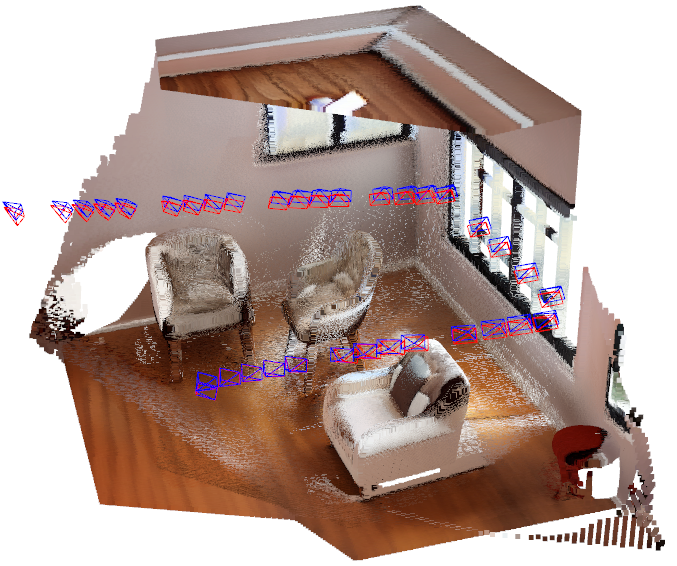} &
    \includegraphics[width=0.24\linewidth]{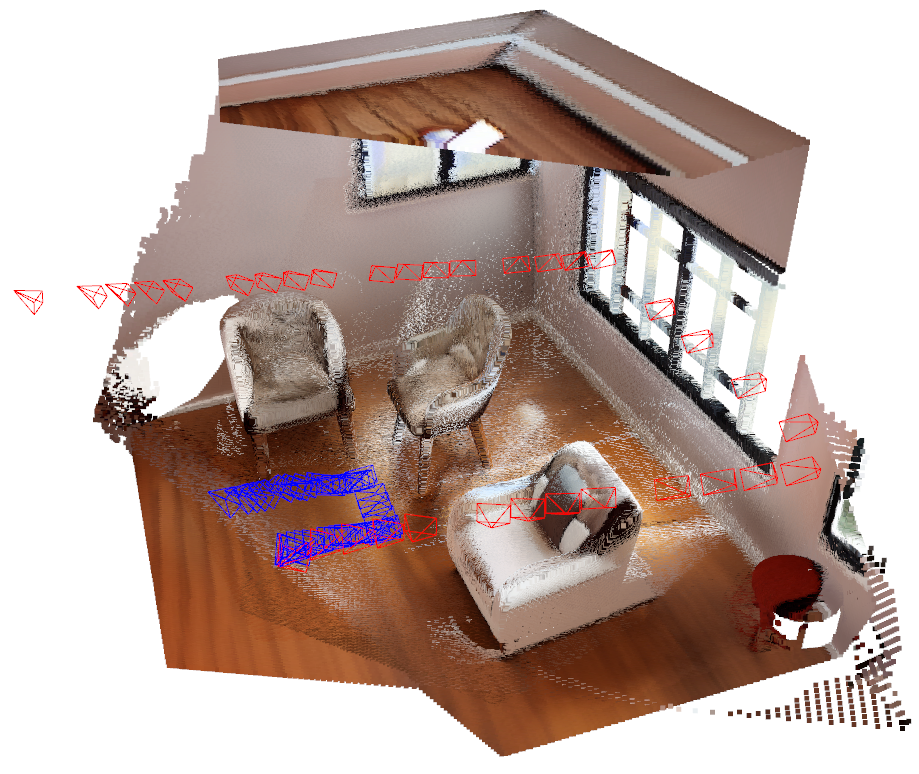} &
  \includegraphics[width=0.24\linewidth]{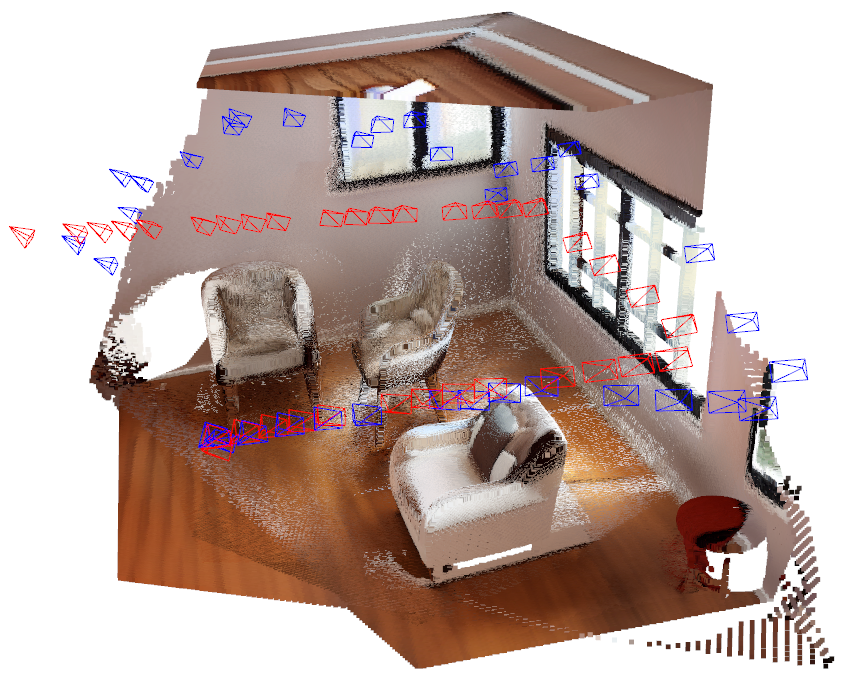}&
    \includegraphics[trim={0.0cm 2.5cm 11.0cm 0.0cm},clip,width=.25\linewidth]{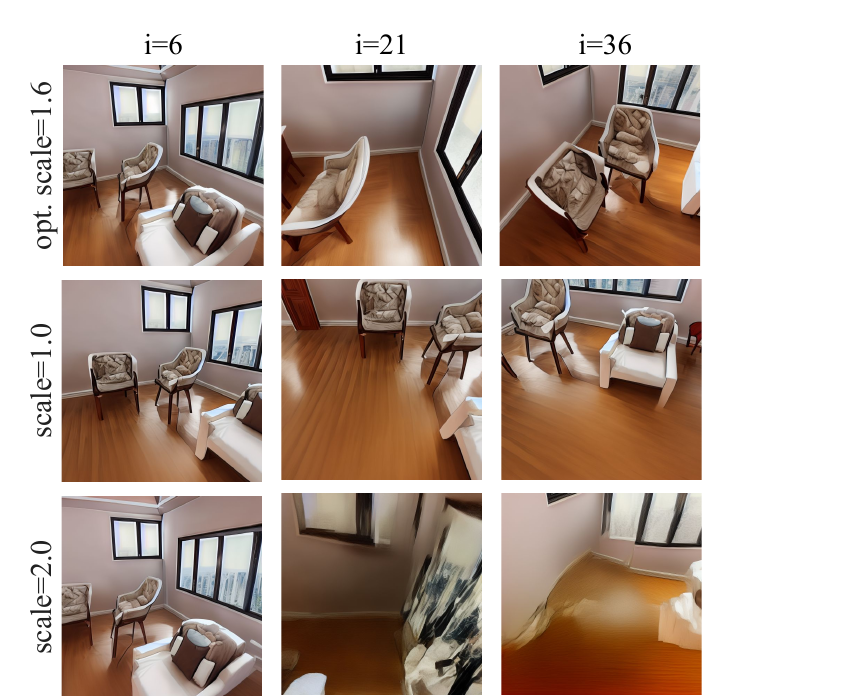}
\\
 Target poses (red) vs          & Target poses (red) vs           & Target poses (red) vs & Generated novel views for \\
 our opt. scale=1.6 (blue) &       default scale=1.0 (blue)              &default scale=2.0 (blue)& varying scales at camera pose $i$ \\

\end{tabular}
}
\caption{\textbf{Visualization of different NVS trajectories.} \textbf{Left:} We utilize MapAnything~\cite{keetha2025mapanything} and compare the predicted camera poses for different scale parameters. Using our strategy, the predicted camera poses are closely aligned with the ground truth. \textbf{Right:} Our optimal scale parameter ensures that the estimated camera trajectories avoid collisions with walls and objects, leading to high quality novel views. In contrast, using default scale parameters can lead to degenerate frames or low scene coverage.}
\label{fig:ablation_trajectory}
\end{figure}

Figure~\ref{fig:ablation_trajectory} shows the visualization of the camera poses from the target trajectory compared to the predicted poses of our optimized trajectory, and two trajectories with the default camera scale parameter, respectively. We observe that the predicted camera poses from our optimized trajectory align well with the target one, whereas the default trajectories can significantly differ. This alignment ensures the accurate coverage of unobserved areas in the scene, and directly enables the usage of the camera poses as ground truth for the generated novel views during the 3DGS optimization.
Furthermore, our generated trajectories avoid collisions with layout elements and objects. This is important, as camera poses too close to objects can result in degenerate frames, as shown in Figure~\ref{fig:ablation_trajectory} (right, see $scale=2.0$).

\begin{figure}[t]
\centering
\scalebox{0.9}{
\begin{tabular}{ccccccc}
\includegraphics[width=0.14\linewidth]{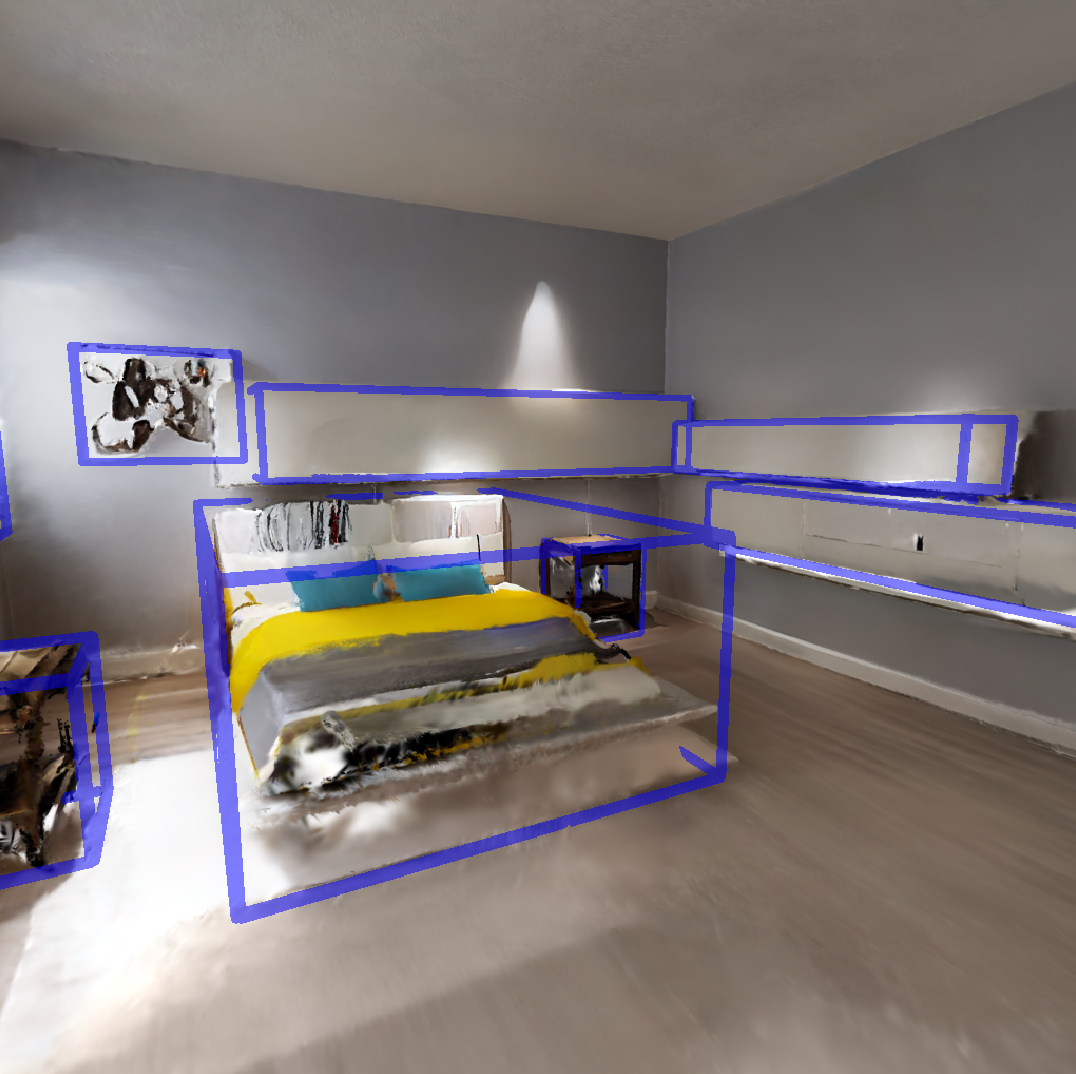}  &
 \includegraphics[width=0.14\linewidth]{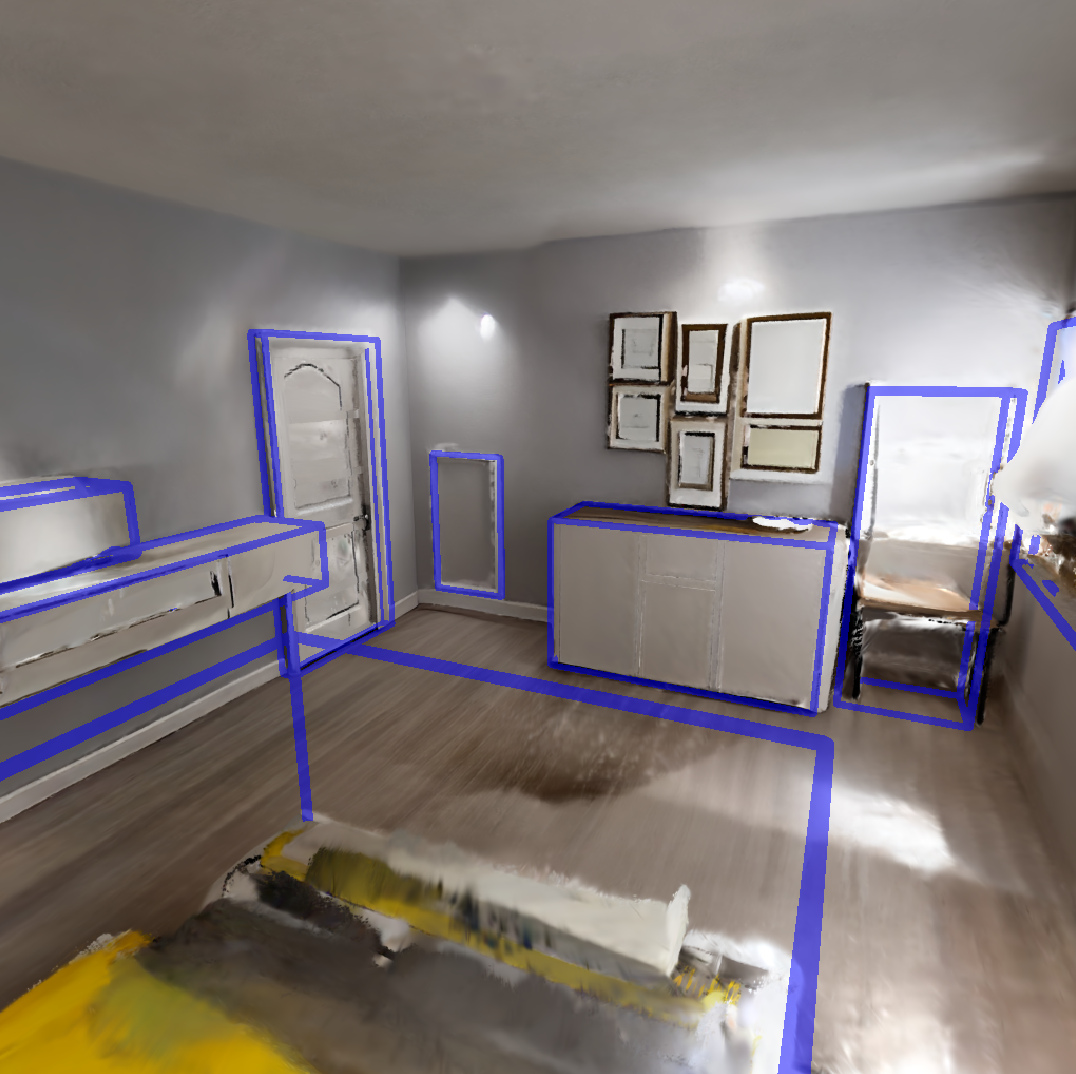} &
 \includegraphics[width=0.14\linewidth]{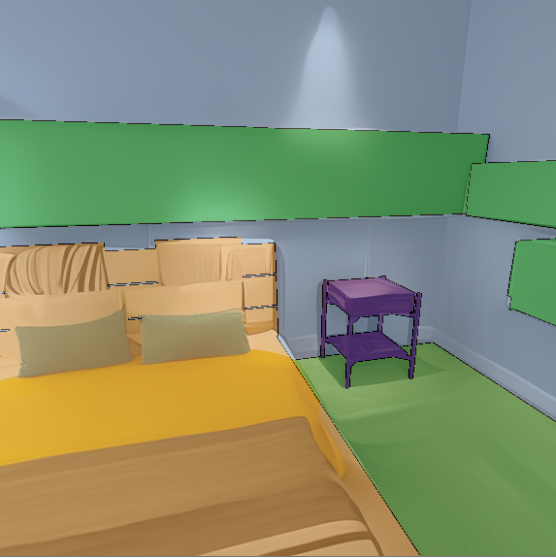} &
 \includegraphics[width=0.14\linewidth]{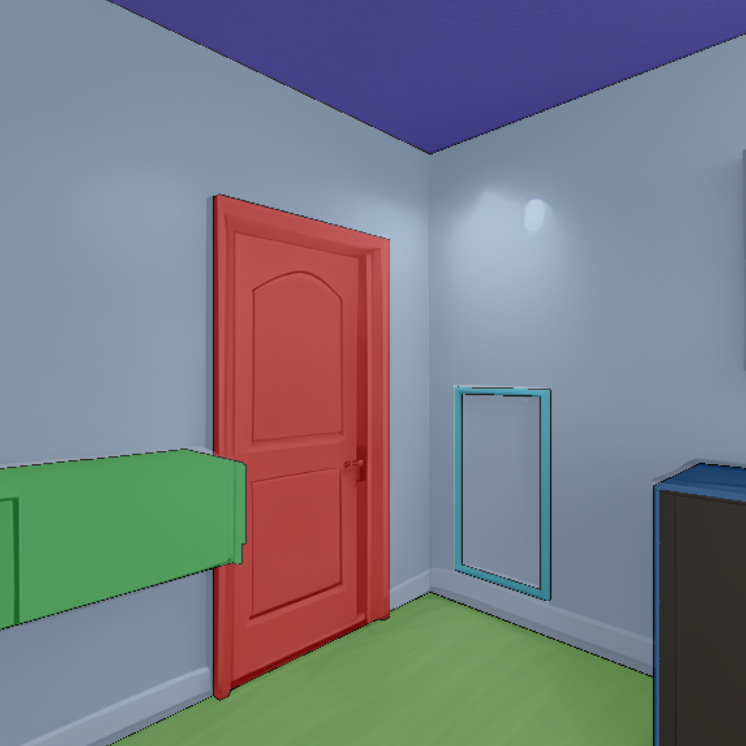} &
 \includegraphics[trim={1cm 0cm 2.8cm 3.cm},clip,width=0.14\linewidth]{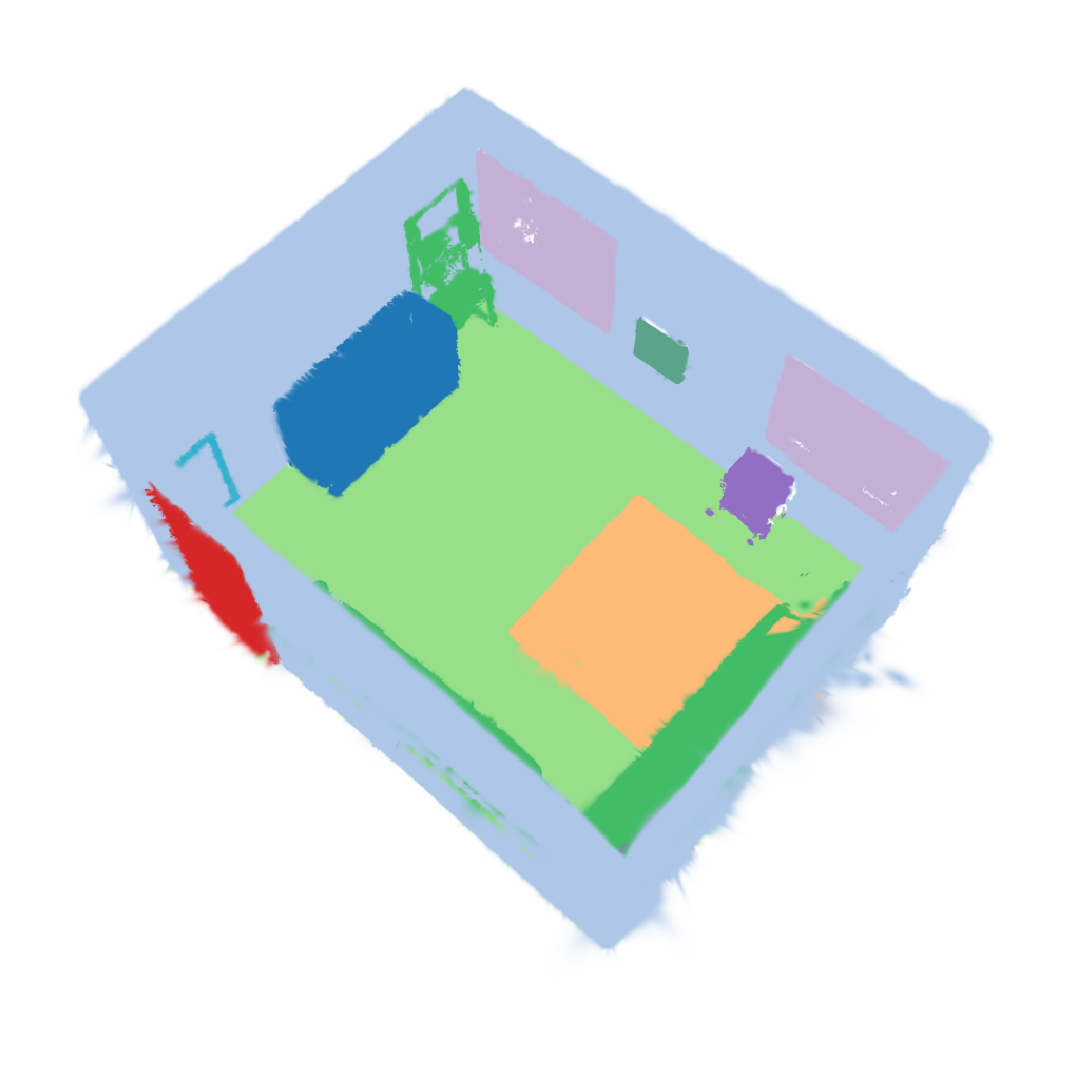}
  &
 \includegraphics[width=0.14\linewidth]{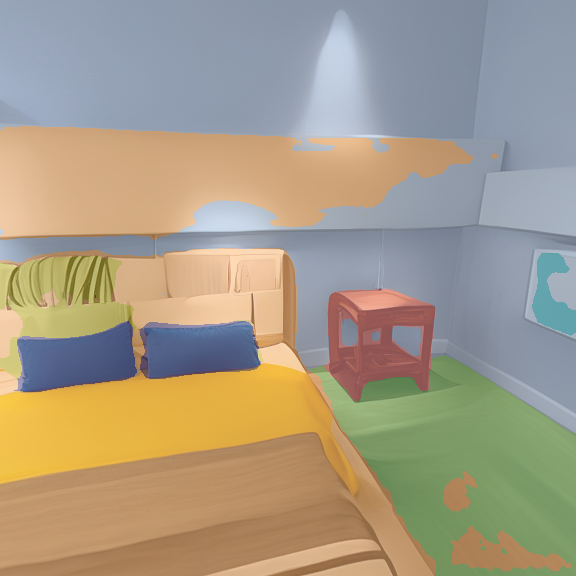} &
 \includegraphics[width=0.14\linewidth]{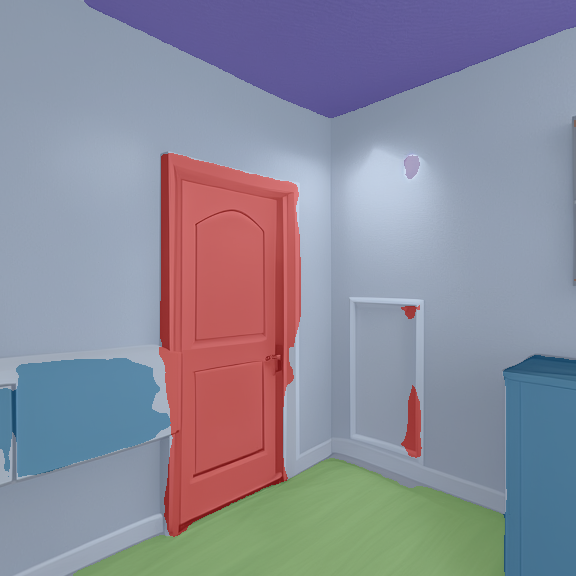}
 \\

\includegraphics[width=0.14\linewidth]{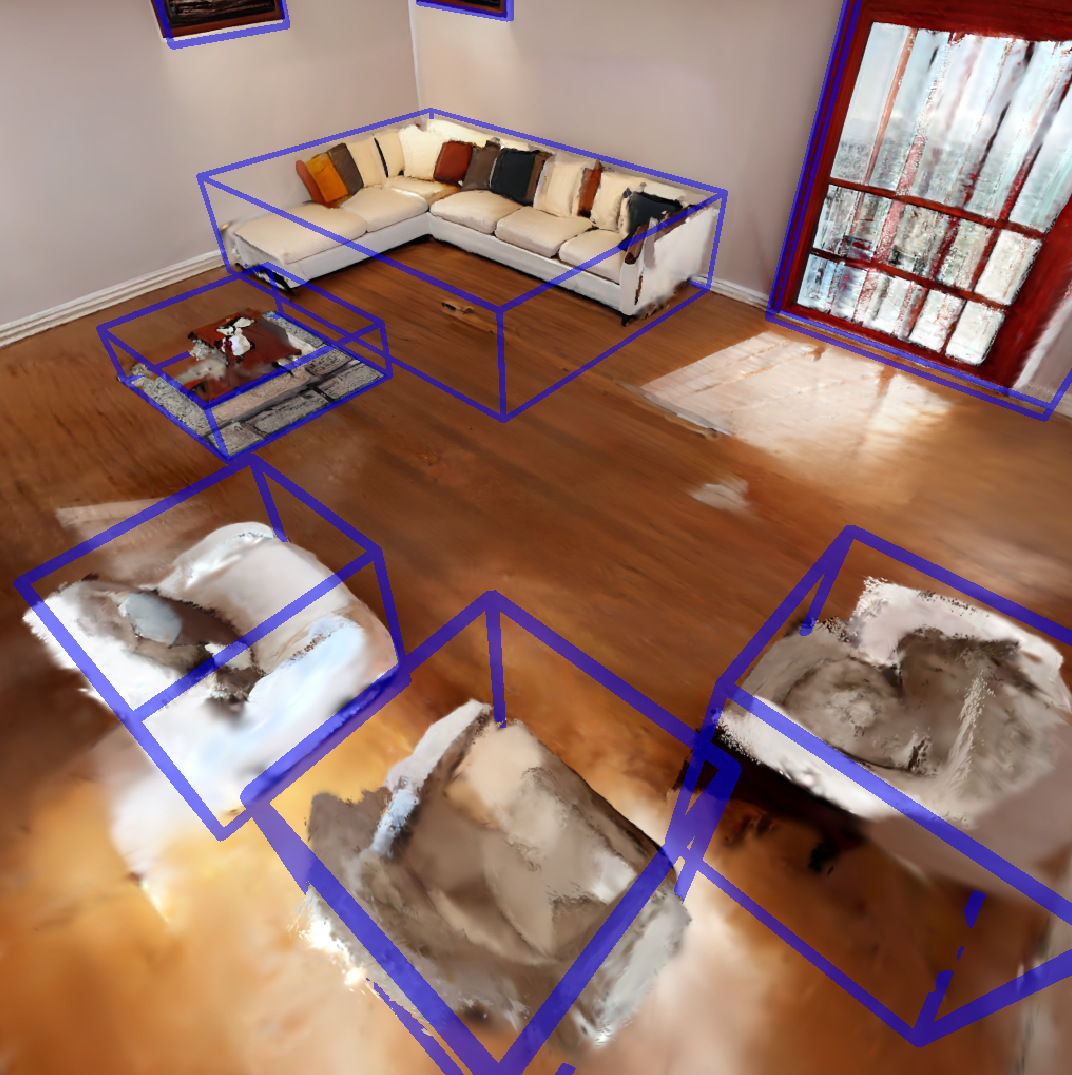}  &
\includegraphics[width=0.14\linewidth]{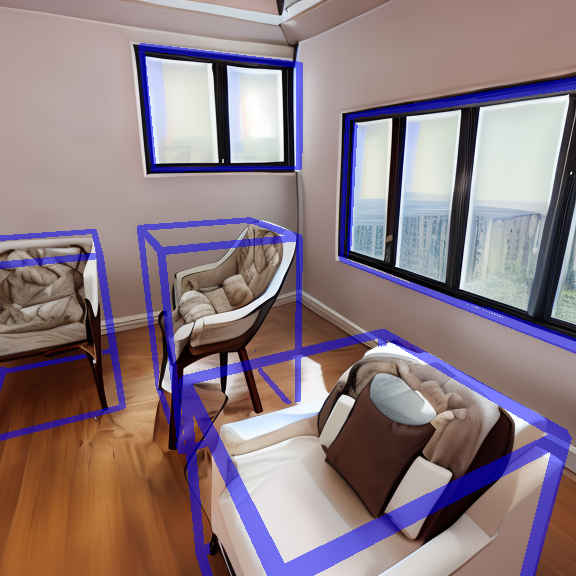} &
\includegraphics[width=0.14\linewidth]{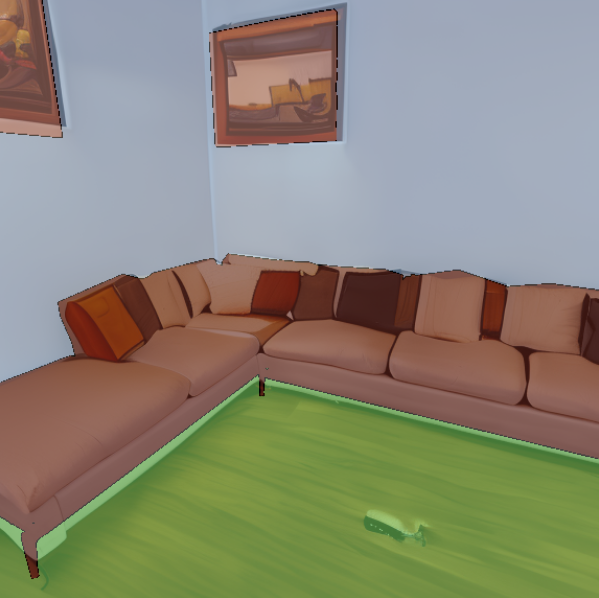} &
\includegraphics[width=0.14\linewidth]{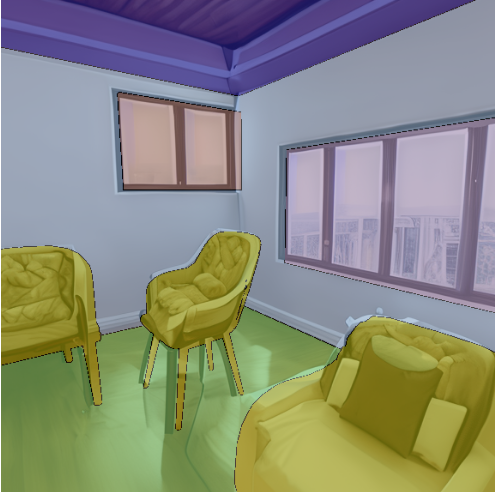} &

\includegraphics[trim={1cm 0cm 3cm 6.5cm},clip,width=0.14\linewidth]{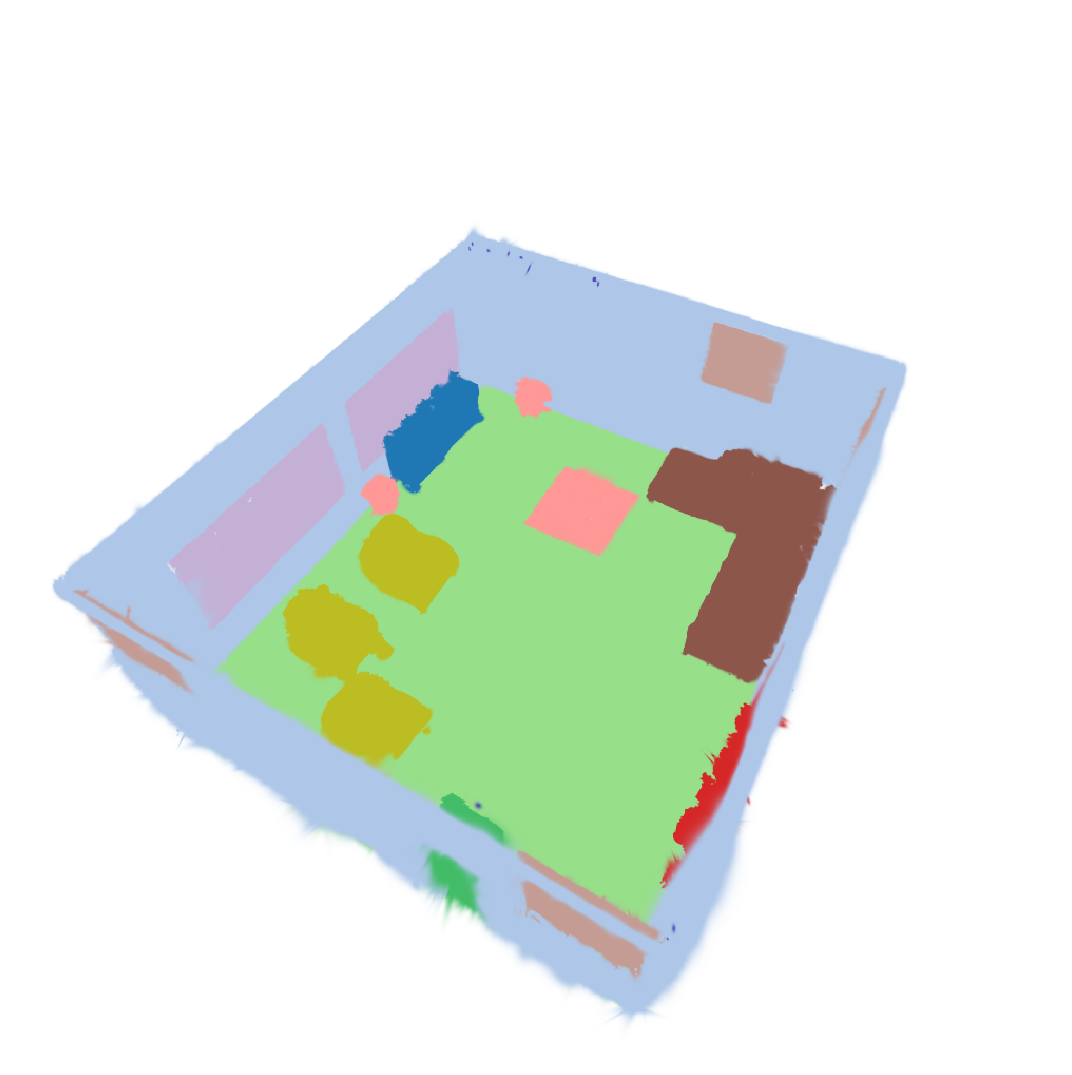}
  &
 \includegraphics[width=0.14\linewidth]{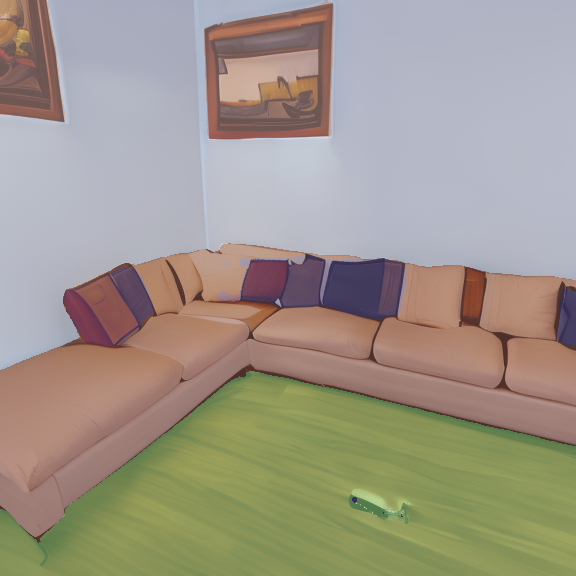} &
 \includegraphics[width=0.14\linewidth]{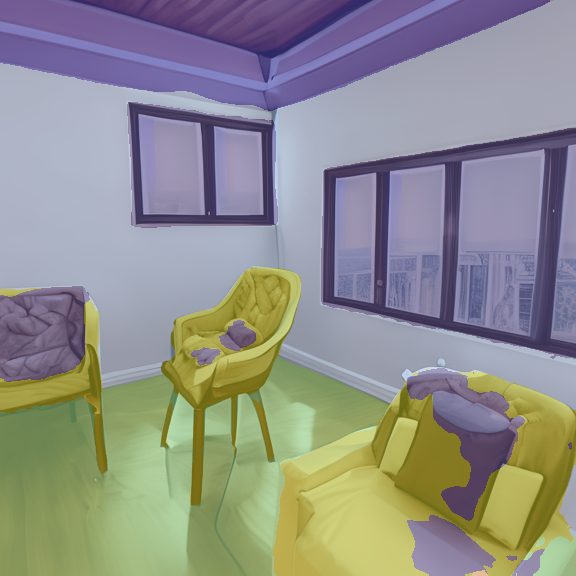}\\

\includegraphics[width=0.14\linewidth]{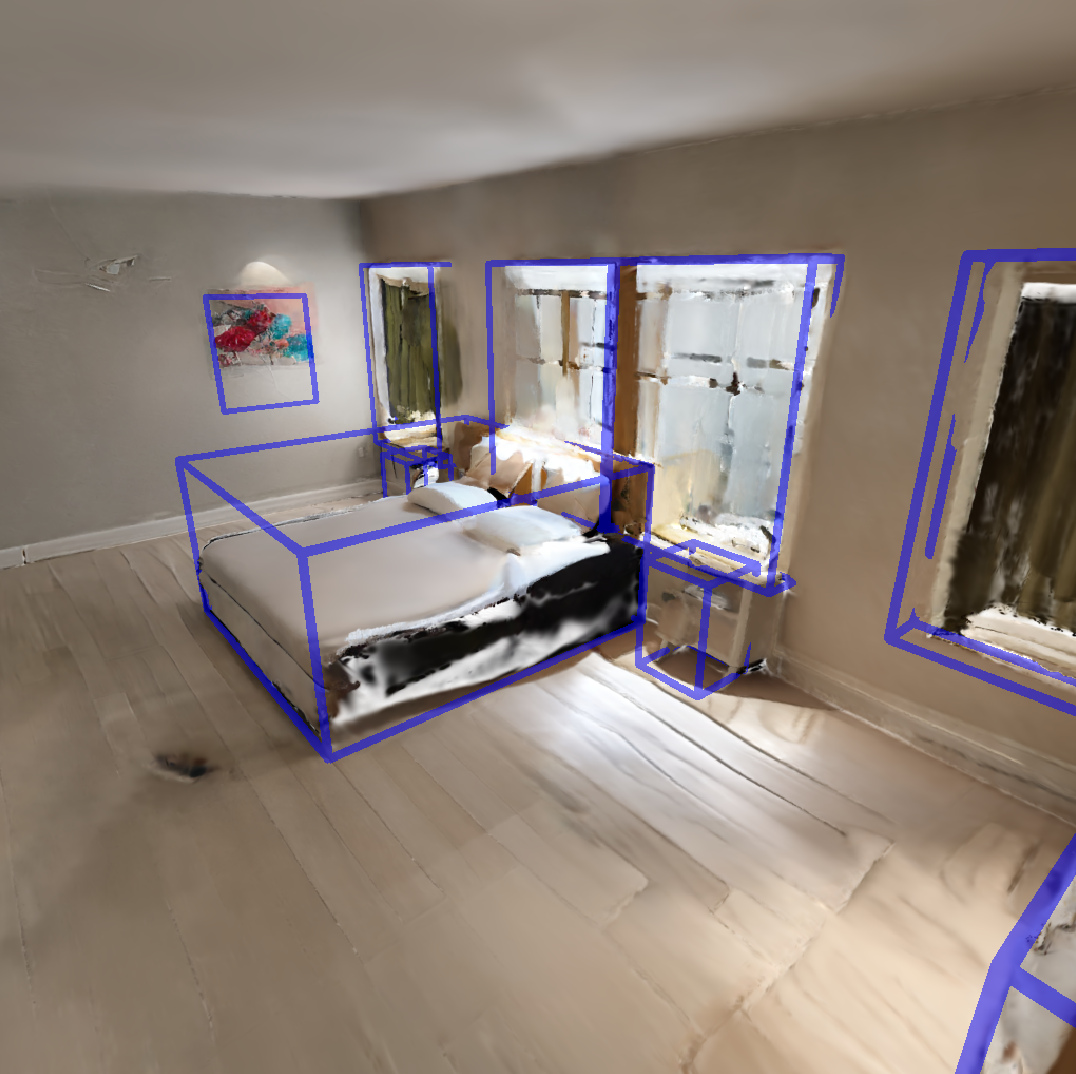}  &
\includegraphics[width=0.14\linewidth]{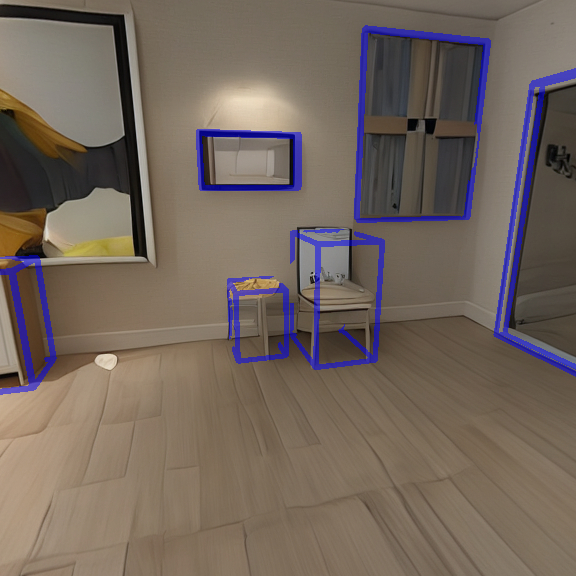} &
\includegraphics[width=0.14\linewidth]{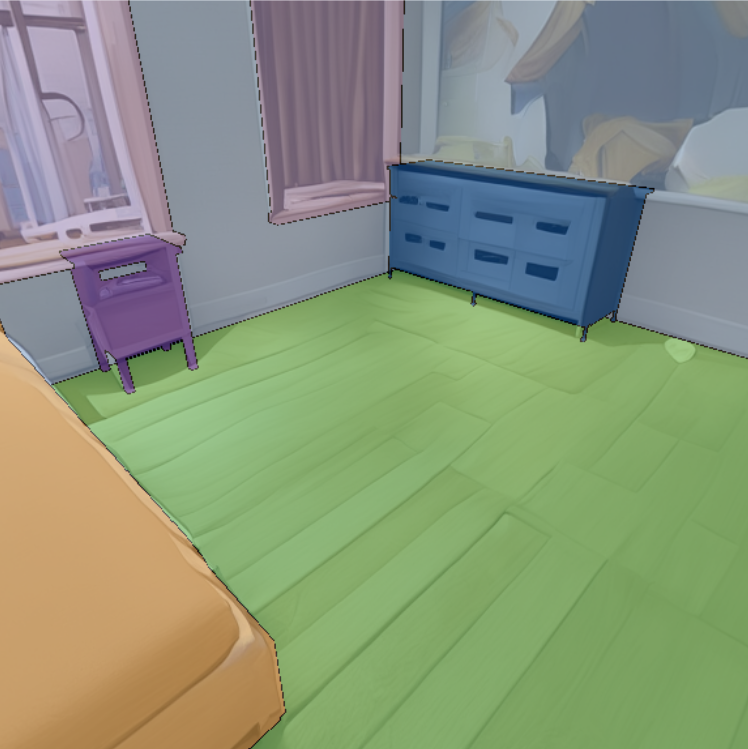} &
\includegraphics[width=0.14\linewidth]{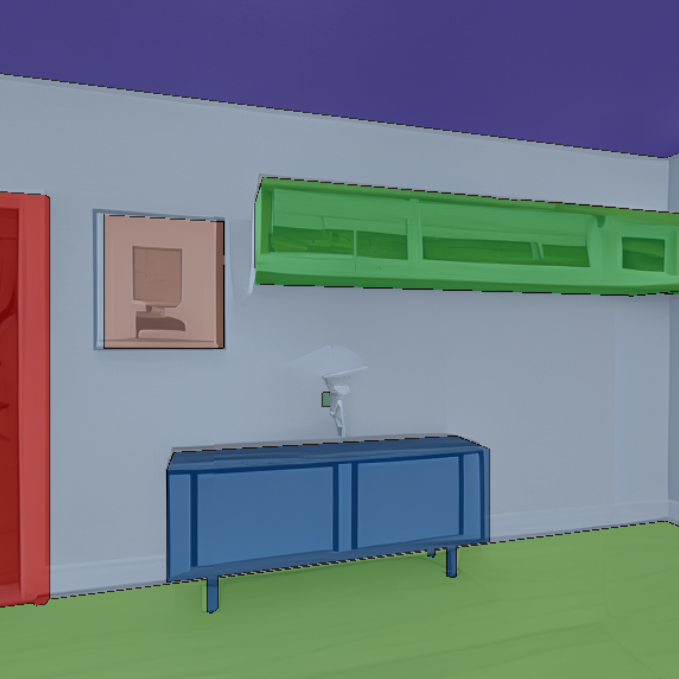} &
\includegraphics[trim={3cm 1.5cm 2.cm 3.5cm},clip,width=0.14\linewidth]{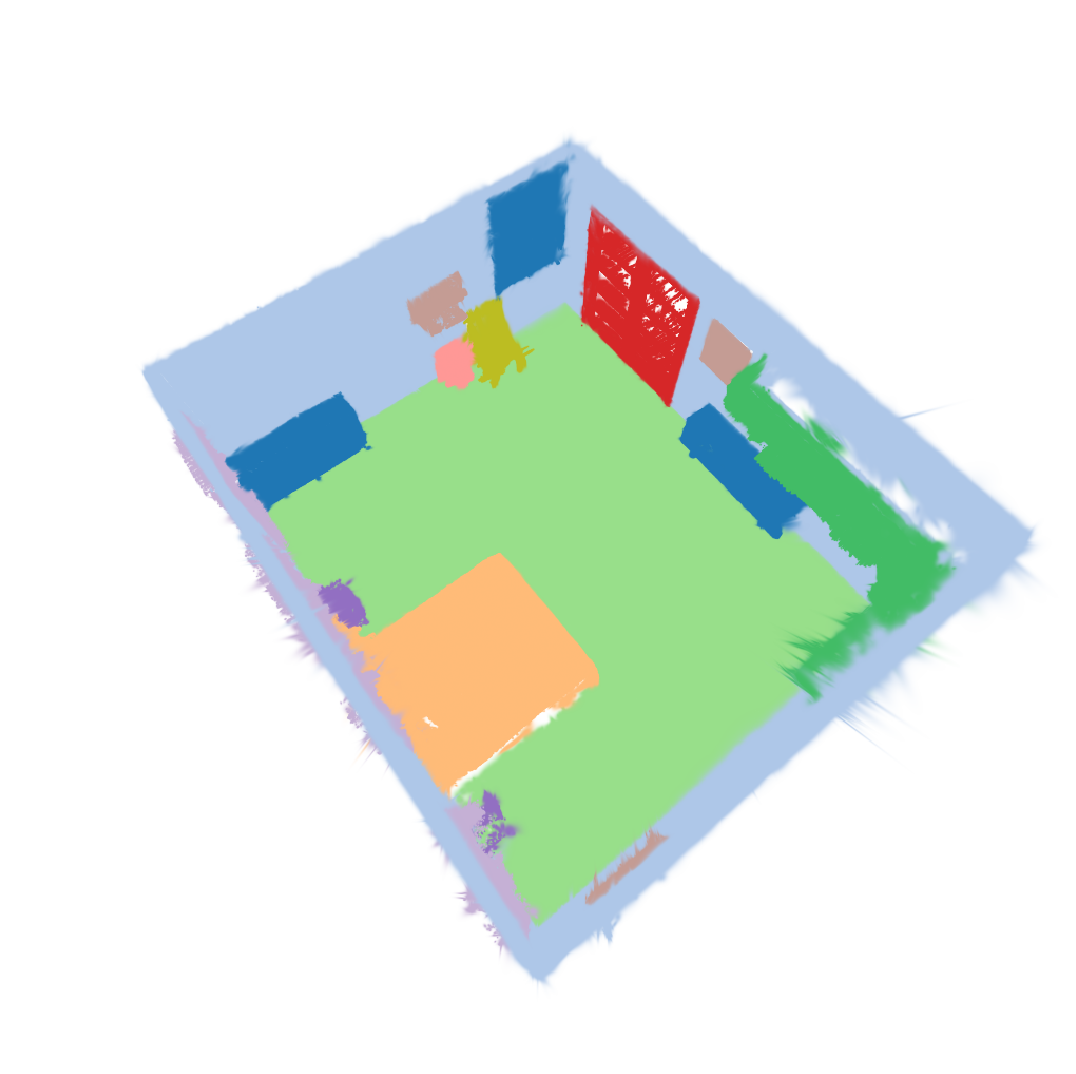}
  &
 \includegraphics[width=0.14\linewidth]{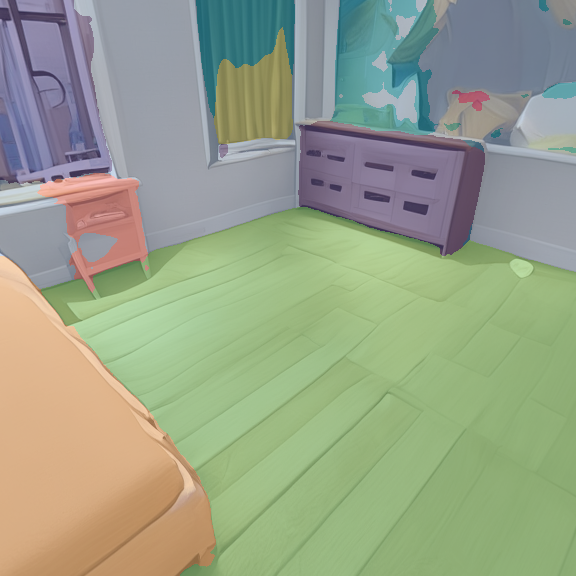} &
 \includegraphics[width=0.14\linewidth]{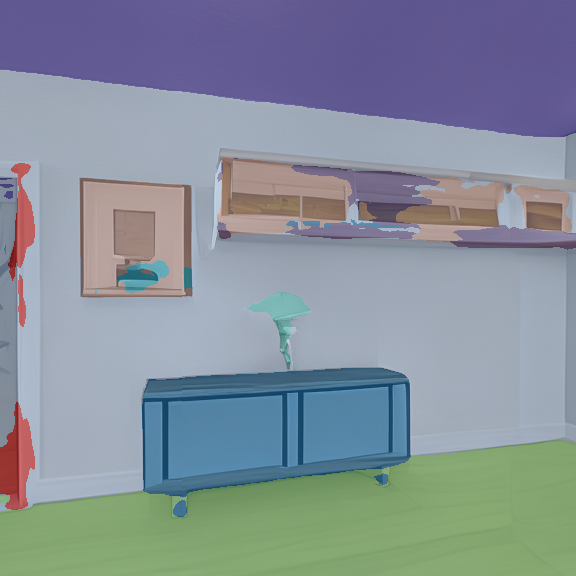}\\
\multicolumn{2}{c}{9D object poses (ours)}&\multicolumn{2}{c}{2D seg. (ours)} & 3D seg. (ours) & \multicolumn{2}{c}{2D seg. using~\cite{xie2021segformer}}\\
\end{tabular}
}

\caption{\textbf{Visualization of our annotations.} 9D object poses and 2D semantic masks transferred from the 3D proxy align well with novel views of the final 3DGS scene and enable reliable 3D Gaussian segmentation via clustering. Compared to predictions of SegFormer~\cite{xie2021segformer}, a specialist model trained for 2D indoor scene segmentation, our transferred annotations are more precise.}
\label{fig:annotations}
\end{figure}

\paragraph{How accurate is the alignment of 3D proxy and final 3DGS scene?} To quantify the alignment, we calculate the depth error between rendered depth maps from the 3D proxy and the generated 3DGS scene. Additionally, the alignment between ground truth camera poses and estimated camera poses for the corresponding generated NVS frames (we use~\cite{keetha2025mapanything} to estimate the metric camera poses) is reported. As shown in Table~\ref{tab:depth_alignment}, the low errors for both depth and camera pose alignment highlight the accuracy of our approach. Figure~\ref{fig:annotations} provides visualizations of transferred 9D poses and semantic maps from 3D proxy to rendered novel views and the generated 3DGS scene, further highlighting the accurate alignment.

\begin{table}[]
\centering
\scalebox{0.85}{
\begin{tabular}{cccccccc}
\toprule
\multicolumn{2}{c}{Cam. Transl. Error [m]}&&\multicolumn{2}{c}{Cam. Rot. Error [\textdegree]} && \multicolumn{2}{c}{Depth Error [m]}\\
\cmidrule{1-2} \cmidrule{4-5} \cmidrule{7-8}
RMSE&Median&&RMSE&Median&&RMSE& AbsRel \\
\midrule
0.036& 0.032&& 1.05 & 0.77&& 0.117  & 0.014\\
\bottomrule
\end{tabular}
}
\vspace{3pt}
\caption{\textbf{Quantitative validation of absolute scale and alignment.} Low errors between GT and estimated camera poses indicate small drifts. Additionally, the small deviation between rendered depth maps from the 3D proxy and the generated 3DGS scene highlights the accurate alignment in the absolute global coordinate frame.}

\label{tab:depth_alignment} 
\vspace{-15pt}
\end{table}

\subsection{Application: Progressive and Seamless Scene Expansion}

\begin{figure}[h!]
\centering
\scalebox{.99}{
\includegraphics[trim={0.5cm 5.5cm 3.4cm .5cm},clip,width=\linewidth]{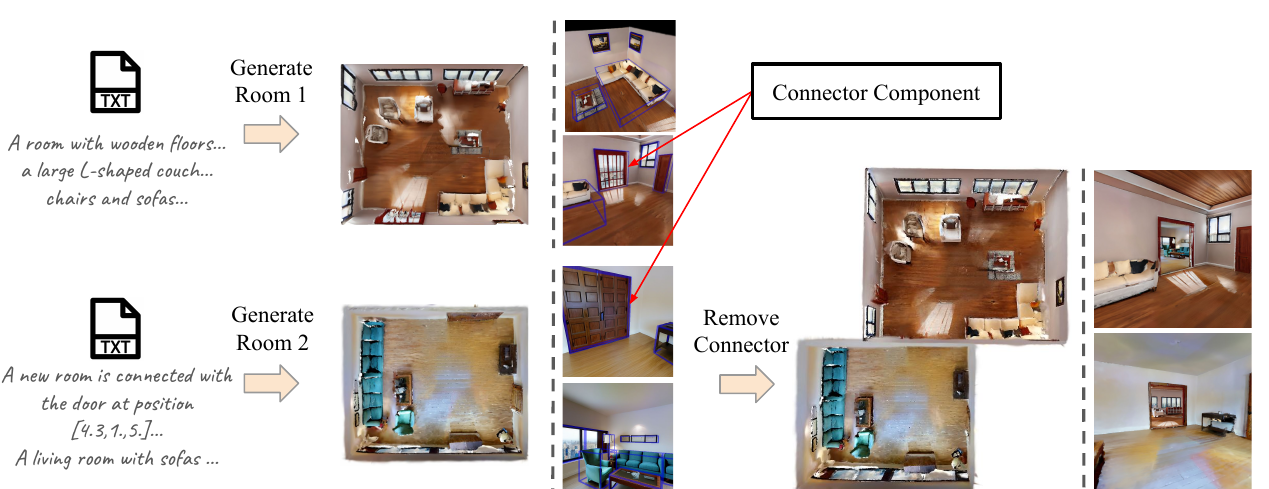}
}
\caption{\textbf{Scene Expansion Example.} After a first scene is generated, our frame work enables seamless iterative scene expansion. First, the previous 3D proxy as well as a textual description of the new scene is given as input to Holodeck~\cite{Yang_2024_CVPR_holodeck}. This enables us to extend the 3D proxy in the global world coordinate frame. Then, each room can be generated independently, whereas the generated results will automatically align due to the 3D proxy guidance. To seamlessly connect the rooms, we can remove the Gaussians of the connector component by utilizing the corresponding semantic and pose information. Notably, our scene expansion application is iterative and single scenes can be generated independently, hence large scenes generation with multiple rooms can be achieved without increasing memory or computation demands.}
\label{fig:scene_expansion}
\end{figure}

To achieve scene expansion, we first utilize Holodeck to extend the initial 3D proxy layout by connecting a new room to the already existing scene utilizing a \textit{connector component}, which can be a door or a (temporary) wall. Next, we reconstruct the new room using our pipeline, which directly ensures correct placement in the global coordinate frame. We then utilize the available 9D poses and semantic information to remove all Gaussian splats that are assigned to the connector component, effectively merging the two scenes. Figure \ref{fig:scene_expansion}  shows an example. We refer to the supplementary material for additional visualizations for scene expansion.

\section{Conclusion}
\label{sec:conclusion}
We introduce a novel text-to-3D scene generation model that uses a metric 3D proxy layout to produce fully navigable 3D scenes with highly coherent novel-view synthesis. By enforcing a metric, absolute global coordinate frame end-to-end, our pipeline preserves and propagates global 3D structure, enabling direct transfer of 9D poses and segmentation from the proxy to the final scene, and seamless scene expansion without extra alignment or rescaling. Future work includes extending our approach to outdoor environments, and exploring alternative proxy generation mechanisms to further increase scene diversity.

%
%
\bibliographystyle{splncs04}
\bibliography{main}

\begin{thebibliography}{10}
\providecommand{\url}[1]{\texttt{#1}}
\providecommand{\urlprefix}{URL }
\providecommand{\doi}[1]{https://doi.org/#1}

\bibitem{asim24met3r}
Asim, M., Wewer, C., Wimmer, T., Schiele, B., Lenssen, J.E.: Met3r: Measuring multi-view consistency in generated images. In: Conference on Computer Vision and Pattern Recognition (2025)

\bibitem{blattmann2023stable}
Blattmann, A., Dockhorn, T., Kulal, S., Mendelevitch, D., Kilian, M., Lorenz, D., Levi, Y., English, Z., Voleti, V., Letts, A., et~al.: Stable video diffusion: Scaling latent video diffusion models to large datasets. arXiv preprint arXiv:2311.15127  (2023)

\bibitem{Matterport3D}
Chang, A., Dai, A., Funkhouser, T., Halber, M., Nie{\ss}ner, M., Savva, M., Song, S., Zeng, A., Zhang, Y.: Matterport3d: Learning from rgb-d data in indoor environments. In: International Conference on 3D Vision (2017)

\bibitem{chen2024pgsr}
Chen, D., Li, H., Ye, W., Wang, Y., Xie, W., Zhai, S., Wang, N., Liu, H., Bao, H., Zhang, G.: Pgsr: Planar-based gaussian splatting for efficient and high-fidelity surface reconstruction. IEEE Transactions on Visualization and Computer Graphics  \textbf{31}(9),  6100--6111 (2024)

\bibitem{chen2025flexworld}
Chen, L., Zhou, Z., Zhao, M., Wang, Y., Zhang, G., Huang, W., Sun, H., Wen, J.R., Li, C.: Flexworld: Progressively expanding 3d scenes for flexiable-view synthesis. arXiv preprint arXiv:2503.13265  (2025)

\bibitem{chen2025janus}
Chen, X., Wu, Z., Liu, X., Pan, Z., Liu, W., Xie, Z., Yu, X., Ruan, C.: Janus-pro: Unified multimodal understanding and generation with data and model scaling. arXiv preprint arXiv:2501.17811  (2025)

\bibitem{chen2025comboverse}
Chen, Y., Wang, T., Wu, T., Pan, X., Jia, K., Liu, Z.: Comboverse: Compositional 3d assets creation using spatially-aware diffusion guidance. In: European Conference on Computer Vision. pp. 128--146. Springer (2024)

\bibitem{cohen2023set}
Cohen-Bar, D., Richardson, E., Metzer, G., Giryes, R., Cohen-Or, D.: Set-the-scene: Global-local training for generating controllable nerf scenes. In: International Conference on Computer Vision Workshops. pp. 2920--2929 (2023)

\bibitem{dominici2025dreamanywhere}
Dominici, E.A., Hladk{\`y}, J., Verhoeven, F., Radl, L., Deixelberger, T., Ainetter, S., Drescher, P., Hauswiesner, S., Coomans, A., Nazzaro, G., et~al.: Dreamanywhere: Object-centric panoramic 3d scene generation. In: IEEE Winter Conference on Applications of Computer Vision. pp. 1--11 (2026)

\bibitem{epstein2024disentangled}
Epstein, D., Poole, B., Mildenhall, B., Efros, A.A., Holynski, A.: Disentangled 3d scene generation with layout learning. In: International Conference on Machine Learning (2024)

\bibitem{fang2025ctrl}
Fang, C., Dong, Y., Luo, K., Hu, X., Shrestha, R., Tan, P.: Ctrl-room: Controllable text-to-3d room meshes generation with layout constraints. In: International Conference on 3D Vision. pp. 692--701. IEEE (2025)

\bibitem{SpatialGen}
Fang, C., Li, H., Liang, Y., Zheng, J., Mao, Y., Liu, Y., Tang, R., Zhou, Z., Tan, P.: Spatialgen: Layout-guided 3d indoor scene generation. In: International Conference on 3D Vision (2026)

\bibitem{feng2023diffusion360}
Feng, M., Liu, J., Cui, M., Xie, X.: Diffusion360: Seamless 360 degree panoramic image generation based on diffusion models. arXiv preprint arXiv:2311.13141  (2023)

\bibitem{scenescape2023}
Fridman, R., Abecasis, A., Kasten, Y., Dekel, T.: Scenescape: Text-driven consistent scene generation. In: Advances in Neural Information Processing Systems. vol.~36, pp. 39897--39914 (2023)

\bibitem{gardner2017laval_indoor}
Gardner, M.A., Sunkavalli, K., Yumer, E., Shen, X., Gambaretto, E., Gagn\'{e}, C., Lalonde, J.F.: Learning to predict indoor illumination from a single image. ACM Trans. Graph.  \textbf{36}(6) (Nov 2017)

\bibitem{gu2025diffusion}
Gu, Z., Yan, R., Lu, J., Li, P., Dou, Z., Si, C., Dong, Z., Liu, Q., Lin, C., Liu, Z., et~al.: Diffusion as shader: 3d-aware video diffusion for versatile video generation control. In: SIGGRAPH 2025 Conference Papers (2025)

\bibitem{clipscore}
Hessel, J., Holtzman, A., Forbes, M., Le~Bras, R., Choi, Y.: {CLIPS}core: A reference-free evaluation metric for image captioning (Nov 2021)

\bibitem{NEURIPS2020_4c5bcfec}
Ho, J., Jain, A., Abbeel, P.: Denoising diffusion probabilistic models. In: Larochelle, H., Ranzato, M., Hadsell, R., Balcan, M., Lin, H. (eds.) Advances in Neural Information Processing Systems. vol.~33, pp. 6840--6851 (2020)

\bibitem{ho2022classifier}
Ho, J., Salimans, T.: Classifier-free diffusion guidance. In: Advances in Neural Information Processing Systems (2021), \url{https://openreview.net/forum?id=qw8AKxfYbI}

\bibitem{VDM}
Ho, J., Salimans, T., Gritsenko, A., Chan, W., Norouzi, M., Fleet, D.J.: Video diffusion models. In: Advances in Neural Information Processing Systems. NIPS '22, Curran Associates Inc., Red Hook, NY, USA (2022)

\bibitem{hoellein2023text2room}
H\"ollein, L., Cao, A., Owens, A., Johnson, J., Nie{\ss}ner, M.: Text2room: Extracting textured 3d meshes from 2d text-to-image models. In: International Conference on Computer Vision. pp. 7909--7920 (October 2023)

\bibitem{hu2022lora}
Hu, E.J., Shen, Y., Wallis, P., Allen-Zhu, Z., Li, Y., Wang, S., Wang, L., Chen, W., et~al.: Lora: Low-rank adaptation of large language models. International Conference for Learning Representations  \textbf{1}(2), ~3 (2022)

\bibitem{voyager}
Huang, T., Zheng, W., Wang, T., Liu, Y., Wang, Z., Wu, J., Jiang, J., Li, H., Lau, R., Zuo, W., et~al.: Voyager: Long-range and world-consistent video diffusion for explorable 3d scene generation. ACM Transactions on Graphics (TOG)  \textbf{44}(6),  1--15 (2025)

\bibitem{huang2025dreamcube}
Huang, Y., Zhou, Y., Wang, J., Huang, K., Liu, X.: {DreamCube: RGB-D Panorama Generation via Multi-plane Synchronization}. In: International Conference on Computer Vision. pp. 24922--24932 (October 2025)

\bibitem{hunyuanworld2025tencent}
HunyuanWorld, T.: Hunyuanworld 1.0: Generating immersive, explorable, and interactive 3d worlds from words or pixels. arXiv preprint  (2025)

\bibitem{kalischek2025cubediff}
Kalischek, N., Oechsle, M., Manhardt, F., Henzler, P., Schindler, K., Tombari, F.: Cubediff: Repurposing diffusion-based image models for panorama generation. In: International Conference for Learning Representations (2025)

\bibitem{keetha2025mapanything}
Keetha, N., M{\"u}ller, N., Sch{\"o}nberger, J., Porzi, L., Zhang, Y., Fischer, T., Knapitsch, A., Zauss, D., Weber, E., Antunes, N., et~al.: Mapanything: Universal feed-forward metric 3d reconstruction. In: International Conference on 3D Vision. pp. 499--509. IEEE (2026)

\bibitem{kerbl20233d}
Kerbl, B., Kopanas, G., Leimk{\"u}hler, T., Drettakis, G.: 3d gaussian splatting for real-time radiance field rendering. ACM Trans. Graph.  \textbf{42}(4) (2023)

\bibitem{flux2024}
Labs, B.F.: Flux. \url{https://github.com/black-forest-labs/flux} (2024)

\bibitem{dreamsceneLi2024}
Li, H., Shi, H., Zhang, W., Wu, W., Liao, Y., Wang, L., Lee, L.h., Zhou, P.Y.: Dreamscene: 3d gaussian-based text-to-3d scene generation via formation pattern sampling. In: European Conference on Computer Vision. pp. 214--230. Springer (2024)

\bibitem{liang2024wonderland}
Liang, H., Cao, J., Goel, V., Qian, G., Korolev, S., Terzopoulos, D., Plataniotis, K.N., Tulyakov, S., Ren, J.: Wonderland: Navigating 3d scenes from a single image. In: Conference on Computer Vision and Pattern Recognition. pp. 798--810 (2025)

\bibitem{viewselectionOLAGUE2002}
Olague, G., Mohr, R.: Optimal camera placement for accurate reconstruction. Pattern Recognition  \textbf{35}(4),  927--944 (2002)

\bibitem{poole2023dreamfusion}
Poole, B., Jain, A., Barron, J.T., Mildenhall, B.: Dreamfusion: Text-to-3d using 2d diffusion. In: International Conference for Learning Representations (2023)

\bibitem{ren2025gen3c3dinformedworldconsistentvideo}
Ren, X., Shen, T., Huang, J., Ling, H., Lu, Y., Nimier-David, M., M{\"u}ller, T., Keller, A., Fidler, S., Gao, J.: Gen3c: 3d-informed world-consistent video generation with precise camera control. In: Conference on Computer Vision and Pattern Recognition. pp. 6121--6132 (2025)

\bibitem{rombach2022high}
Rombach, R., Blattmann, A., Lorenz, D., Esser, P., Ommer, B.: High-resolution image synthesis with latent diffusion models. In: Conference on Computer Vision and Pattern Recognition. pp. 10684--10695 (2022)

\bibitem{salimans2016improved}
Salimans, T., Goodfellow, I., Zaremba, W., Cheung, V., Radford, A., Chen, X.: Improved techniques for training gans. Advances in Neural Information Processing Systems  \textbf{29} (2016)

\bibitem{schneider_hoellein_2025_worldexplorer}
Schneider, M.A., H{\"o}llein, L., Nie{\ss}ner, M.: Worldexplorer: Towards generating fully navigable 3d scenes. In: SIGGRAPH Asia 2025 Conference Papers. pp. 1--11 (2025)

\bibitem{schult2024controlroom3d}
Schult, J., Tsai, S., H{\"o}llein, L., Wu, B., Wang, J., Ma, C.Y., Li, K., Wang, X., Wimbauer, F., He, Z., et~al.: Controlroom3d: Room generation using semantic proxy rooms. In: Conference on Computer Vision and Pattern Recognition. pp. 6201--6210 (2024)

\bibitem{shi2023MVDream}
Shi, Y., Wang, P., Ye, J., Mai, L., Li, K., Yang, X.: Mvdream: Multi-view diffusion for 3d generation. In: International Conference for Learning Representations. vol.~2024, pp. 39838--39859 (2024)

\bibitem{shriram2024realmdreamertextdriven3dscene}
Shriram, J., Trevithick, A., Liu, L., Ramamoorthi, R.: Realmdreamer: Text-driven 3d scene generation with inpainting and depth diffusion. arXiv preprint arXiv:2404.07199  (2024)

\bibitem{silberman2012indoor}
Silberman, N., Hoiem, D., Kohli, P., Fergus, R.: Indoor segmentation and support inference from rgbd images. In: European Conference on Computer Vision. pp. 746--760. Springer (2012)

\bibitem{song2021denoising}
Song, J., Meng, C., Ermon, S.: Denoising diffusion implicit models. In: International Conference for Learning Representations (2021), \url{https://openreview.net/forum?id=St1giarCHLP}

\bibitem{viewselectionsun2021}
Sun, Y., Huang, Q., Hsiao, D.Y., Guan, L., Hua, G.: Learning view selection for 3d scenes. In: Conference on Computer Vision and Pattern Recognition. pp. 14464--14473 (June 2021)

\bibitem{szymanowicz2025bolt3d}
Szymanowicz, S., Zhang, J.Y., Srinivasan, P., Gao, R., Brussee, A., Holynski, A., Martin-Brualla, R., Barron, J.T., Henzler, P.: {Bolt3D: Generating 3D Scenes in Seconds}. International Conference on Computer Vision  (2025)

\bibitem{tang2023MVDiffusion}
Tang, S., Zhang, F., Chen, J., Wang, P., Furukawa, Y.: Mvdiffusion: Enabling holistic multi-view image generation with correspondence-aware diffusion. In: Advances in Neural Information Processing Systems (2023)

\bibitem{wan2025wan}
Wan, T., Wang, A., Ai, B., Wen, B., Mao, C., Xie, C.W., Chen, D., Yu, F., Zhao, H., Yang, J., et~al.: Wan: Open and advanced large-scale video generative models. arXiv preprint arXiv:2503.20314  (2025)

\bibitem{vistadream2025}
Wang, H., Liu, Y., Liu, Z., Wang, W., Dong, Z., Yang, B.: Vistadream: Sampling multiview consistent images for single-view scene reconstruction. In: International Conference on Computer Vision. pp. 26772--26782 (October 2025)

\bibitem{clipiqa}
Wang, J., Chan, K.C., Loy, C.C.: Exploring clip for assessing the look and feel of images. In: American Association for Artificial Intelligence Conference (2023)

\bibitem{qalign}
Wu, H., Zhang, Z., Zhang, W., Chen, C., Li, C., Liao, L., Wang, A., Zhang, E., Sun, W., Yan, Q., Min, X., Zhai, G., Lin, W.: Q-align: Teaching lmms for visual scoring via discrete text-defined levels (2023), equal Contribution by Wu, Haoning and Zhang, Zicheng. Project Lead by Wu, Haoning. Corresponding Authors: Zhai, Guangtai and Lin, Weisi.

\bibitem{xie2021segformer}
Xie, E., Wang, W., Yu, Z., Anandkumar, A., Alvarez, J.M., Luo, P.: Segformer: Simple and efficient design for semantic segmentation with transformers. Advances in Neural Information Processing Systems  \textbf{34},  12077--12090 (2021)

\bibitem{yang2024depth}
Yang, L., Kang, B., Huang, Z., Zhao, Z., Xu, X., Feng, J., Zhao, H.: Depth anything v2. Advances in Neural Information Processing Systems  \textbf{37},  21875--21911 (2024)

\bibitem{yang2024layerpano3d}
Yang, S., Tan, J., Zhang, M., Wu, T., Wetzstein, G., Liu, Z., Lin, D.: Layerpano3d: Layered 3d panorama for hyper-immersive scene generation. In: SIGGRAPH 2025 Conference Papers (2025)

\bibitem{yang2024scenecraft}
Yang, X., Man, Y., Chen, J., Wang, Y.X.: Scenecraft: Layout-guided 3d scene generation. Advances in Neural Information Processing Systems  \textbf{37},  82060--82084 (2024)

\bibitem{Yang_2024_CVPR_holodeck}
Yang, Y., Sun, F.Y., Weihs, L., VanderBilt, E., Herrasti, A., Han, W., Wu, J., Haber, N., Krishna, R., Liu, L., Callison-Burch, C., Yatskar, M., Kembhavi, A., Clark, C.: Holodeck: Language guided generation of 3d embodied ai environments. In: Conference on Computer Vision and Pattern Recognition. pp. 16227--16237 (June 2024)

\bibitem{yu2025wonderworld}
Yu, H.X., Duan, H., Herrmann, C., Freeman, W.T., Wu, J.: Wonderworld: Interactive 3d scene generation from a single image. In: Conference on Computer Vision and Pattern Recognition. pp. 5916--5926 (2025)

\bibitem{yu2023wonderjourney}
Yu, H.X., Duan, H., Hur, J., Sargent, K., Rubinstein, M., Freeman, W.T., Cole, F., Sun, D., Snavely, N., Wu, J., Herrmann, C.: Wonderjourney: Going from anywhere to everywhere. In: Conference on Computer Vision and Pattern Recognition (2024)

\bibitem{yu2024viewcrafter}
Yu, W., Xing, J., Yuan, L., Hu, W., Li, X., Huang, Z., Gao, X., Wong, T.T., Shan, Y., Tian, Y.: Viewcrafter: Taming video diffusion models for high-fidelity novel view synthesis. arXiv preprint arXiv:2409.02048  (2024)

\bibitem{Zhang2023Text2NeRFT3}
Zhang, J., Li, X., Wan, Z., Wang, C., Liao, J.: Text2nerf: Text-driven 3d scene generation with neural radiance fields. IEEE Transactions on Visualization and Computer Graphics  \textbf{30}(12),  7749--7762 (2024)

\bibitem{zhang2023controlnet}
Zhang, L., Rao, A., Agrawala, M.: Adding conditional control to text-to-image diffusion models. In: International Conference on Computer Vision. pp. 3836--3847 (2023)

\bibitem{zhang2023scenewiz3d}
Zhang, Q., Wang, C., Siarohin, A., Zhuang, P., Xu, Y., Yang, C., Lin, D., Zhou, B., Tulyakov, S., Lee, H.Y.: Scenewiz3d: Towards text-guided 3d scene composition. arXiv preprint arXiv:2312.08885  (2023)

\bibitem{zhang2025generating}
Zhang, Z., Hold-Geoffroy, Y., Ha{\v{s}}an, M., Chen, Z., Luan, F., Dorsey, J., Hu, Y.: Generating 360° video is what you need for a 3d scene. In: Proceedings of the SIGGRAPH Asia 2025 Conference Papers. pp. 1--12 (2025)

\bibitem{Structured3D}
Zheng, J., Zhang, J., Li, J., Tang, R., Gao, S., Zhou, Z.: Structured3d: A large photo-realistic dataset for structured 3d modeling. In: European Conference on Computer Vision (2020)

\bibitem{zhou2025stable}
Zhou, J., Gao, H., Voleti, V., Vasishta, A., Yao, C.H., Boss, M., Torr, P., Rupprecht, C., Jampani, V.: Stable virtual camera: Generative view synthesis with diffusion models. In: International Conference on Computer Vision. pp. 12405--12414 (2025)

\bibitem{zhou2024dreamscene360}
Zhou, S., Fan, Z., Xu, D., Chang, H., Chari, P., Bharadwaj, T., You, S., Wang, Z., Kadambi, A.: Dreamscene360: Unconstrained text-to-3d scene generation with panoramic gaussian splatting. In: European Conference on Computer Vision. pp. 324--342. Springer (2024)

\end{thebibliography}

\newpage
\appendix

{\LARGE{\textbf{Supplementary Material}}}\\

\section{Implementation Details about Guided Panoramic Image Generation}
\label{sec:supp_guided_pano}
We propose a novel architecture for our guided panoramic image generator, which predicts multiple perspective views of a scene in the form of a cubemap and utilizes depth and semantic information as guidance signal. Hereafter, we describe the data pre-processing, and provide further details about implementation, training and inference.

\subsection{Data Pre-processing}
\label{supp:data_preprocessing}
We collect panoramic images from multiple publicly available datasets~\cite{Structured3D,Matterport3D,gardner2017laval_indoor}. For each panorama image, we convert the equirectangular representations into six perspective views in the form of a cubemap. Each perspective view has a resolution of $512 \times 512$px. 
Following CubeDiff~\cite{kalischek2025cubediff}, we ensure that the perspective views have a slight overlap by choosing a 95\degree\ field of view.
Providing these overlap information to the network during training helps to enhance color and geometric consistency across cube faces. The resulting generated views are then cropped and re-assembled into a 90\degree\ field of view cubemap during post-processing. 

In addition to the image pre-processing, we extract a textual description for each panoramic image using Janus Pro 7B~\cite{chen2025janus}. This textual description is provided to the diffusion model as text condition for each cube face.

In total, our dataset for training the cubemap-based multi-view diffusion model consists of approximately 54k panoramic images.

\subsection{Model Architecture}
We broadly follow the model architecture introduced by CubeDiff~\cite{kalischek2025cubediff}, which utilizes a noise-free reference view as image condition during training. All cube faces are first encoded into latent space using a VAE encoder, and noise is added to each face unless it is a reference view. We then concatenate a per-view binary mask and xyz-positional encoding~\cite{huang2025dreamcube} to the extracted latents, and use this as input for the U-Net of our diffusion model. Note that the binary mask indicates if the cube face is a clean reference view or noisy (noisy views have to be generated by the model), and the positional encoding provides the Latent Diffusion Model (LDM) with spatial awareness within the cubemap. Due to the fact that we concatenate a binary mask and positional encodings to the latents, we have to extend the first convolutional layer of the U-Net to eight input channels per view. In detail, the new layer consists of four channels for the extracted latents, one additional channel for the binary mask, and three additional channels for the xyz-positional encoding, whereas the newly added weights are initialized with zeros. Additionally, all self- and cross-attention layers are extended to inflated attention.

We initialize the LDM with a pre-trained Stable Diffusion 2~\cite{rombach2022high} model. During training, we fine-tune the first convolutional layer of the U-Net, as well as the inflated self- and cross-attention layers. All other parts of the LDM are kept frozen.

\subsection{Training Details and Inference Setup}
Our training procedure consists of two consecutive steps. First, we train our cubemap-based multi-view diffusion model. Next, we extend it with a Multi-ControlNet~\cite{zhang2023controlnet} structure and finetune it to enable semantic and depth guided panoramic image generation. Hereafter, we give details about the training procedures for both steps.

\paragraph{Multi-view Diffusion Model Training.}
The model is trained for 120,000 iterations with a batch size of 2. We use the AdamW optimizer ($\beta_1=0.9, \beta_2=0.999, \gamma=1e^{-2}$) and a learning rate of $1e^{-5}$, linearly ramped up over 10,000 warm-up steps. To prevent gradient explosion, we apply gradient clipping with a maximum norm of 1.
The model is fine-tuned using v-prediction~\cite{ho2022classifier}, and we employ the DDPM noise scheduler~\cite{NEURIPS2020_4c5bcfec} with 1000 time steps.
Note that we provide a clean reference image as input during training of the multi-view diffusion model, but randomly drop this input signal with a probability of $20\%$.

\paragraph{Finetuning the Multi-ControlNet.}
We extend the architecture of our multi-view diffusion model by adding two separate ControlNet branches for depth and semantic information. Each of the ControlNet branches is initialized with a trained copy of the U-Net from our trained multi-view diffusion model. Hence, all architectural adaptations, namely the inflated attention and the expanded first layer of the U-Net, are also integrated in the ControlNet branches.
For finetuning, we utilize the Structured3D dataset, which provides depth and semantic panoramic images as ground truth. We employ the same pre-processing strategy as detailed in Subsection~\ref{supp:data_preprocessing} for the semantic and depth data. Regarding the optimizer and hyper-parameters, we use the same configuration as for the multi-view diffusion model, and fine-tune for 90,000 iterations with a batch size of 1.

\paragraph{Inference Setup.}
At inference time we use the textual description of the scene, as well as the corresponding depth and semantic renderings from our global scene layout as input. We employ the DDIM noise scheduler~\cite{song2021denoising} with 50 sampling steps and use a value of 8.0 for classifier-free guidance. 

\section{Details about NVS with Stable Virtual Camera}
We first extract eight perspective views from the initially generated panoramic image, which we denote as $\{I_k\}_{k=1}^8$. For each of these images, we define a fixed camera position at the center of the panoramic view, with an image specific camera rotation $R_k$ around the up-axis of the scene defined as
\begin{equation}
\ R_k = R_{up}(\phi_k),\phi_k \in \{ 0^{\circ},45^{\circ},90^{\circ},135^{\circ},180^{\circ},225^{\circ},270^{\circ},315^{\circ} \}.
\end{equation}
For each of the initial frames, we define the pinhole camera intrinsics with a field of view (FoV) of $72^{\circ}$, whereas each frame has a resolution of $576 \times 576$px. We empirically found that while Stable Virtual Camera produces reasonable results for various FoVs, this configuration in combination with our depth-guided alignment led to the highest 3D consistency of generated novel views and best alignment with the global scene layout.
For each of our trajectories, we select one frame from $\{I_k\}_{k=1}^8$ as starting point, depending on which area of the scene we want to cover. Additionally, all initial frames are used as context frames for Stable Virtual Camera, which conditions the novel view synthesis to the information from the initial panoramic image.
For each trajectory, we therefore use the initial eight frames as context frames, and then sample novel views with a sequence length of 42.

\section{Implementation Details about 3DGS Reconstruction}
Our 3DGS implementation is based on PGSR~\cite{chen2024pgsr}, utilizing their unbiased depth and normal convention. We employ the same training strategy and hyperparameter setup as described in~\cite{chen2024pgsr}, however, we reduced the number of training iterations to 7000. This led to sufficient results in our experiments, as the initial point cloud already provides a highly accurate initialization. 

Our training loss $\mathcal{L}_{\text{}}$ is defined as
\begin{equation}
\label{eq:our_full_loss}
\mathcal{L}_{\text{}} =\mathcal{L}_{\text{3DGS}} +  \mathcal{L}_{\text{geom}} +\mathcal{L}_{\text{NN}} + \mathcal{L}_{\text{depth}},
\end{equation}
where $\mathcal{L}_{\text{3DGS}}$ is the original 3DGS loss proposed in~\cite{kerbl20233d}, $\mathcal{L}_{\text{geom}}$ was introduced in~\cite{chen2024pgsr}, and $\mathcal{L}_{\text{NN}}$ and $\mathcal{L}_{\text{depth}}$ are our 3D nearest neighbor loss and masked depth loss, respectively. Hereafter, we provide additional information for each loss term.
\paragraph{3DGS Loss.}
The 3DGS loss $\mathcal{L}_{\text{3DGS}}$ is defined as
\begin{equation}
\mathcal{L}_{\text{3DGS}} =(1-\lambda) \mathcal{L}_1( I,\hat{I} )+ \lambda \mathcal{L}_{DSSIM}(I,\hat{I}),
\end{equation}
where $\hat{I}$ is a rendered frame from the 3DGS representation, $I$ is the corresponding generated frame, and $\lambda$ is a weighting parameter.
\paragraph{Geometric Loss $\mathcal{L}_{\text{geom}}$.}
The loss term $\mathcal{L}_{\text{geom}}$ was introduced in~\cite{chen2024pgsr}, and is defined as
\begin{equation}
\mathcal{L}_{\text{geom}} = \lambda_2 \mathcal{L}_{svgeom} + \lambda_3 \mathcal{L}_{mvrgb} + \lambda_4 \mathcal{L}_{mvgeom}.
\end{equation}
Here, $\mathcal{L}_{svgeom}$ is the image edge-aware single-view loss where the difference between the blended normals from Gaussian splats referring to a pixel, and the calculated normal of the local plane of the neighboring pixels are minimized. $\mathcal{L}_{mvrgb}$ describes the multi-view photometric consistency loss created through normalized cross correlation of image patches of a neighboring frame, projected into to the current frame. The multi-view geometric consistency loss $\mathcal{L}_{mvgeom}$ minimizes the error of forward and backward projection of pixels in a neighboring frame. We refer the interested reader to the original paper~\cite{chen2024pgsr} for a more detailed explanation.

\paragraph{Masked Depth Loss.}
To enforce metrically accurate geometry, we supervise the rendered depth using a ground-truth depth map rendered from our global scene layout. To achieve this, we define our masked depth loss $\mathcal{L}_{\text{depth}}$ as
\begin{equation}
\mathcal{L}_{\text{depth}}=\frac{1}{\sum_{j} M^j}\sum_{j}M^j \,||D^j_{gt} - D^j_{rd} ||_1.
\end{equation}
Here, $D^j_{rd}$ denotes the rendered depth at pixel $j$, and $D^j_{gt}$ defines the corresponding metric ground-truth depth rendered from our global scene layout. 
We additionally utilize a binary segmentation mask $M^j \in \{0,1\}$, which indicates regions that we want to exclude from supervision, for example windows, doors and mirrors as the rendered depth can potentially be undefined in this regions, and all objects since they do not follow the new scene geometry. This loss encourages the rendered geometry to match the metric depth of the scene while preventing object regions from biasing the supervision signal.

\paragraph{3D Nearest-Neighbor Loss.}
To encourage geometric consistency between the reconstructed representation and the initial point cloud, we employ a nearest-neighbor loss. For each 3D Gaussian mean $m_l \in \mathbb{R}^3$, we retrieve its closest point $n_l \in \mathbb{R}^3$ from the initial point cloud using a 1-NN lookup. This loss serves as regularization term that explicitly constrains the predicted geometry to remain close to the initial scene layout, while allowing the freedom to represent the new structure. To reduce computational complexity, the nearest neighbor search is done only every 1000 iterations.
Let $\mathcal{GM} = \{m_l\}_{l=1}^{C}$ denote the set of Gaussian means as points and $\mathcal{NN} = \{n_l\}_{l=1}^{C}$ their corresponding nearest neighbors. $\mathcal{L}_{\text{NN}}$ is defined as mean of the square-root distance between each pair:
\begin{equation}
\mathcal{L}_{\text{NN}}
= \frac{1}{C}
\sum_{l=1}^{C}
\sqrt{| m_l - n_l |}.
\end{equation}

Figure~\ref{fig:losses2} provides a visualization of the influence of our different losses.

\begin{figure*}
\centering
\scalebox{0.7}{
\begin{tabular}{ccccc}

\includegraphics[width=0.20\linewidth]{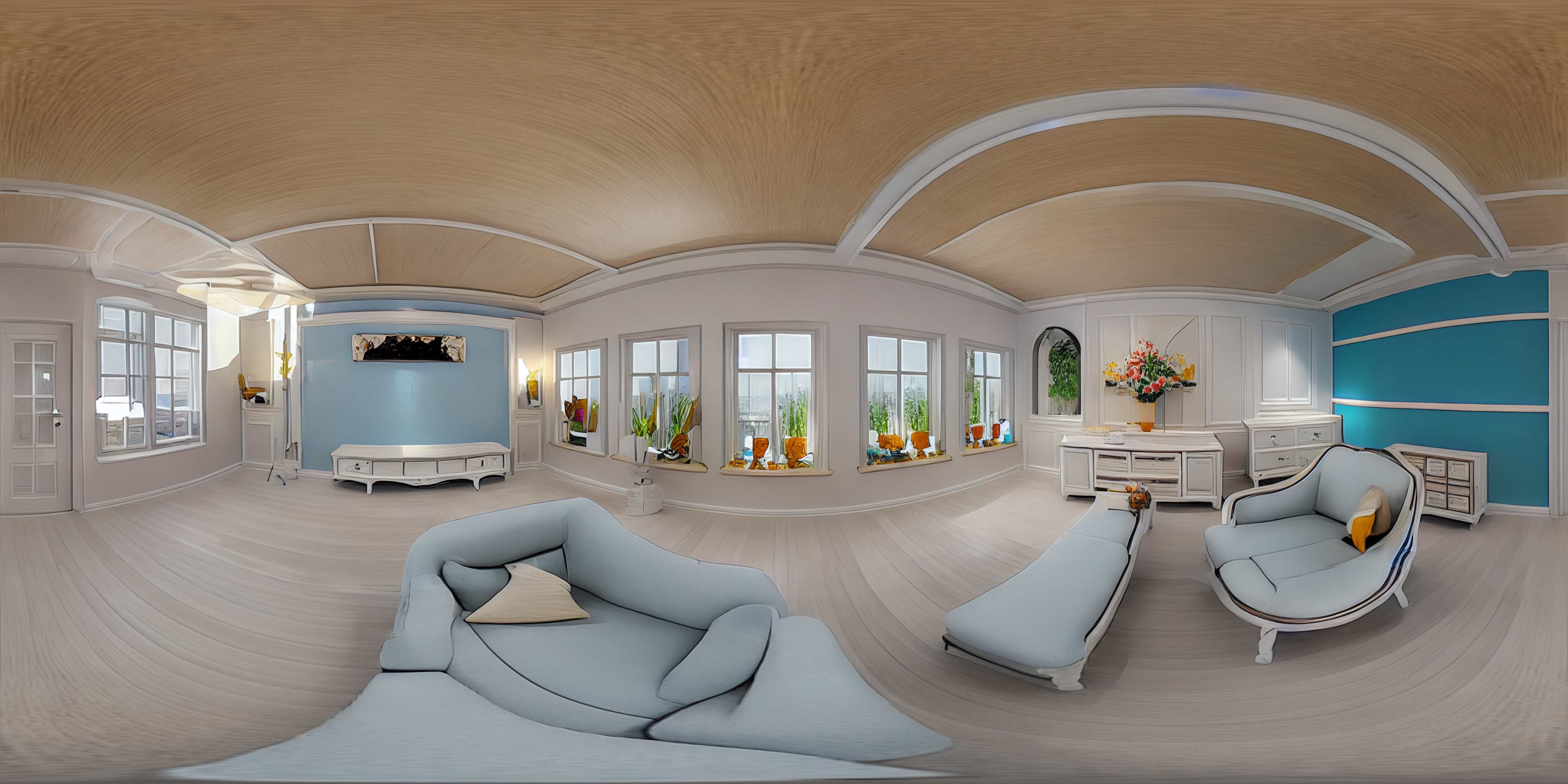} &
\includegraphics[width=0.20\linewidth]{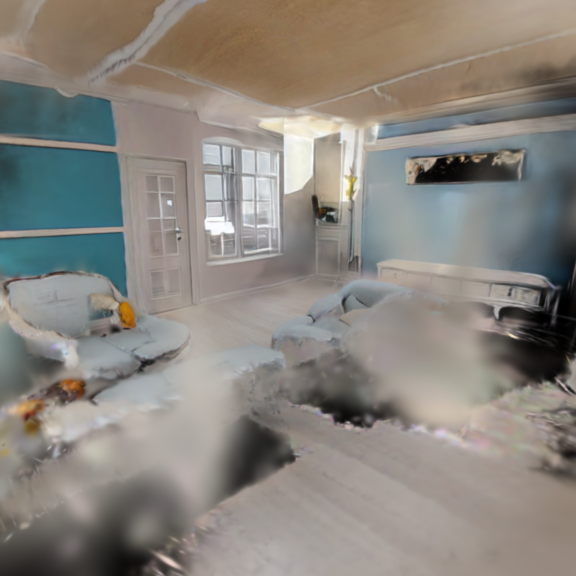} &
\includegraphics[width=0.20\linewidth]{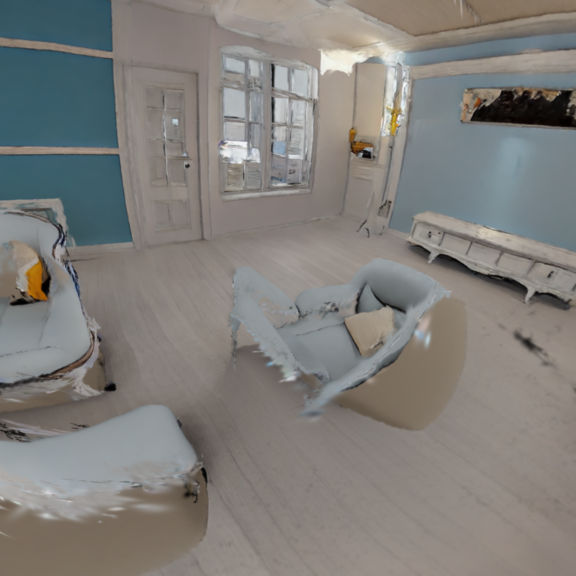} &
\includegraphics[width=0.20\linewidth]{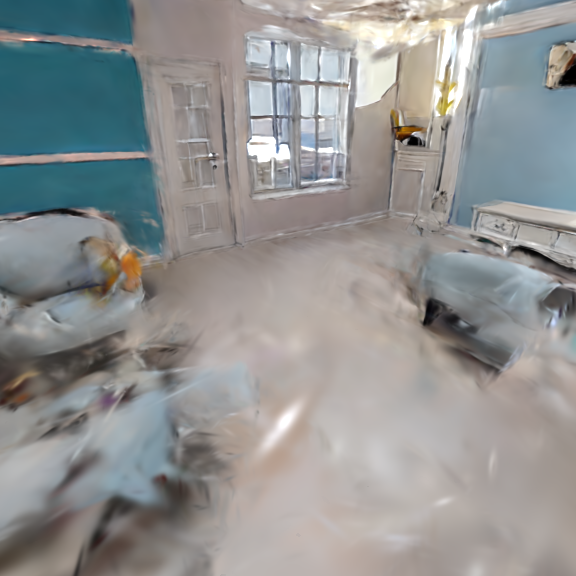} &
\includegraphics[width=0.20\linewidth]{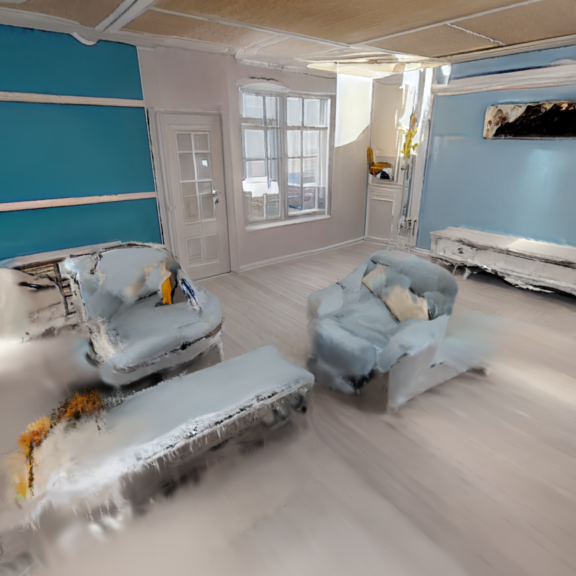}\\

\includegraphics[width=0.20\linewidth]{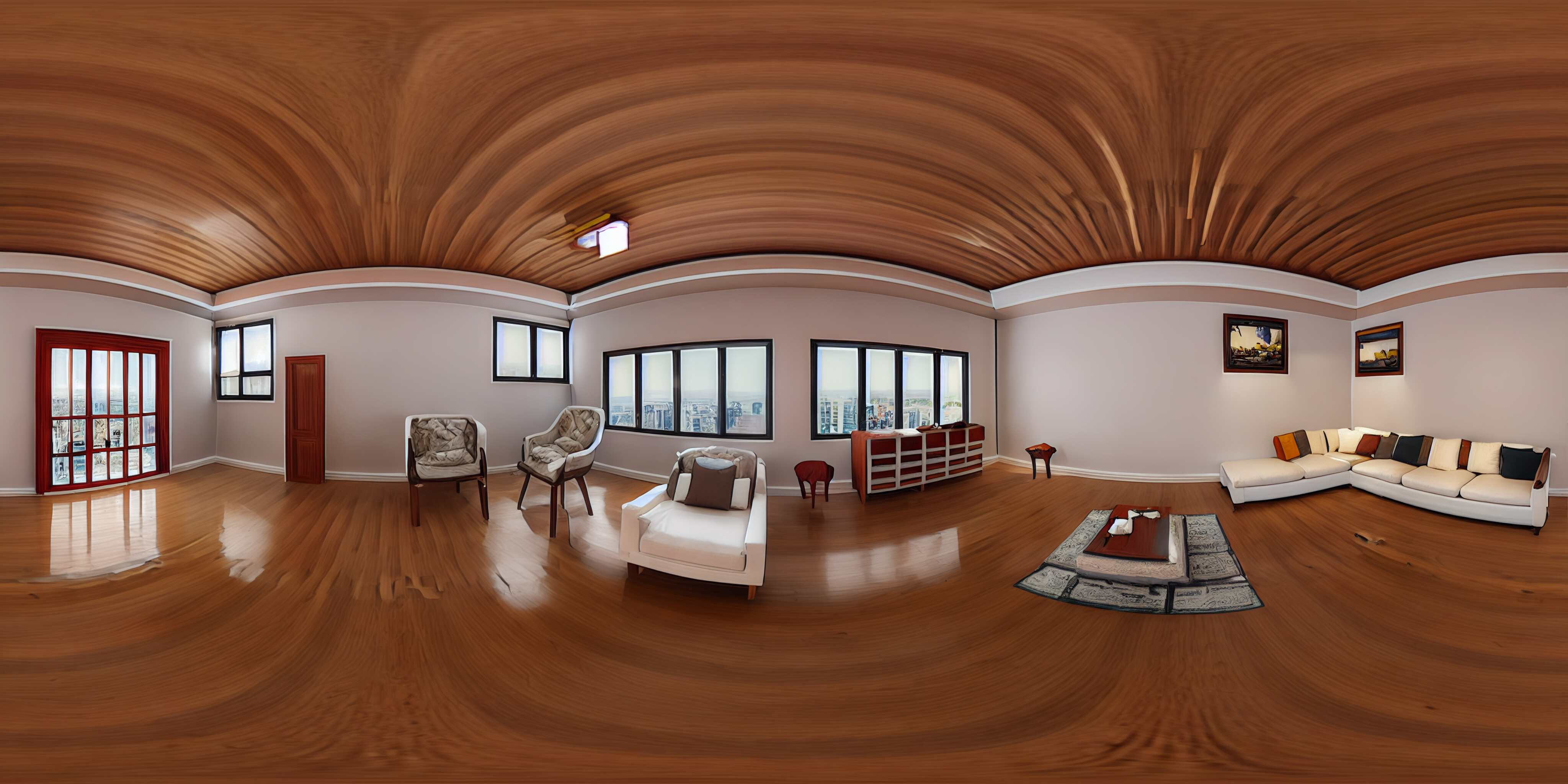} &
\includegraphics[width=0.20\linewidth]{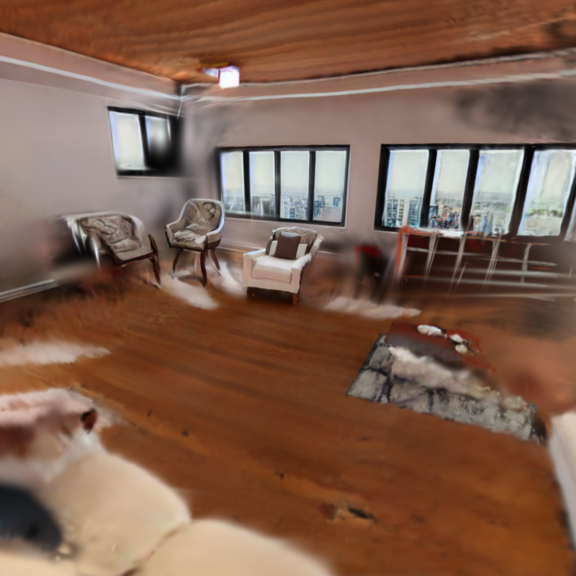} &
\includegraphics[width=0.20\linewidth]{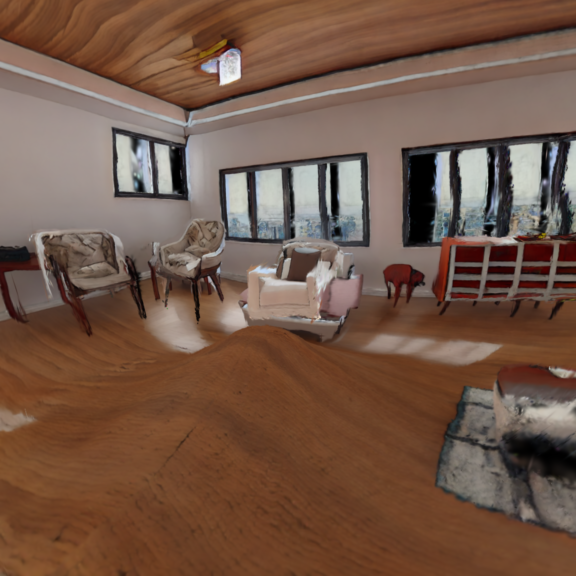} &
\includegraphics[width=0.20\linewidth]{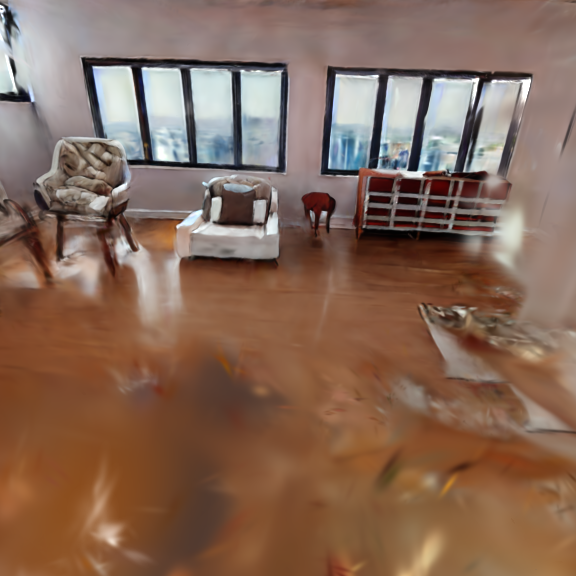} &
\includegraphics[width=0.20\linewidth]{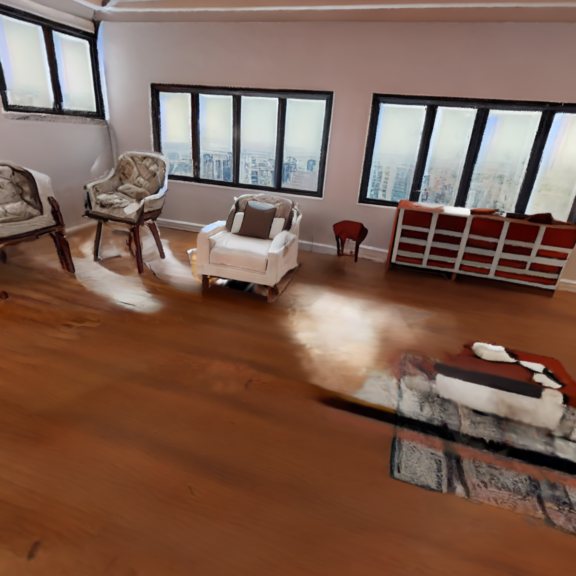}\\

\includegraphics[width=0.20\linewidth]{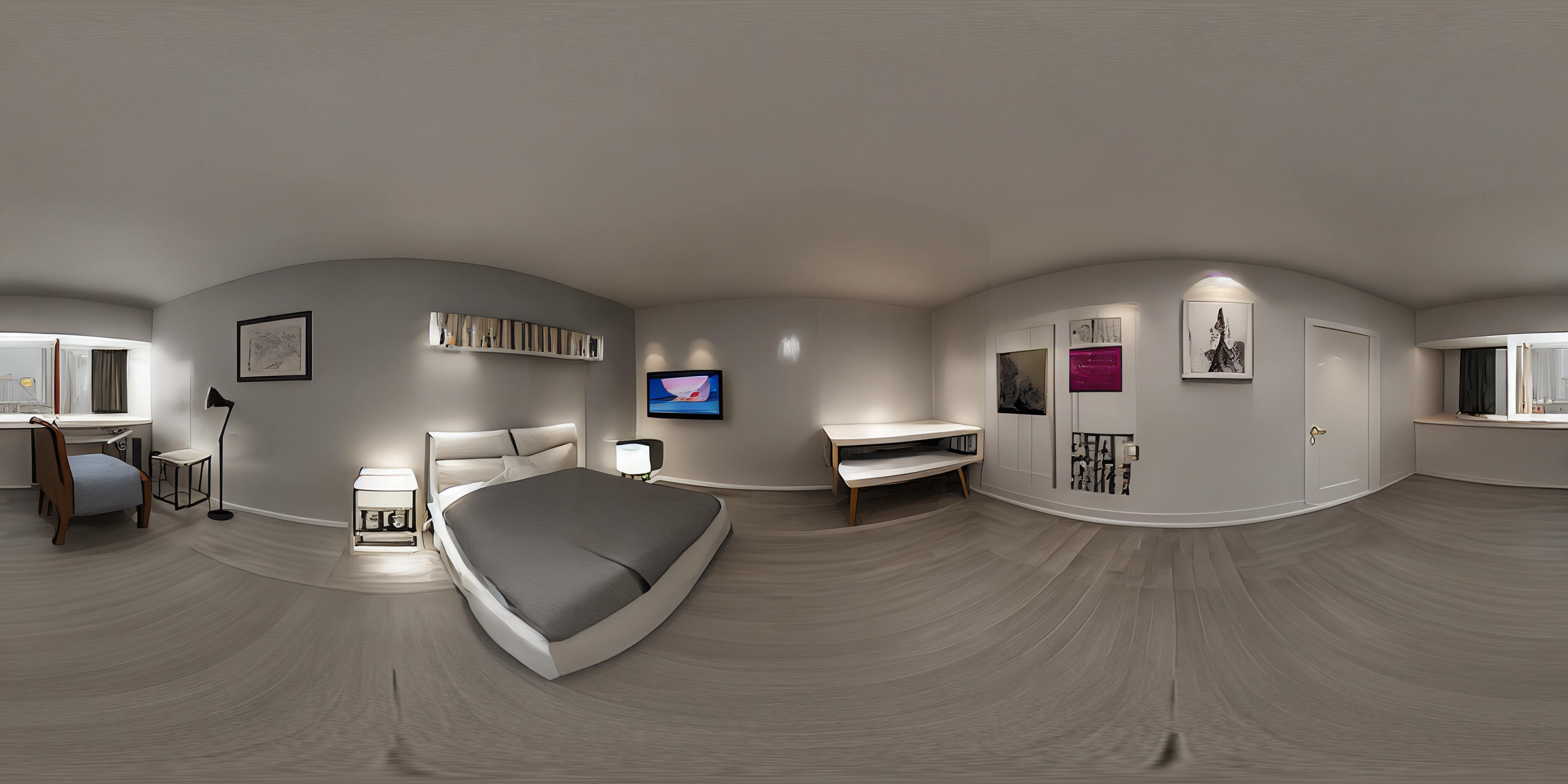} &
\includegraphics[width=0.20\linewidth]{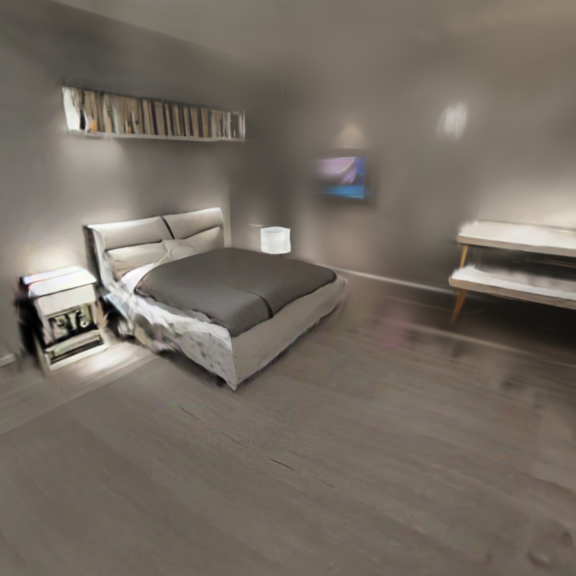} &
\includegraphics[width=0.20\linewidth]{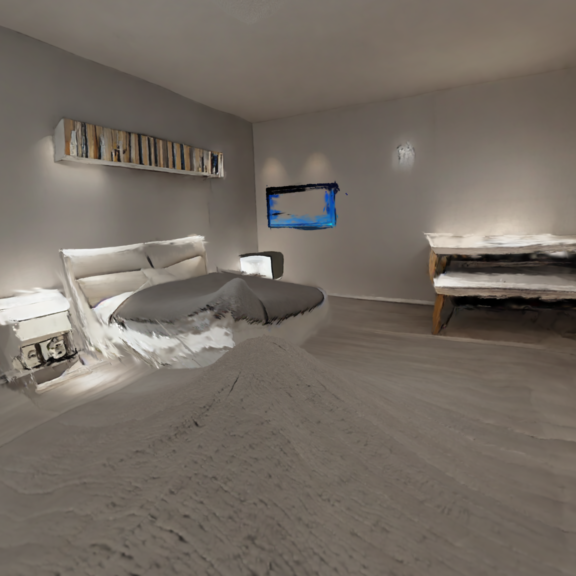} &
\includegraphics[width=0.20\linewidth]{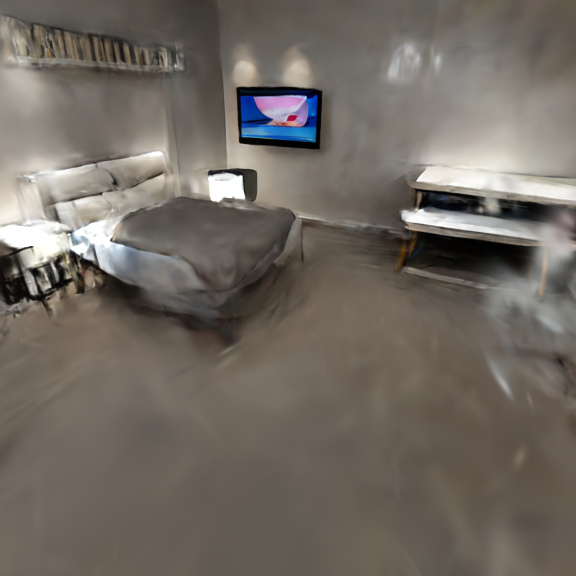} &
\includegraphics[width=0.20\linewidth]{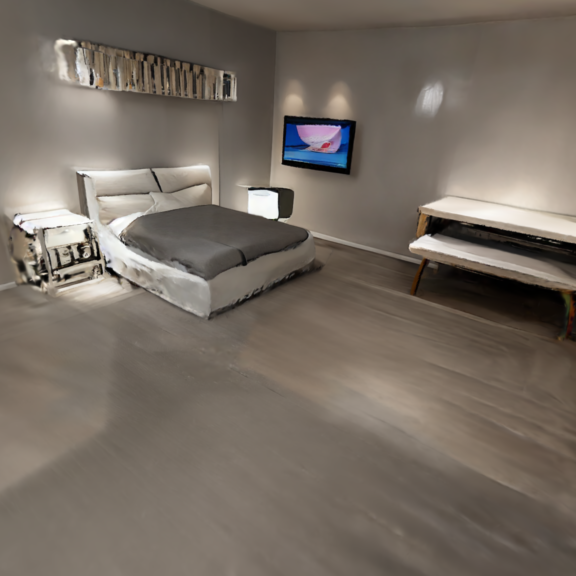}\\

\includegraphics[width=0.20\linewidth]{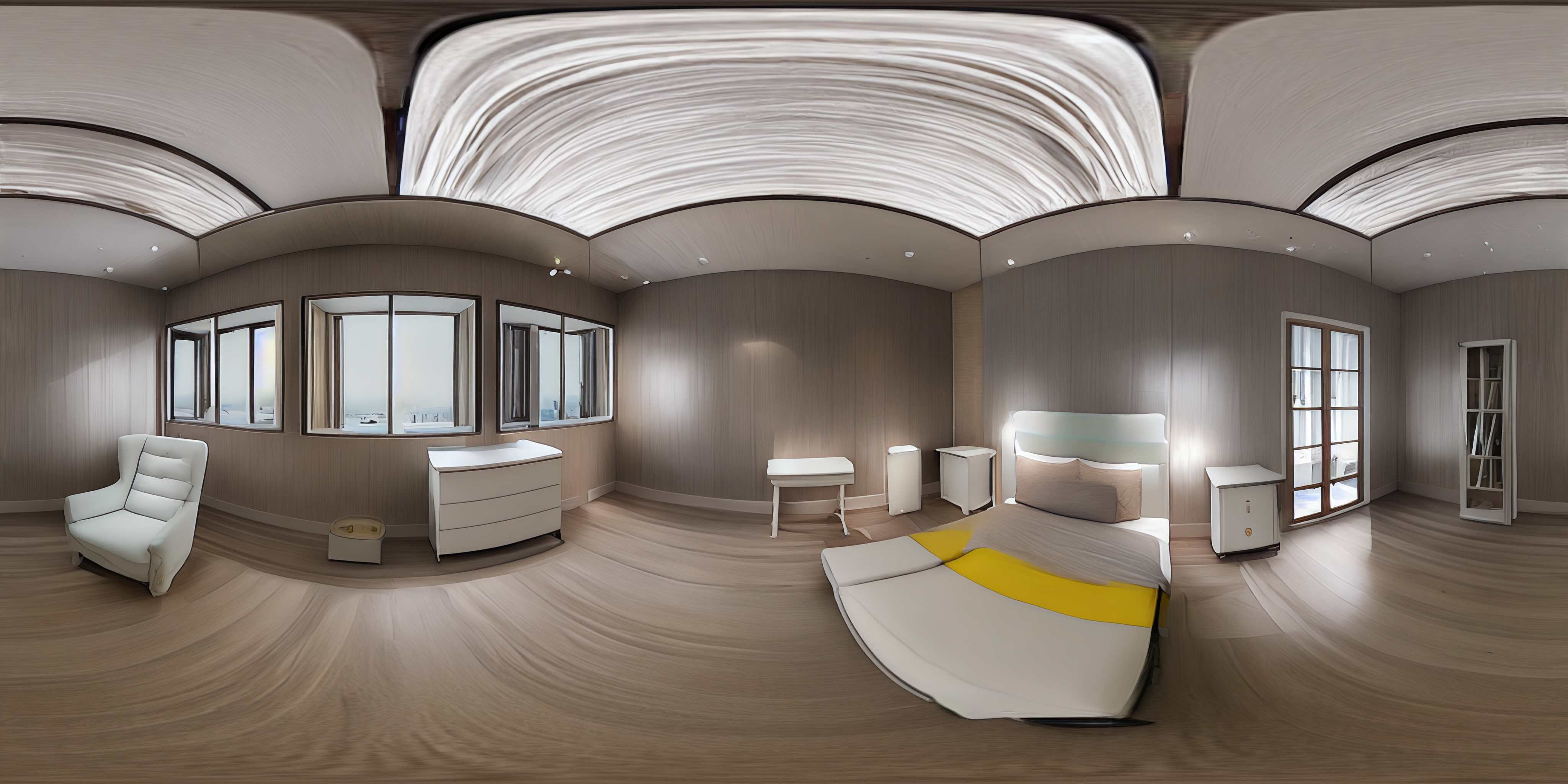} &
\includegraphics[width=0.20\linewidth]{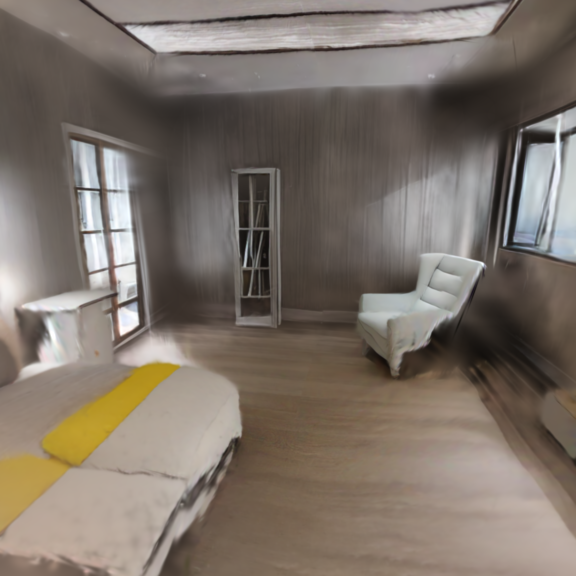} &
\includegraphics[width=0.20\linewidth]{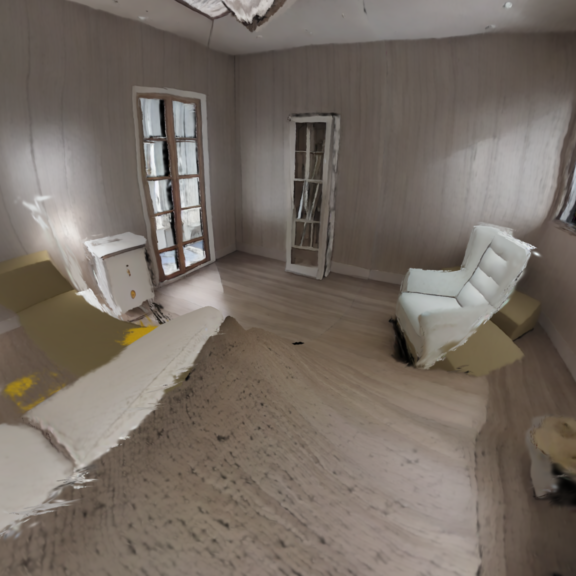} &
\includegraphics[width=0.20\linewidth]{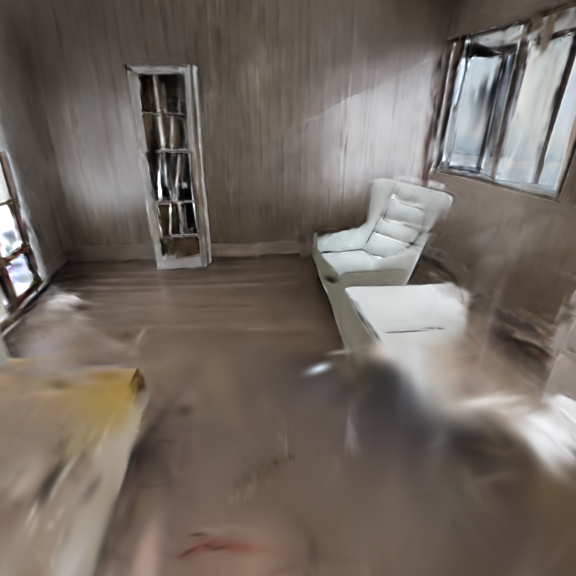} &
\includegraphics[width=0.20\linewidth]{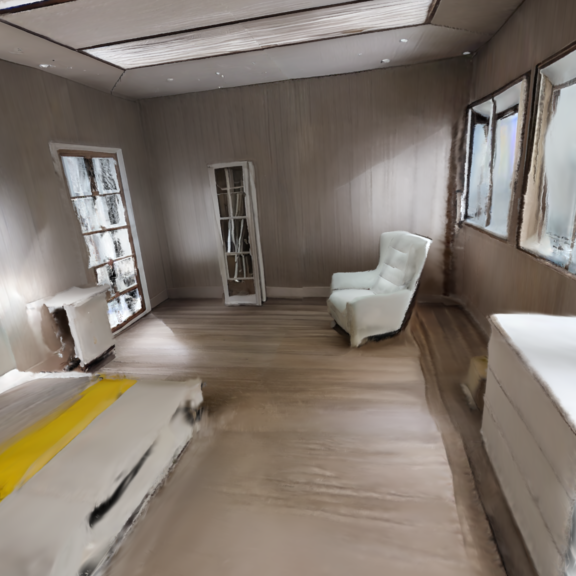}\\

 Initial Panorama & DreamScene360 & LayerPano3D & WorldExplorer & Ours \\
 
\end{tabular}
}
\caption{\textbf{Additional visual comparison of 3D reconstructions for different methods.} The visualizations show rendered novel views from comparable camera positions for each method. Note that the camera poses for the rendered novel views highly deviate from the training views to showcase the limitations of the different methods.  
}
\label{fig:qual_results_supp}
\end{figure*}

\paragraph{Additional Qualitative Results for 3D Reconstruction.}
Figure~\ref{fig:qual_results_supp} shows additional qualitative results for NVS from 3D reconstructions for different methods. 
Furthermore, a side-by-side comparison of rendered videos for different methods, as well as a free flowing camera path through an multi-room scene is provided in the supplementary video.

\newcommand{\zoomA}{(-0.6,-0.4)}
\newcommand{\zoomB}{(0.8,-0.4)}
\newcommand{\zoomD}{(1.1,-0.2)}
\newcommand{\zoomC}{(0.25,0.7)}
\begin{figure}
\centering
\scalebox{0.45}{
\begin{tabular}{cccccc}
\doublezoom{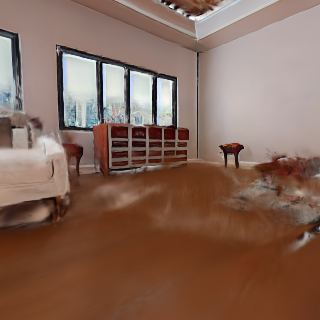}{\zoomA}{\zoomB} &
\doublezoom{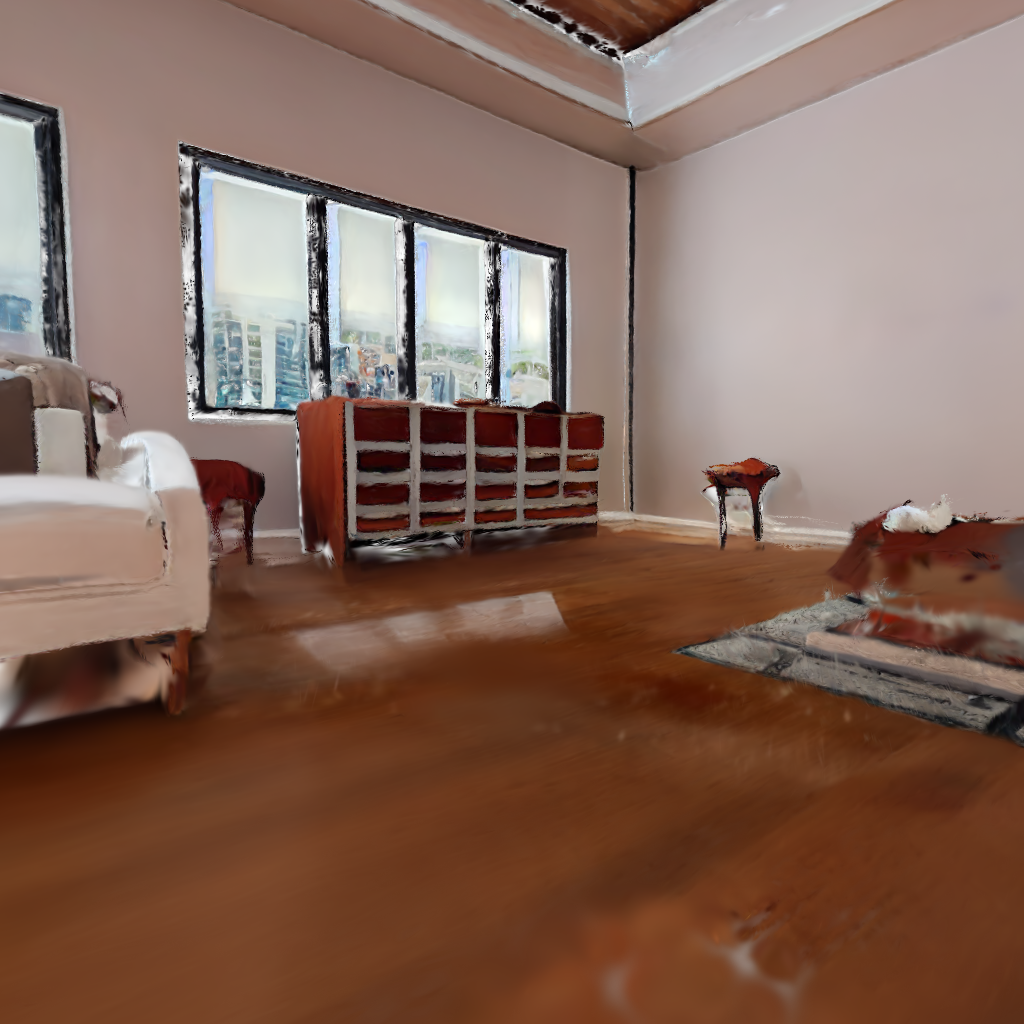}{\zoomA}{\zoomB} &
\doublezoom{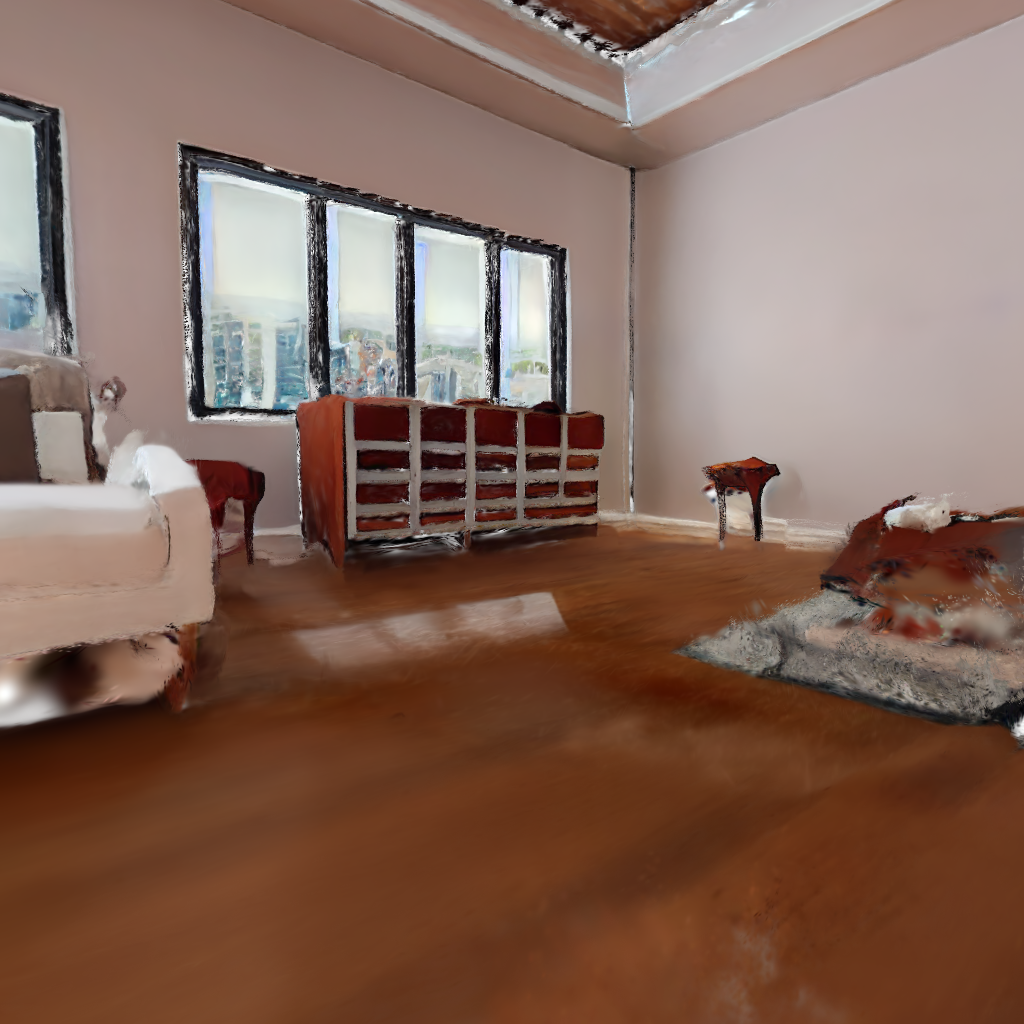}{\zoomA}{\zoomB} &
\doublezoom{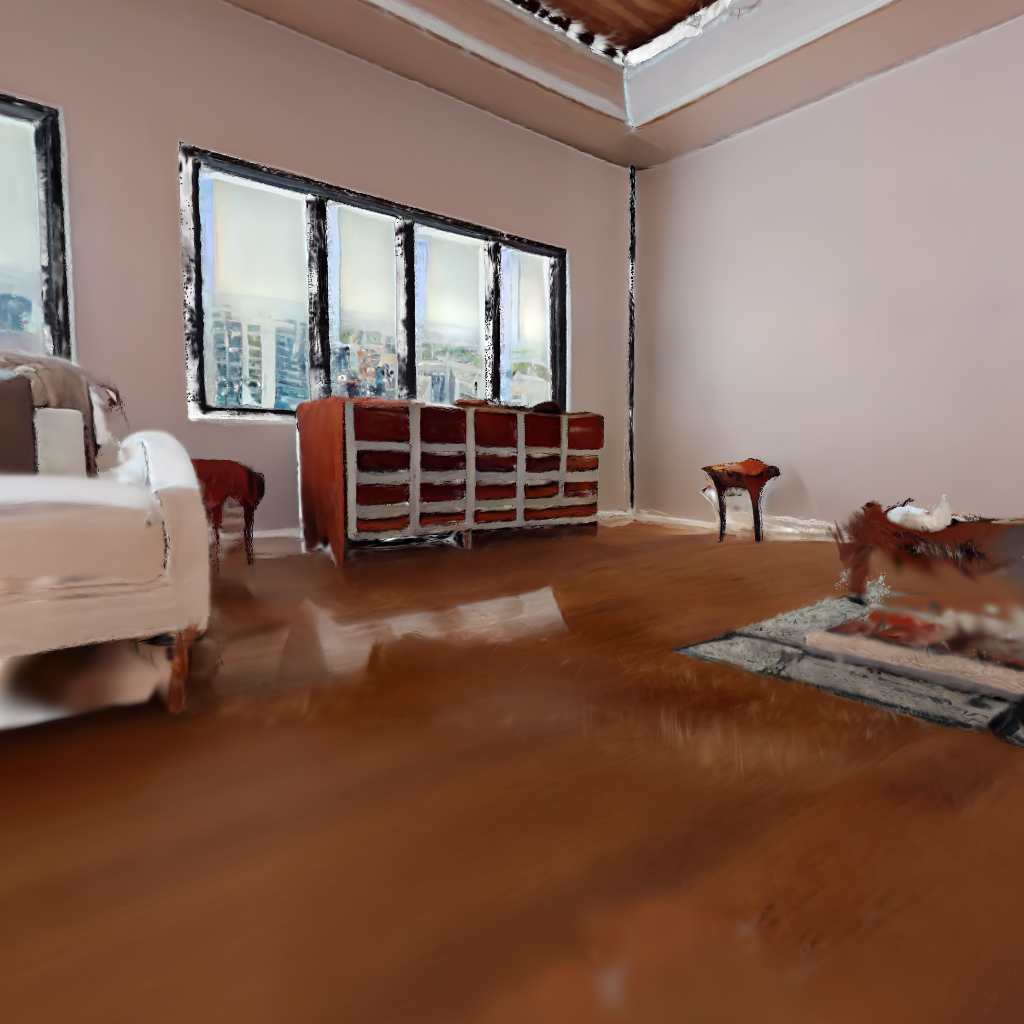}{\zoomA}{\zoomB} &
\doublezoom{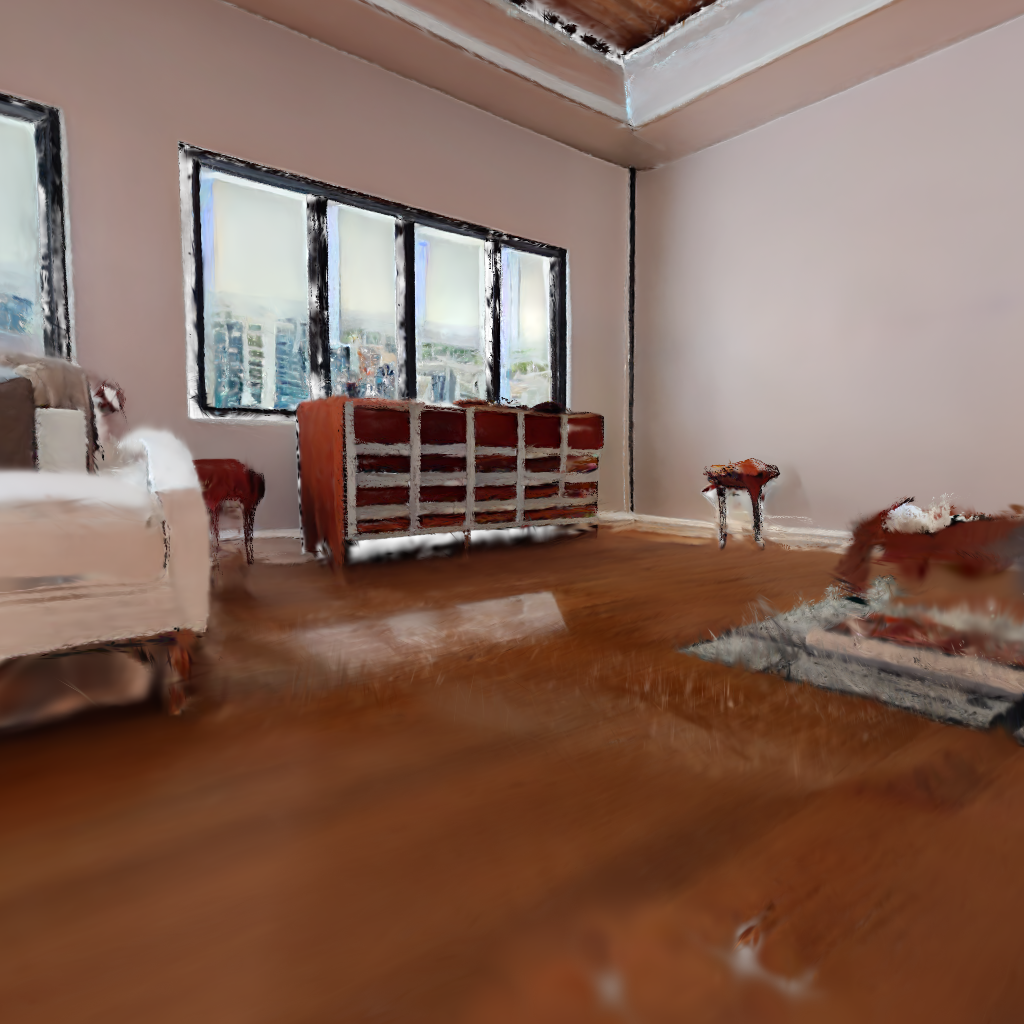}{\zoomA}{\zoomB} \\

\doublezoom{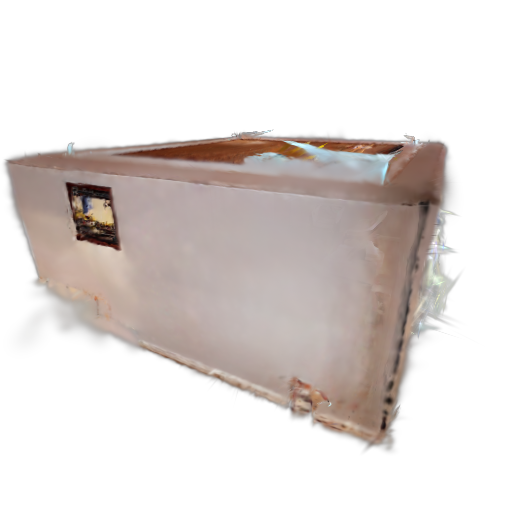}{\zoomC}{\zoomD} &
\doublezoom{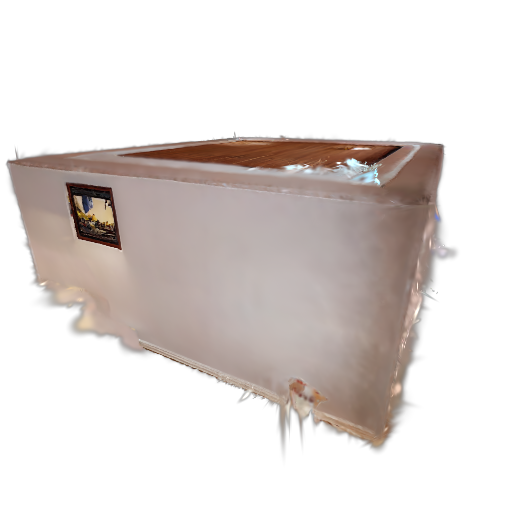}{\zoomC}{\zoomD)} &
\doublezoom{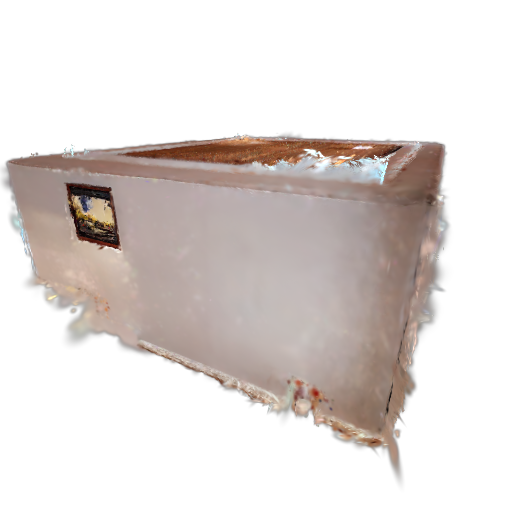}{\zoomC}{\zoomD} &
\doublezoom{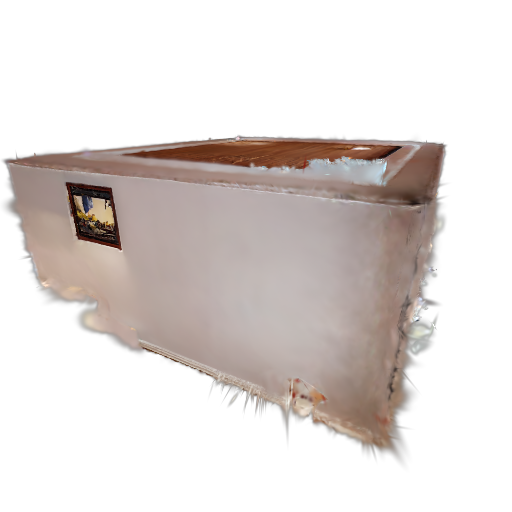}{\zoomC}{\zoomD} &
\doublezoom{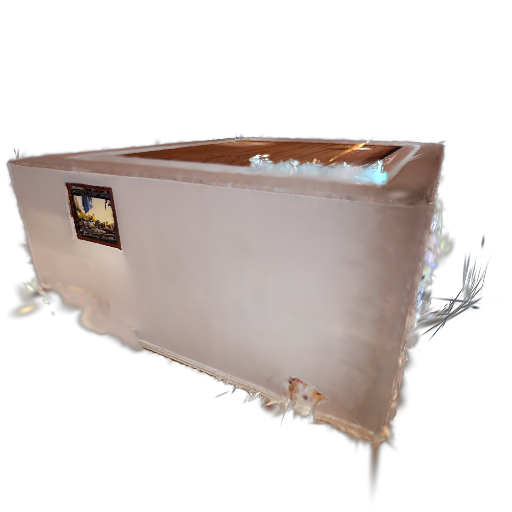}{\zoomC}{\zoomD} \\

3DGS & Ours & Ours w/o $\mathcal{L}_{NN}$ & Ours w/o $\mathcal{L}_{depth}$ & Ours w/o $\mathcal{L}_{geom}$\\

\end{tabular}
}
\caption{\textbf{Comparison of 3D reconstruction results using different loss terms.} First, we show the 3D reconstruction using the original 3DGS~\cite{kerbl20233d} loss for reference. Next, we show the results using our loss as defined in Equation~\ref{eq:our_full_loss}. For comparison, we show the results when we remove a specific loss term. In general, $\mathcal{L}_{NN}$ improves scene bounding, $\mathcal{L}_{depth}$ removes wobbles at reflections and $\mathcal{L}_{geom}$ reduces spikes by aligning the normals of the Gaussians with the scene geometry.}
\label{fig:losses2}
\end{figure}

\section{Additional Results for Guided Panorama Generation}

\begin{figure*}[]
\centering
\scalebox{0.98}{
\includegraphics[trim={0.0cm 1.7cm .5cm .0cm},clip,width=\linewidth]{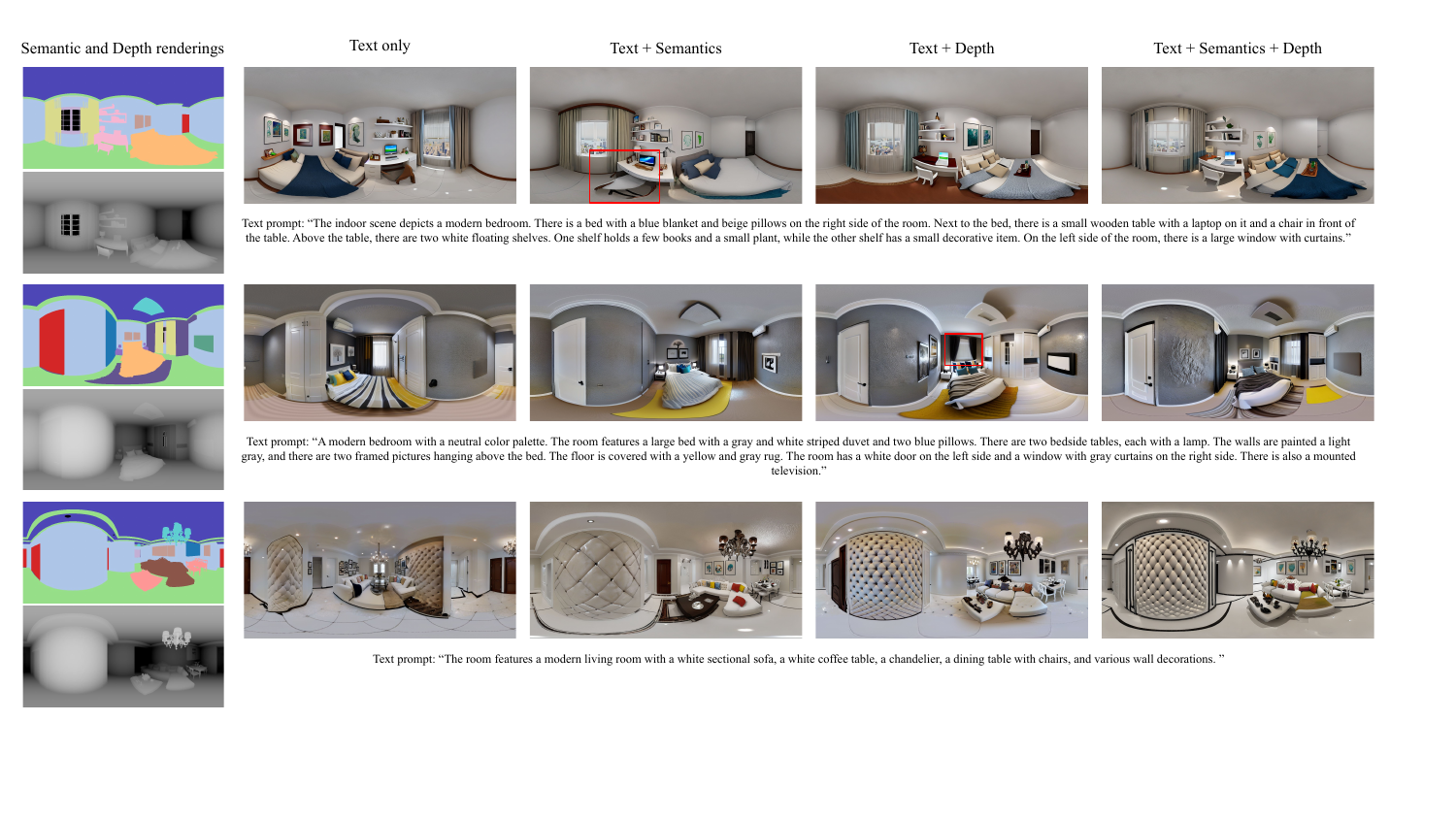}
}
\caption{\textbf{Qualitative validation of generated panoramic image with and without depth and semantic ControlNets.} When only using the text prompt as input, the resulting panoramic image significantly differs from the depth and semantic renderings, making this unusable for our approach of generating a 3D proxy to ground the subsequent generation. Using semantics in addition to the text prompt already leads to a good alignment of generated scenes, however it can still generate degenerate object shapes (see first row highlighted in red) due to missing 3D geometry information. In contrast,  using the text prompt and the depth map results in generating detailed geometry and accurate object poses, but can mix up object types with very similar geometry (see second row highlighted in red: a window above the bed instead of two framed pictures). Our approach, using text as well as semantics and depth as guidance signals eliminates both shortcomings.}
\label{fig:ablation_controlnet}
\end{figure*}

\paragraph{Influence of Depth and Semantic ControlNet.}
Figure~\ref{fig:ablation_controlnet} shows a qualitative ablation of the depth and semantic ControlNet branches of our guided panoramic image generator. These results show that our approach of using text, as well as semantics and depth input as guidance generates semantically and geometrically coherent panoramic images, strongly adhering to the 3D proxy layout.

\begin{figure}[]
\centering
\scalebox{0.98}{
\includegraphics[trim={0.0cm 1.7cm .5cm .0cm},clip,width=\linewidth]{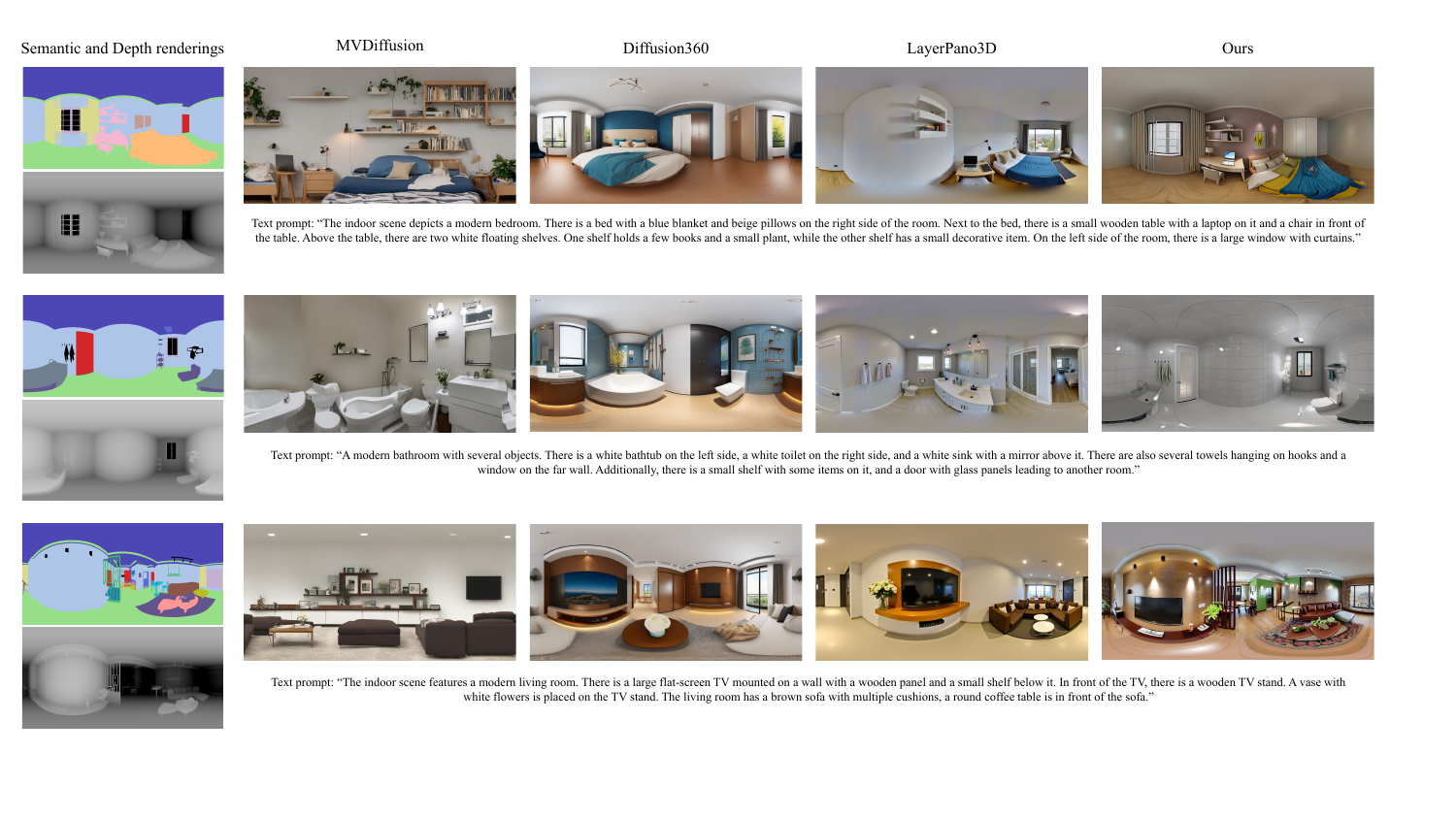}
}
\caption{\textbf{Visual comparison of generated panoramic images from different panorama generators.} Each row depicts a separate example, whereas we first show the semantic and depth renderings corresponding to the text prompt, and then the generated panoramic image from the different methods. The visualization highlights limitations of different panorama generators: MVDiffusion~\cite{tang2023MVDiffusion} often fails to generate plausible scene layouts. Diffusion360~\cite{feng2023diffusion360} tends to ignore important objects described in the text prompt, for example the table with laptop in the first row. This is also backed by the low CLIPScore in the corresponding quantitative evaluation in the main paper. LayerPano3D~\cite{yang2024layerpano3d} often generates panorama images with blurry borders on the top and the bottom (see first and second row), which leads to artifacts when used for 3D reconstruction. Our guided panoramic image generator is able to accurately follow the text prompt and to generate plausible layouts by utilizing the depth and semantic guidance signals provided by our global scene layout.
}
\label{fig:pano_comparison}
\end{figure}

\paragraph{Qualitative Results for Guided Panorama Generation.} To show the influence of our guided panorama generation, we use text prompts extracted from examples of the Structured3D test set as input for different panoramic image generators. 
Figure~\ref{fig:pano_comparison} shows a visual comparison of the resulting images. While all methods are able to generate high quality and photorealistic 360\degree\ images, competitor methods often struggle to accurately follow the text prompt, or to deliver plausible and complete layouts.

\section{Details on Depth-guided Trajectory Alignment}
\paragraph{Binary Search Algorithm for Camera Scale Parameter.}
Algorithm~\ref{alg:bin_search} illustrates the implementation of our bisection search for the optimal camera scale parameter based on the alignment between depth renderings of the 3D proxy layout and generated NVS frames from Stable Virtual Camera~\cite{zhou2025stable}.

\begin{algorithm}
\caption{Coarse-to-Fine Bisection Search for Camera Scale Parameter Optimization} \label{alg:bin_search}
\begin{algorithmic}
\Input Initial camera scale candidates $[\theta_{cs}^{\min}, \theta_{cs}^{\max}]$, resolution \\ schedule $\mathcal{R} = \{r_1, \dots, r_K\}$
\Output Optimal camera scale parameter $\theta_{cs}^\ast$

\State Initialize search candidates $\Theta_0 \gets [\theta_{cs}^{\min}, \theta_{cs}^{\max}]$
\State Initialize empty sets $\Theta^\ast$, $\mathcal{L}^\ast$
\State Interval range $l \gets  \theta_{cs}^{\max} - \theta_{cs}^{\min}$

\For{$k = 1$ to $K$}
    \State $\theta_{cs,k}^\ast \gets 
    \arg\min\limits_{\theta_{cs} \in \Theta_{k-1}} \mathcal{L}_{dpt}(\theta_{cs})$
    \State $\Theta^\ast \gets \Theta^\ast \cup \{\theta_{cs,k}^\ast\}$

    \State $\hat{\theta}_{cs,k} \gets \arg\min\limits_{\theta_{cs} \in \Theta^\ast} \mathcal{L}_{dpt}(\theta_{cs})$
    
    \State Update search interval:\[
        \Theta_k \gets 
        \left[
        \hat{\theta}_{cs,k} - \frac{l}{r_k},
        \;
        \hat{\theta}_{cs,k} + \frac{l}{r_k}
        \right]
    \]
\EndFor

\State \Return $\theta_{cs}^\ast \gets \hat{\theta}_{cs,k}$
\end{algorithmic}
\end{algorithm}

\paragraph{Influence of Resolution Schedule Depth on Depth Alignment Accuracy.} We selected a Resolution Schedule of $\mathcal{R} = \{2,4,8,16\}$ to efficiently narrow down the search space for the optimal camera scale parameter. Table~\ref{tab:camera_scale_results} shows the depth error during our camera scale estimation in Algorithm~\ref{alg:bin_search} for a representative example.

\begin{table}[]
\centering
\scalebox{0.85}{
\begin{tabular}{c c c c}
\toprule
\textbf{Resolution $r$} &
\textbf{Candidates for $\theta_{cs}$} &
\textbf{Selected $\theta_{cs}$} &
$\boldsymbol{\mathcal{L}_{dpt}}$ \\ 
\midrule
$1$   & $[0.8,\;1.60]$         & $1.60$ & $0.199$ \\
$2$   & $[1.20,\;1.60,\;2.00]$ & $1.20$ & $0.157$ \\
$4$   & $[1.00,\;1.20,\;1.40]$ & $1.20$ & $0.157$ \\
$8$   & $[1.10,\;1.20,\;1.30]$ & $1.30$ & $0.095$ \\
$16$  & $[1.25,\;1.30,\;1.35]$ & $1.25$ & $0.094$ \\
\bottomrule
\end{tabular}
}
\vspace{3pt}
\caption{Depth alignment error $\mathcal{L}_{dpt}$ over the coarse-to-fine resolution schedule for different camera scale parameter settings for a common trajectory.}
\label{tab:camera_scale_results}
\end{table}

\section{Ablation for Trajectory Estimation Algorithm}
\label{sec:view_planning}
This sections describes and compares a mesh-based view planning approach for camera trajectory estimation, which we see as valid alternative to our heuristic trajectory sampling described in the main paper. Given a triangle mesh of an indoor scene obtained by Holodeck~\cite{Yang_2024_CVPR_holodeck}, the aim is to select a compact set of camera viewpoints and a collision-free trajectory through them that covers the scene surfaces for downstream multi-view reconstruction. 
\paragraph{Algorithm.}
We pose this as a weighted face-coverage problem solved by greedy selection, followed by a path-aware traversal that stays inside the room's free space. We rasterize the floor polygon and a top-down footprint of the mesh into an occupancy grid and apply a clearance margin, yielding a walkable mask, then place candidate camera poses on the free cells at a fixed eye height. For each candidate we cast a low-resolution ray grid against the mesh and mark a face \emph{covered} when its first hit lies within a usable distance band and below a maximum incidence angle. We pick a view set by area-weighted greedy maximum coverage and order the views by a nearest-neighbor traveling-salesman tour with shortest-path ($A^{\ast}$) edge costs on the walkable grid, so transitions stay in free space. We use the same protocol as described in the main paper for creating four trajectories (one per quadrant): each samples within one quadrant but scores against the full mesh, and the resulting anchors are linked by shortest paths, resampled to a fixed number of frames at uniform spacing along the path. Additionally, we implemented a \emph{depth-aware} variant using the same per-view minimum-depth filter as implemented in our main algorithm.
\paragraph{Experiments.}
We evaluate with four planned $42$-frame quadrant trajectories. Table~\ref{tab:vp_coverage} shows the final MEt3R score and scene coverage for all algorithms. The depth-aware variant of the standard view planning algorithm trades coverage ($74.29{\to}60.01\%$) for quality, leading to a higher 3D consistency score. The experiment shows that optimizing coverage converges to non-overlapping views, while 3D reconstruction benefits from overlapping views from different viewpoints. Due to this, our proposed algorithm leads to the highest MEt3R score compared to the simple baselines.

\begin{table}[]
\centering
\setlength{\tabcolsep}{4pt}
\scalebox{0.9}{
\begin{tabular}{lcc}
\toprule
Algorithm & MEt3R $\downarrow$ & Coverage (\%)\\
\midrule
Standard view planning & 0.029 & 74.29 \\
Standard view planning with depth filter & \underline{0.027}& 60.01 \\
\midrule
Ours& \textbf{0.024}  & 62.39\\
\bottomrule
\end{tabular}
}
\vspace{3pt}
\caption{\textbf{Comparison of different algorithms for trajectory sampling.}}
\label{tab:vp_coverage}
\end{table}
\section{Additional Visualizations for Scene Expansion}
\begin{figure*}
\centering
\scalebox{0.9}{
\begin{tabular}{cccc}

\includegraphics[width=0.24\linewidth]{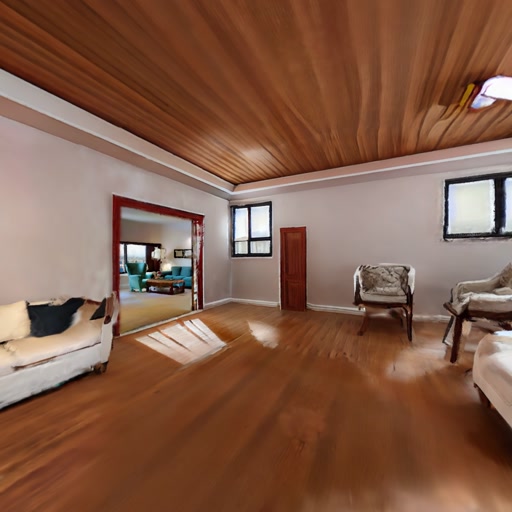} &
\includegraphics[width=0.24\linewidth]{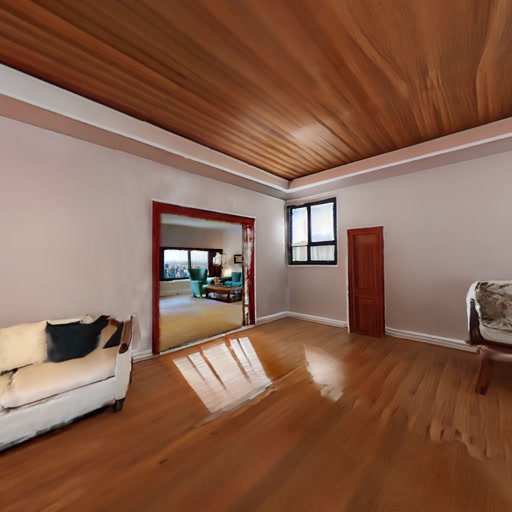} &
\includegraphics[width=0.24\linewidth]{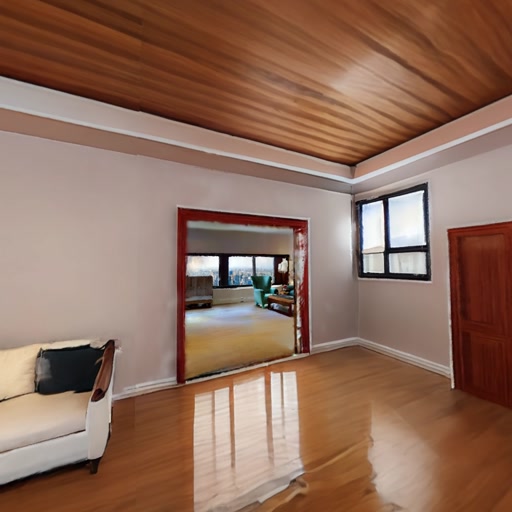}&
\includegraphics[width=0.24\linewidth]{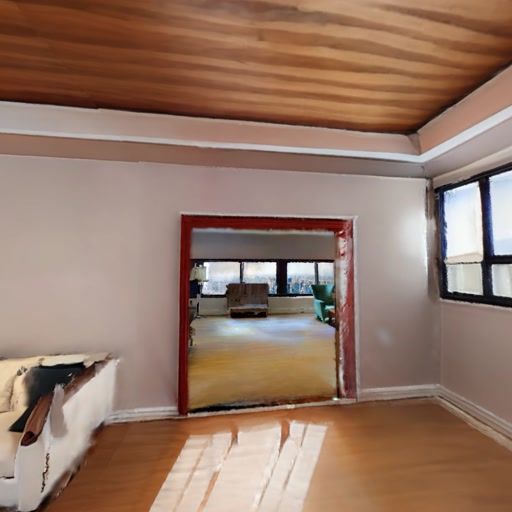} \\

\includegraphics[width=0.24\linewidth]{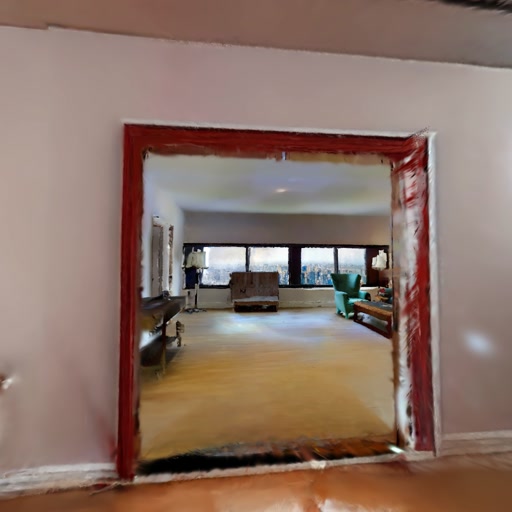} &
\includegraphics[width=0.24\linewidth]{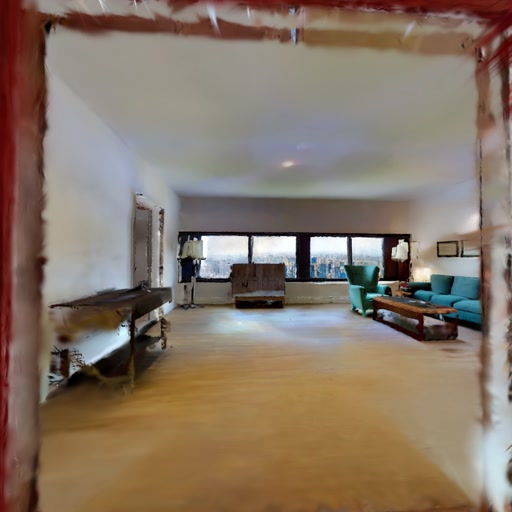} &
\includegraphics[width=0.24\linewidth]{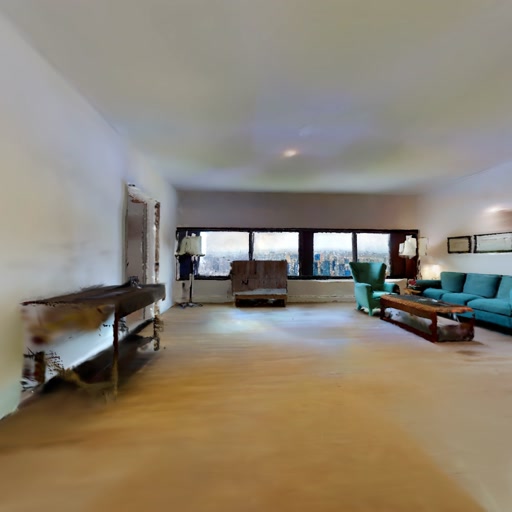}&
\includegraphics[width=0.24\linewidth]{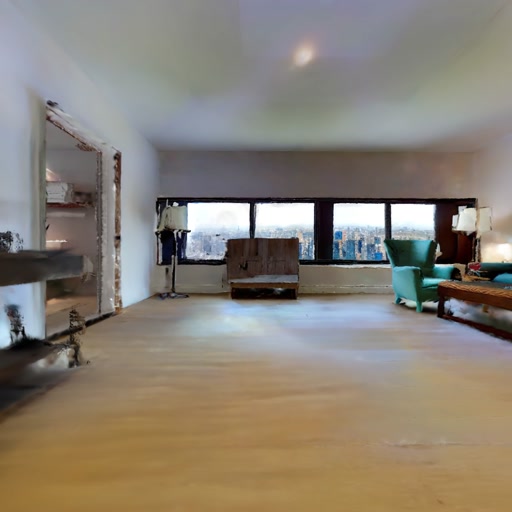} \\

\includegraphics[width=0.24\linewidth]{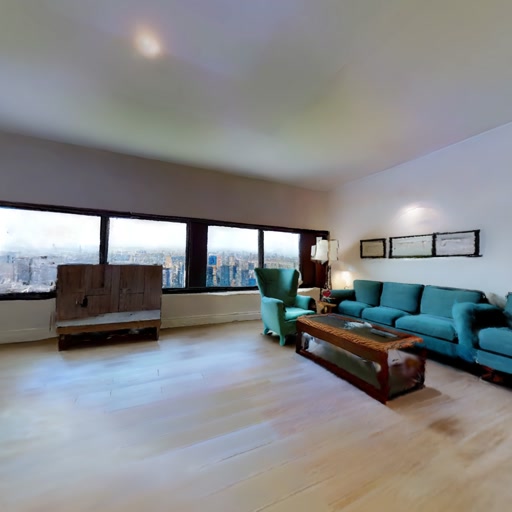} &
\includegraphics[width=0.24\linewidth]{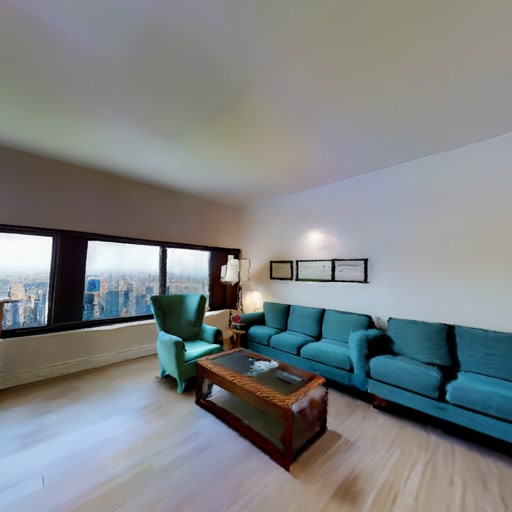} &
\includegraphics[width=0.24\linewidth]{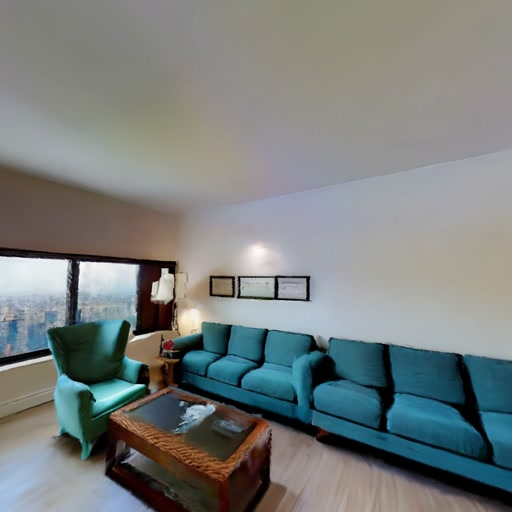}&
\includegraphics[width=0.24\linewidth]{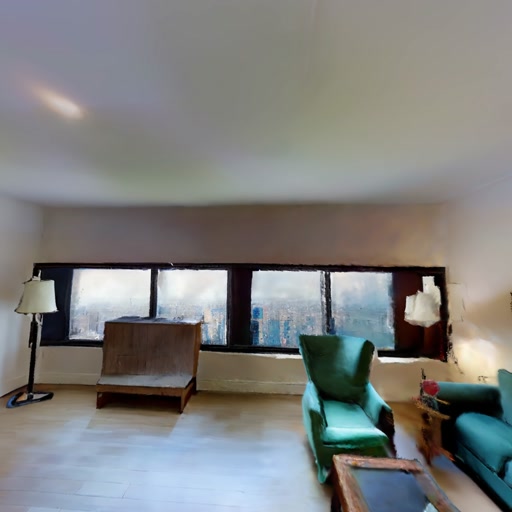} \\

\includegraphics[width=0.24\linewidth]{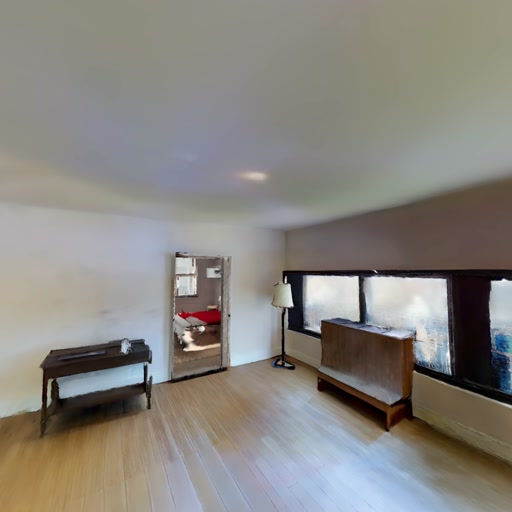} &
\includegraphics[width=0.24\linewidth]{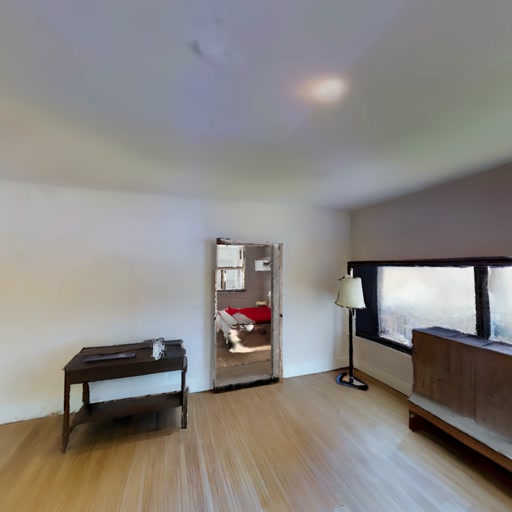} &
\includegraphics[width=0.24\linewidth]{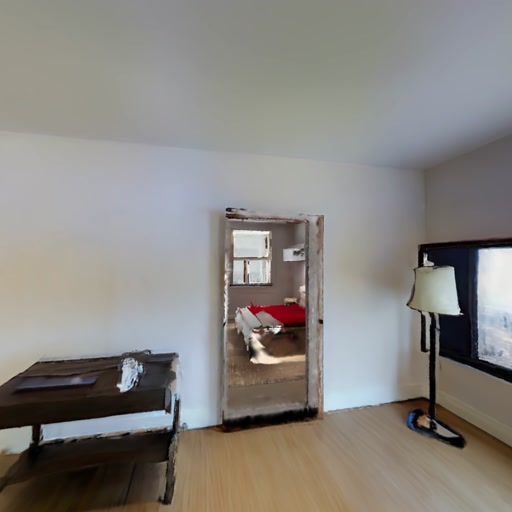}&
\includegraphics[width=0.24\linewidth]{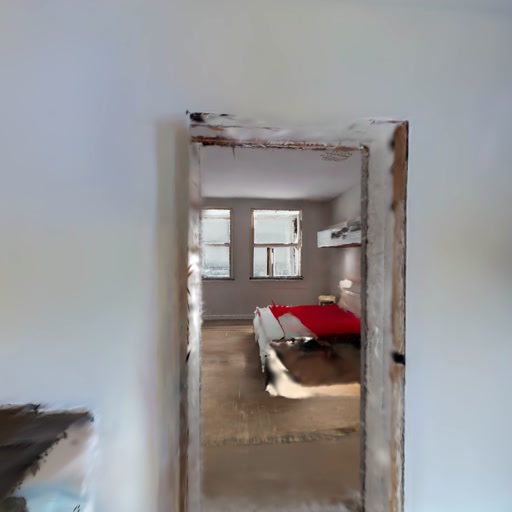} \\

\includegraphics[width=0.24\linewidth]{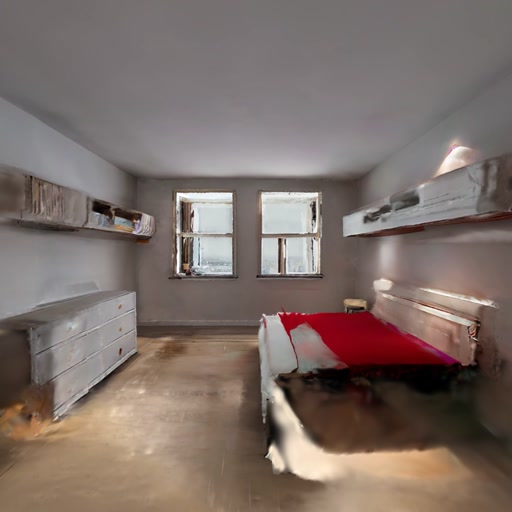} &
\includegraphics[width=0.24\linewidth]{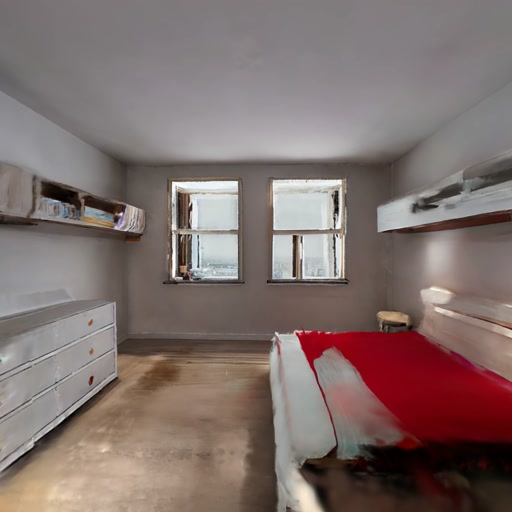} &
\includegraphics[width=0.24\linewidth]{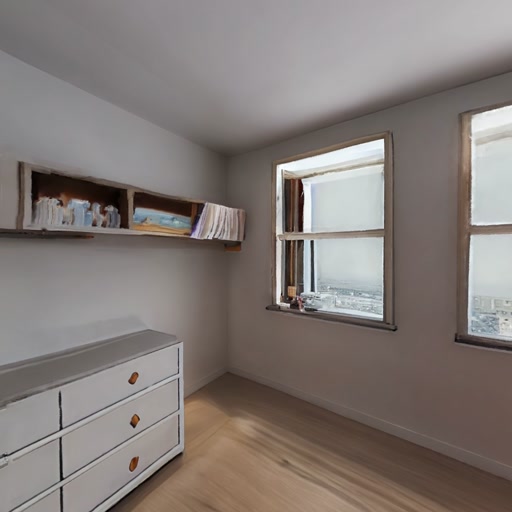}&
\includegraphics[width=0.24\linewidth]{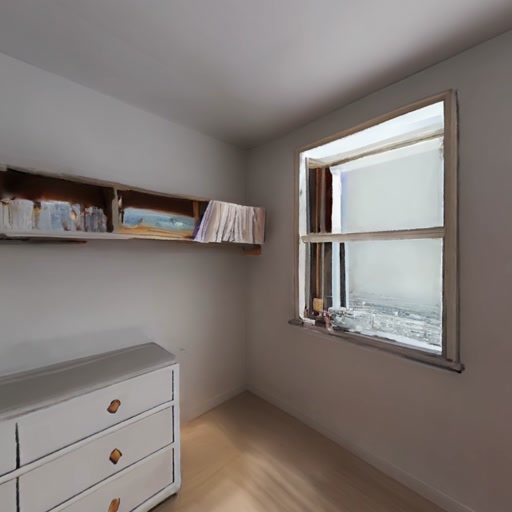} \\
\end{tabular}
}
\caption{\textbf{Visualization of a free-flowing camera path through an expanded scene.} From top left to bottom right, the visualization shows selected frames of a free-flowing camera path through a 3-room scene generated with our scene expansion algorithm.}
\label{fig:scene_exp_vis}
\end{figure*}

Figure~\ref{fig:scene_exp_vis} shows captured frames from a free-flowing camera path through a progressively expanded scene.

\section{Additional Information about the User Study}
For the user study, we collected 2945 data points from 31 participants with medium to high computer vision/graphics knowledge. The participants were shown a 15 second long video of a camera orbiting at a consistent distance from the origin and were asked to assess the following criteria:
\begin{itemize}
    \item  \textbf{Perceptual Quality:} From 1 to 5, how would you rate the quality of the scene? Does the video look plausible and perceptually accurate?
    \item \textbf{3D Geometric Consistency:} From 1 to 5, how complete and good is the 3D geometry (floor, ceiling, wall) and individual objects?
    \item \textbf{Overall Scene Quality:} From 1 to 5, how would you rate the scene overall? You might consider personal taste.
\end{itemize}
Each video received 3 scores from 1 to 5. To evaluate the statistical significance of the user study, we first analyze the variance of the different groups to see if they differ from one another (ANOVA F-statistic $113$, p-value $<0.05$). We then compare all methods pairwise and compute the p-value using the adjusted significance threshold, all comparisons were significant with p $<0.00833$.

\end{document}